\documentclass[a4paper, 11pt, oneside]{Thesis}  
\graphicspath{Figures/}  
\usepackage{multirow}
\usepackage{epigraph}
\usepackage[round]{natbib}
\usepackage{lipsum}
\usepackage{xspace}
\newcommand{\system}[1]{\textsc{#1}\xspace}
\newcommand{\Geo}{\mbox{\textsc{GeoQuery}}\xspace}
\newcommand{\Jobs}{\mbox{\textsc{Jobs}}\xspace}
\newcommand{\Atis}{\mbox{\textsc{ATIS}}\xspace}

\newcommand{\gen}[1]{\textit{GEN}(#1)\xspace}
\newcommand{\reduce}[1]{\textit{REDUCE}(#1)\xspace}
\newcommand{\parser}{ProtoParser\xspace}
\newcommand{\pre}[1]{\textsl{#1}\xspace}
\newcommand{\data}[1]{\textsc{#1}\xspace}
\newcommand{\atis}{\data{ATIS}}
\newcommand{\geoquery}{\data{GeoQuery}}
\newcommand{\jobs}{\data{Jobs}}

\newcommand{\geode}{\data{GeoQuery(De)}}
\newcommand{\geoel}{\data{GeoQuery(El)}}
\newcommand{\geoth}{\data{GeoQuery(Th)}}

\newcommand{\geo}{\data{GeoQuery}}
\newcommand{\nlmap}{\data{NLMap}}
\newcommand{\nlmapde}{\data{NLMap(De)}}

\newcommand{\almsp}{\system{AL-MSP}}

\newcommand{\rand}{\system{Random}}
\newcommand{\cluster}{\system{Cluster}}
\newcommand{\csse}{\system{CSSE}}
\newcommand{\rttl}{\system{RTTL}}
\newcommand{\lcsfw}{\system{LCS(FW)}}
\newcommand{\ssfw}{\system{S2S(FW)}}
\newcommand{\lcsbw}{\system{LCS(BW)}}
\newcommand{\traffic}{\system{Traffic}}
\newcommand{\oracle}{\system{Oracle}}
\newcommand{\amspnbest}{\system{ABE(n-best)}}
\newcommand{\amspmax}{\system{ABE(max)}}
\newcommand{\amsp}{\system{ABE}}

\newcommand{\lfsd}{\system{LFSD}}
\newcommand{\lcd}{\system{LCD}}
\newcommand{\lfslcd}{\system{LFS-LC-D}}

\newcommand{\maxcompound}{\system{Max Compound}}

\newcommand{\coarsefine}{\system{Coarse2Fine}}
\newcommand{\ptmaml}{\system{PT-MAML}}

\newcommand{\da}{\system{DA}}
\newcommand{\irnet}{\system{IRNet}}
\newcommand{\ratsql}{\system{RATSQL}}
\newcommand{\pt}{\textit{pt}}
\newcommand{\os}{\textit{os}}
\newcommand{\comb}{\textit{cb}}

\newcommand{\bertlstm}{\system{BERT-LSTM}}

\newcommand{\Seq}{\mbox{\textsc{Seq2Seq}}\xspace}
\newcommand{\seqseq}{\mbox{\textsc{Seq2Seq}}\xspace}

\newcommand{\blfs}{\system{DLFS}}
\newcommand{\aemr}{\system{TR}}
\newcommand{\awu}{\system{FSCL}}
\newcommand{\TR}{\system{TR}}
\newcommand{\tr}{\system{Total Recall}}

\newcommand{\dlfs}{\system{DLFS}}

\newcommand{\overnight}{\data{Overnight}}
\newcommand{\nlmapc}{\data{NLMap(city)}}
\newcommand{\nlmapq}{\data{NLMap(qt)}}

\newcommand{\finetune}{\system{Fine-tune}}
\newcommand{\hatt}{\system{HAT}}
\newcommand{\ewc}{\system{EWC}}

\newcommand{\gem}{\system{GEM}}
\newcommand{\emar}{\system{EMAR}}
\newcommand{\arper}{\system{ARPER}}
\newcommand{\emr}{\system{EMR}}
\newcommand{\proto}{\system{ProtoParser}}

\newcommand{\btmetric}[1]{BT Discrepancy Rate}

\newcommand{\gss}{\system{GSS}}
\newcommand{\balance}{\system{Balance}}
\newcommand{\fss}{\system{FSS}}
\newcommand{\prior}{\system{Prior}}
\newcommand{\lfs}{\system{LFS}}
\newcommand{\random}{\system{Random}}
\newcommand{\blfst}{\system{DLFS}}
\newcommand{\aemrt}{\system{TR}}

\usepackage[ruled,vlined]{algorithm2e}

\usepackage{amsthm,amsmath,amsfonts,bm,xspace}
\usepackage{color}

\newcommand{\comment}[1]{}

\def\eqref#1{(\ref{#1})}

\def\1{\bm{1}}

\def\va{{\bm{a}}}

\def\vc{{\bm{c}}}
\def\vd{{\bm{d}}}
\def\ve{{\bm{e}}}

\def\vx{{\bm{x}}}
\def\vy{{\bm{y}}}
\def\vz{{\bm{z}}}

\DeclareMathAlphabet{\mathsfit}{\encodingdefault}{\sfdefault}{m}{sl}
\SetMathAlphabet{\mathsfit}{bold}{\encodingdefault}{\sfdefault}{bx}{n}

\DeclareMathOperator*{\argmax}{arg\,max}
\DeclareMathOperator*{\argmin}{arg\,min}

\usepackage{graphicx}
\usepackage[dvipsnames]{xcolor}
\usepackage{verbatim}  
\usepackage{vector}  
\hypersetup{urlcolor=blue, colorlinks=true}  
\usepackage{listings}

\lstdefinestyle{simple_lst_style}{
    columns=flexible,
    keywordstyle=\color{red},
    numberstyle=\color{gray},
    stringstyle=\color{green},
    basicstyle=\ttfamily\small,
    identifierstyle=\color{black},
    commentstyle=\color{blue},
    breakatwhitespace=false,
    breaklines=true,
    captionpos=b,
    keepspaces=false,
    numbersep=5pt,
    showspaces=false,
    showstringspaces=false,
    showtabs=false,
    tabsize=2,
    frame=single,
}

\lstdefinelanguage{Prompt}{
    morekeywords={Human, Computer},
    sensitive=true,
    morestring=[b]",
}

\lstdefinelanguage{LFPrompt}{
    morekeywords={Utterance, LF},
    sensitive=true,
    morestring=[b]",
}

\begin{document}
\frontmatter      

\UNIVERSITY{{Monash University}}
%
\school{{School of Information Technology}}
\gradtime{{2023}}

%
\title  {Semantic Parsing in Limited Resource Conditions}
\authors  {\texorpdfstring
            {\href{zhuang.li@monash.edu}{Zhuang Li}}
            {Zhuang Li}
            }
\addresses  {\groupname\\\deptname\\\univname}  
\date       {\today}
\subject    {}
\keywords   {Semantic Parsing, Limited Resource, Neural Network}

\maketitle

\setstretch{1.3}  

\fancyhead{}  
\rhead{\thepage}  
\lhead{}  

\pagestyle{fancy}  

\copyrightnotice{







\textsuperscript{\textcopyright  \authornames (2023).}

I certify that I have made all reasonable efforts to secure copyright permissions for third-party content included in this thesis and have not knowingly added copyright content to my work without the owner's permission.

}
\clearpage  

\abstract{
\addtocontents{toc}{}  

Semantic parsing has become important in facilitating human-system interaction due to its unique ability to convert natural language utterances into structured semantic representations. These representations can then be executed against systems to obtain the desired responses for human users. However, the resources available for training an effective semantic parser are sometimes limited. Most supervised semantic parsers are data-hungry, and collecting training data for them is often challenging due to the strong domain-specific knowledge required by data annotators. Moreover, limited computational resources can constrain semantic parsing, especially when an on-device parser requires frequent retraining to adapt to new tasks. Such processes can incur high computational costs that local devices might struggle with. Given these resource constraints, this thesis aims to address semantic parsing challenges under limited data and computational conditions using techniques from automatic data curation, knowledge transfer, active learning, and continual learning.

The first challenge this thesis addresses involves adapting the parser to a new task when there is no or only limited parallel training data available in the target domain or task. In the absence of parallel data for a new task, the thesis proposes a method that uses structured data from a database to automatically generate a significant number of synthetic parallel training examples, thereby training the semantic parser for this task. If abundant parallel training data exists from a source task but only limited data from the target task, the thesis introduces a technique that utilizes the knowledge from the source task to enhance the semantic parser for the new task.

In the second scenario, the thesis addresses multilingual situations where no data are available in the target language and there is a limited annotation budget for that language. However, abundant data exist in the source language. This thesis proposes a method to adapt multilingual parsers to the target languages using a limited human translation budget. The thesis leverages active learning to sample source-language examples for manual translation. The translation of these selected samples aims to maximize the performance of the multilingual parser in the target language. In a more relaxed scenario, where both a manual translation budget and a machine translation service are accessible, this thesis suggests an alternative method. This approach uses a machine translation system to translate all source-language data, and subsequently, an active learning approach is adopted to obtain human-translated data, which supplements the machine-translated data. A parser trained on a combination of machine- and human-translated data typically surpasses those trained exclusively on machine-translated or purely human-translated data.

In situations constrained by computational resources, this thesis introduces a continual learning approach to minimize the training time and computational memory required by the semantic parser when adapting the parser for a new task. At the same time, this method ensures that the efficiency of the parser on previously seen tasks is preserved, mitigating catastrophic forgetting and fostering knowledge transfer between tasks.

Overall, this thesis paves the way for using various techniques to address challenges associated with semantic parsing under low resource conditions.


}

\clearpage  

\Declaration{
\addtocontents{toc}{}  

This thesis is an original work of my research and contains no material which has been accepted for the award of any other degree or diploma at any university or equivalent institution and that, to the best of my knowledge and belief, this thesis contains no material previously published or written by another person, except where due reference is made in the text of the thesis.
\vfil\vfil\null
Signature:\\
\rule[1em]{25em}{0.5pt}  

Print Name: Zhuang Li\\
\rule[1em]{25em}{0.5pt}  

Date:\\
\rule[1em]{25em}{0.5pt}  
}

\clearpage  


\Publicationslist{

Publications during PhD candidature on which the thesis is based:
\begin{itemize}
   \item \cref{chap:back} previously presented in:\\ \textbf{Z. Li}, L. Qu, G. Haffari. ``Context Dependent Semantic Parsing: A Survey'' The 28th International Conference on Computational Linguistics (COLING). 2020.
      \item \cref{chap:auto} previously presented in:\\  F. Shiri$^{*}$, T. Y. Zhuo$^{*}$, \textbf{Z. Li}$^{*}$, V. Nguyen, S. Pan, W. Wang, G. Haffari, Y. F. Li. ``Paraphrase Techniques for Maritime QA System'' $^{*}$ denotes the equal contribution. 25th International Conference on Information Fusion (FUSION). 2022.
     \item \cref{chap:meta} previously presented in:\\ \textbf{Z. Li}, L. Qu, S. Huang, G. Haffari. ``Few-shot Semantic Parsing for New Predicates'' The 16th conference of the European Chapter of the Association for Computational Linguistics (EACL). 2021.
        \item \cref{chap:continual} previously presented in:\\ \textbf{Z. Li}, L. Qu, G. Haffari. ``Total Recall: a Customized Continual Learning Method for Neural Semantic Parsers'' The 2021 Conference on Empirical Methods in Natural Language Processing (EMNLP). 2021.
\end{itemize}

Other articles during PhD candidature not relevant to the thesis:
\begin{itemize}
\item \textbf{Z. Li}, L. Qu, Q. Xu, T. Wu, T. Zhan, G. Haffari. ``Variational Autoencoder with Disentanglement Priors for Low-Resource Task-Specific Natural Language Generation'' The 2022 Conference on Empirical Methods in Natural Language Processing (EMNLP). 2022.
\item S. Huang,  \textbf{Z. Li}, L. Qu, G. Haffari. ``On Robustness of Neural Semantic Parsers'' The 16th conference of the European Chapter of the Association for Computational Linguistics (EACL). 2021.
\item T. Wu, M. Caccia,  \textbf{Z. Li}, Y. F. Li, G. Qi, G. Haffari ``Pretrained language model in continual learning: A comparative study'' The 2022 International Conference on Learning Representations (ICLR). 2022.
\end{itemize}

}

\Publicationsafterlist{

Publications that arose from the thesis during the examination period, following the completion of PhD candidature:
\begin{itemize}
      \item Subsequent to PhD enrollment, research findings from~\cref{chap:al_msp}, along with additional new results, have been published in the following paper:\\  \textbf{Z. Li}, G. Haffari. ``Active Learning for Multilingual Semantic Parsing'' Findings of the 17th conference of the European Chapter of the Association for Computational Linguistics (EACL). 2023. \\
      \item Subsequent to PhD enrollment, research findings from~\cref{chap:comb_human_auto}, along with additional new results, have been published in the following paper:\\ \textbf{Z. Li}, L. Qu, P. R. Cohen, R. V. Tumuluri, G. Haffari ``The Best of Both Worlds: Combining Human and Machine Translations for Multilingual Semantic Parsing with Active Learning'' The 61st Annual Meeting of the Association for Computational Linguistics (ACL). 2023.
\end{itemize}
}

\clearpage  


\setstretch{1.3}  
\ack{
\addtocontents{toc}{}  

I am profoundly grateful to Prof. Gholamreza (Reza) Haffari, Dr. Lizhen Qu, and Dr. Phil Cohen. Without their guidance, this thesis would not have been possible. Dr. Lizhen Qu and Dr. Phil Cohen's decision to bring me to Monash University marked the beginning of an enlightening three-year Ph.D. journey. I am also indebted to Prof. Reza Haffari for welcoming me into his lab and introducing me to a remarkable community of Ph.D. students. His unwavering support during the challenging times of the COVID era was invaluable. Their collective guidance enriched not only my research but also my daily life.

Special acknowledgment goes to my esteemed panel members: Prof. Guido Tack, Dr. Mario Boley, Prof. Shirui Pan, and Dr. Lan Du, as well as the Director of Graduate Research, Prof. Arun Konagurthu. Their incisive feedback and insights have been indispensable. I was fortunate to intern at both Adobe and Openstream and wish to extend my gratitude to my mentor at Adobe, Dr. Quan Hung Tran, as well as to my colleagues at Openstream for their constructive suggestions and guidance.

The Faculty of Information Technology Research Scholarship played a pivotal role in funding my research endeavors. Additionally, I owe much to the computational resources provided by the Multi-modal Australian Sciences Imaging and Visualisation Environment (MASSIVE). My appreciation extends to Julie Holden, whose ALS workshop significantly enhanced my writing capabilities. Diligent editors from Elite Editing Australia revised Chapters 1 and 2 to ensure they met the Australian Standards for Editing Practice D and E.

During the unprecedented challenges of the COVID era, a robust network of colleagues and fellow Ph.D. students from Monash and ANU proved invaluable. Specifically, I owe a debt of gratitude to Tongtong Wu, Qiongkai Xu, Terry Yue Zhuo, Shuo Huang, Michelle Zhao, Minghao Wu, Xuanli He, and Thuy-Trang Vu for their invaluable assistance. My journey was further enriched by the support from Haolan Zhan, Yunchen Hua, Yufei Wang, Farhad Moghimifar, Fahimeh Saleh, Xiaoyu Guo, Snow Situ, Lin Chuang, Yujin Huang, and Tao Feng.

Last but certainly not least, my deepest gratitude goes to my parents and my beloved Siyu. Their unconditional love and steadfast faith in me have empowered me to confront any challenge that comes my way.

}
\clearpage  

\pagestyle{fancy}  

\lhead{\emph{Contents}}  
\tableofcontents  

\lhead{\emph{List of Figures}}  
\listoffigures  

\lhead{\emph{List of Tables}}  
\listoftables  

\setstretch{1.5}  
\clearpage  
\lhead{\emph{Abbreviations}}  
\listofsymbols{ll}  
{

\textbf{AL} & \textbf{A}ctive \textbf{L}earning \\
\textbf{AST} & \textbf{A}bstract \textbf{S}yntax \textbf{T}ree \\
\textbf{CL} & \textbf{C}ontinual \textbf{L}earning \\
\textbf{FSL} & \textbf{F}ew-\textbf{S}ot \textbf{L}earning \\
\textbf{HT} & \textbf{H}uman \textbf{T}ranslation \\
\textbf{LF} & \textbf{L}ogical \textbf{F}orm \\
\textbf{MR} & \textbf{M}eaning \textbf{R}epresentation \\
\textbf{MSP} & \textbf{M}ultilingual \textbf{S}emantic \textbf{P}arsing \\
\textbf{MT} & \textbf{M}achine \textbf{T}ranslation \\
\textbf{NL} & \textbf{N}atural \textbf{L}anguage \\
\textbf{NLP} & \textbf{N}atural \textbf{L}anguage \textbf{P}rocessing \\
\textbf{NLU} & \textbf{N}atural \textbf{L}anguage \textbf{U}nderstanding \\
\textbf{RNN} & \textbf{R}ecurrent \textbf{N}eural \textbf{N}etwork \\
\textbf{SCFG} & \textbf{S}ynchronous \textbf{C}ontext-\textbf{F}ree \textbf{G}rammar \\
\textbf{SOTA} & \textbf{S}tate-\textbf{O}f-\textbf{T}he-\textbf{A}rt \\
\textbf{SQL} & \textbf{S}tructured \textbf{Q}uery \textbf{L}anguage

}




\makeatletter
\newcommand{\shiftleft}{\hspace*{-\@totalleftmargin}}
\makeatother
\mainmatter	  
\pagestyle{fancy}  


\fancyhead{}  
\rhead{\thepage}  
\lhead{}  
\chapter{Introduction}
\epigraph{Science is organized knowledge.}{\textit{Albert Einstein}}
\label{intro}
\begin{figure}
\centering
  \includegraphics[width=0.98\textwidth]{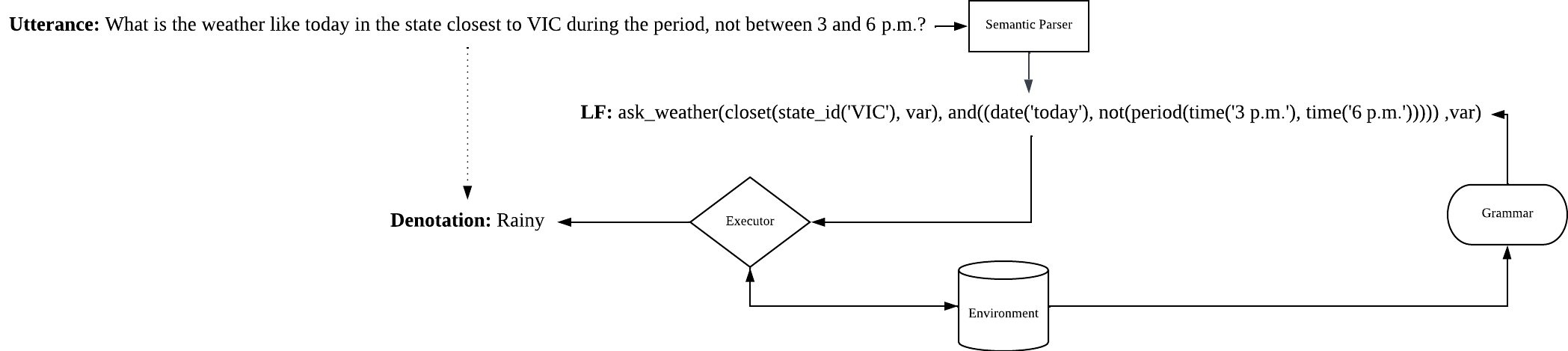}%
  \caption{An example of the semantic parsing process. \label{fig:sp_diagram}}
\end{figure}

\section{Background}
Due to their unambiguous representations of meanings, structured meaning representations (MRs) or logical forms (LFs), such as SQL, Prolog, and lambda calculus, are widely used in industrial applications, including expert and database systems, as well as in the research communities of logic and semantic webs. However, owing to the steep learning curve and the heavy manual effort required to manipulate LFs, non-technical users find it difficult to interact with the system using symbolic MRs/LFs. On the other hand, natural language (NL) is easy to use but ambiguous, such that the system cannot comprehend NL as humans do. Therefore, natural language understanding (NLU) applications need to be developed to serve as a bridge between humans and systems.

The semantic parser converts NL into machine-readable MRs/LFs, which makes it an ideal NLU interface between humans and machines. The LFs can then be executed for various purposes on the machine. For example, as depicted in~\fref{fig:sp_diagram}, the query `What is the weather like today in the state closest to VIC during the period, not between 3 and 6 p.m.?' is transformed into an LF and then executed against the database to obtain the answer. In terms of supporting complex human-system interactions, semantic parsing is also superior to other NLU tasks. For instance, the utterance in~\fref{fig:sp_diagram} contains complex compositional semantics, such as a superlative, a negation and a conjunction, which cannot be easily represented through shallow NLU annotations, such as intent and slots. A well-designed LF, conversely, can represent the meaning of such complex NLs. Because of its unique abilities, semantic parsing has now become popular in a variety of applications, such as those in smart speakers for answering questions, robotic navigation and understanding queries~\citep{kamath2018survey}.

With the advent of deep learning, semantic parsing has transitioned from the era of statistical models to that of neural models. Currently, neural semantic parsing models have attained state-of-the-art (SOTA) performance on most benchmarks and leaderboards. Nevertheless, despite its recent advancements, there are still open problems in neural semantic parsing due to data and computational resource limitations, which are discussed next.

\subsection{Limited Data Resources}
\begin{figure}
\centering
  \includegraphics[width=0.5\textwidth]{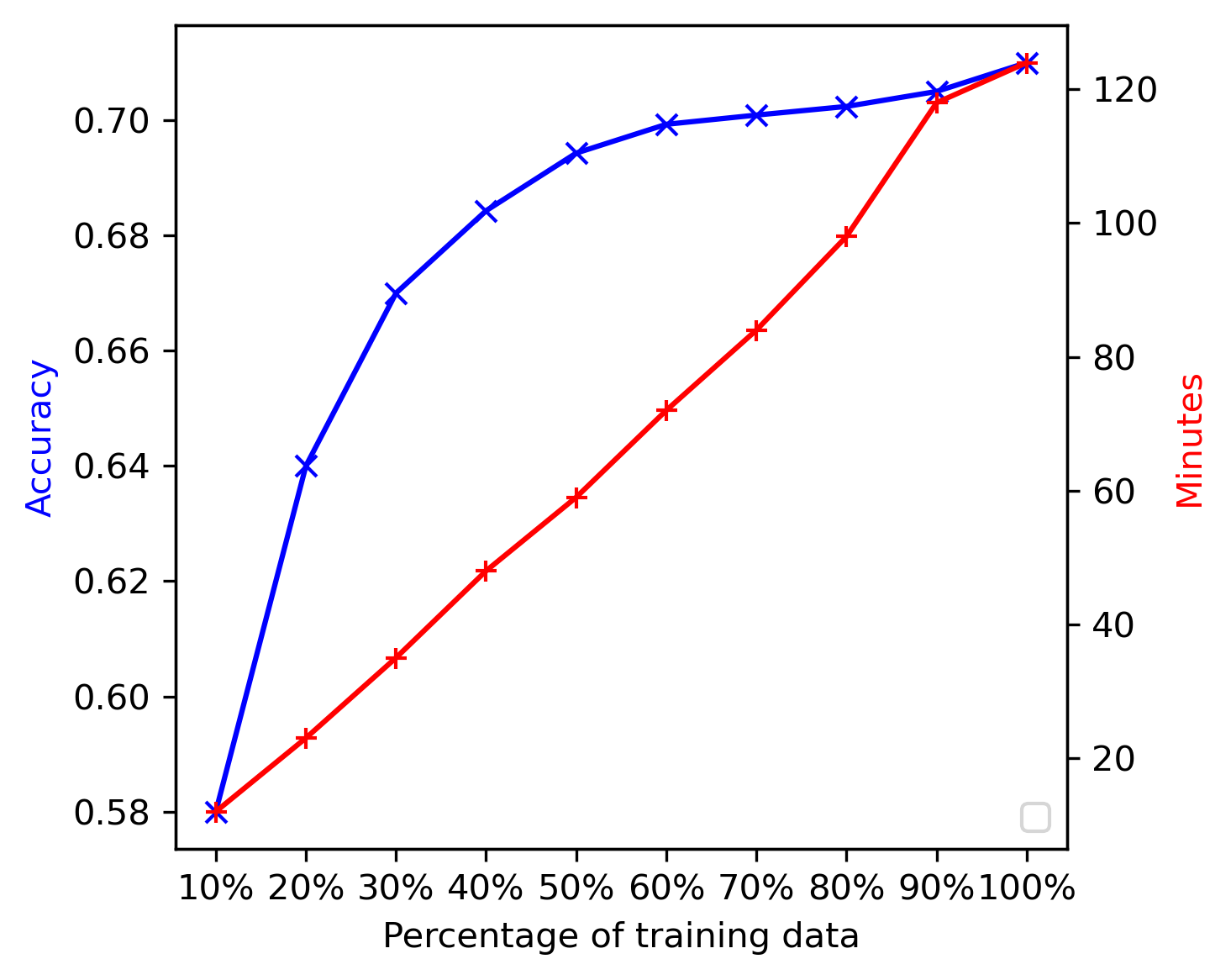}%
  \caption{A neural semantic parser,~\bertlstm~\citep{xu2020autoqa}, is trained on various subsets of the total NLMapV2~\citep{lawrence2018improving} training set of 18,000 examples. The accuracies of the semantic parsers (blue line). The training time for each semantic parser (red line).  \label{fig:nlmap}}
\end{figure}
Data resource limitations occur mainly due to a lack of annotated training data, and sometimes, the neural semantic parser’s incapacity to exploit data.


The primary reason for the lack of data for semantic parsing is that data collection is expensive. Typically, the training data of a semantic parser consists of parallel utterance-LF pairs. First, collecting the utterances is challenging, given that many semantic parsing scenarios are highly task- and domain-specific. For instance, only users who are interested in an insurance company’s product would pose a question regarding it. Therefore, only the customers of this company can provide the utterances needed for training the semantic parsers. Second, annotating the LFs, as opposed to collecting data for machine learning tasks such as text classification and named entity recognition, requires strong domain knowledge of the LFs. Thus, it is typically difficult to scale up the annotation project owing to the impracticality of educating many annotators on the skills necessary to annotate task-specific LFs. Owing to these annotation limitations, collecting semantic parsing data can be costly, cumbersome, and time-consuming. For instance, gathering the Semantic Parsing in Context (SParC) dataset~\citep{yu2019sparc}, which contains only approximately 10,000 utterance-LF pairs, required 1,000 person-hours from 15 computer science students.

As with other neural models, neural semantic parsers are data hungry. A substantial amount of parallel training data, including NL utterances with annotated LFs, is necessary for training a neural semantic parser through supervised training~\citep{jia2016datarecombination}. \fref{fig:nlmap} demonstrates a strong correlation between the amount of training data and the parser performance on the NLMapV2~\citep{haas2016} benchmark. More than 18,000 training examples are required to attain merely an accuracy of 71\% on the test sets of NLMapV2. When data are lacking, semantic parsing is consequently impeded by data hungriness.

\subsection{Limited Computational Resources}
The computational resource limitations are mainly due to the time complexity and memory cost of a semantic parser. Commonly, when adapting it to a new task, the parser is retrained from scratch by using the training data of all previous tasks instead of being fine-tuned only on the new task. Retraining can guarantee parser performance across all observed tasks, whereas fine-tuning suffers from catastrophic forgetting. That is, the model forgets the knowledge about previous tasks after learning a new task. However, the rising number of new tasks raises training costs. As shown in~\fref{fig:nlmap}, the training time of semantic parsers is proportional to the amount of training data. Consequently, the cost of retraining would increase linearly as the amount of training data or the number of tasks increases. Because of the volume of old tasks, frequent model updating through retraining would be time-consuming. One alternative choice to prevent catastrophic forgetting is training one parser per task, which may incur a large memory cost when the number of tasks is large. Due to privacy leakage concerns, there is a need to deploy the semantic parser of the personalized applications (e.g., personalized chatbot) on local devices with low memories, such as mobile phones, which cannot support such a large memory cost. In addition, the small memory might not be able to store all the data of old tasks, which makes the retraining infeasible as well.

In light of these situations, there is a clear need to improve the neural semantic parser in order to maximize parser performance with limited resources. This thesis will shed light on improving semantic parsers in different low-resource conditions.

\section{Limited Resource Conditions}

Many current research efforts aim to improve semantic parsers under resource-constrained conditions with different types of available resources, but many problems remain unsolved. Due to the time constraints of this thesis, issues under three different resource conditions have been investigated. Specifically, this section will categorize two distinct limited data conditions and one limited computational resource condition according to the types of available resources. Moreover, it will identify current research gaps that require attention, from which it will derive the primary research questions.

\subsection{Limited Parallel Data for Semantic Parsing} In the zero/few-shot data condition, it is assumed that there are no, or only a handful of, parallel training examples available for a new task or a new domain. Sometimes, some prior resources are also available, such as lexicons extracted from databases or the training data from a related source task/domain. This scenario frequently occurs when a small business begins operations in a new domain that has limited data resources. However, it is crucial to prototype a semantic parser in the new domain rapidly. Given the available prior resources, recent studies have solved the problem by either automatically generating synthetic training examples using the database lexicons~\citep{xu2020schema2qa,herzig2019don,xu2020autoqa,yin2021ingredients} or transferring knowledge from the source tasks~\citep{rongali2022training,li2021domain}. However, ways to explore prior knowledge efficiently still need to be further explored.

\subsection{Limited Annotation Budget for Multilingual Semantic Parsing} This condition assumes an abundance of monolingual data in the current domain or task. Simultaneously, the semantic parser will generalize to a new language that has no data but will require a limited budget for human annotation. Some studies~\citep{moradshahi2020localizing,sherborne2020bootstrapping} also assume machine translation (MT) systems are available in this condition. This scenario frequently occurs when a multinational corporation desires to launch an international business, which requires using a multilingual semantic parser. Rather than the collection and annotation of utterances in the target language, current methods involve the direct translation of the utterances in the existing semantic parsing datasets into the target languages by either human translators~\citep{li2021mtop} or MT services~\citep{moradshahi2020localizing,sherborne2020bootstrapping}. By using these methods, the expenditure required to collect new utterances and LFs can be avoided. However, human translation is expensive, time-consuming, and laborious, whereas machine-translated data are of low quality.

\subsection{Limited Computational Resources for Semantic Parsing} This condition assumes that the training time and computational memory are constrained. Typically, this occurs when there is a pressing need to deliver the parser for a new task within a short timeframe. However, machine resources for training the parser are at capacity (e.g., training the parser on the mobile devices). Recent studies have used continual learning (CL) to train the parser under this condition. However, most studies have focused on CL for general sequence generation settings and have disregarded customization for the semantic parsing problem.


\section{Research Questions}

Given the limited resource conditions described in the previous section, this thesis will address five research questions. Each research question addresses a problem under a resource condition with different amounts of resources. The first four relate to limited data resources in two distinct data conditions, whereas the fifth relates to limited computational resources.

\textbf{Part I: Limited Parallel Data for Semantic Parsing}

\begin{itemize}
    \item \textbf{RQ1} How can a semantic parser be rapidly adapted to a new task or a new domain when there is no parallel data available but just structured data in a database? (\cref{chap:auto})
    \item \textbf{RQ2} How can a semantic parser be adapted rapidly to a new task or domain when the target task/domain has limited parallel data but a related source task/domain has abundant parallel data?  (\cref{chap:meta})
\end{itemize}
\textbf{Part II: Limited Annotation Budget for Multilingual Semantic Parsing}
\begin{itemize}
    \item \textbf{RQ3} How can a semantic parser be adapted to a new language when the annotation budget for manual translation is limited?  (\cref{chap:al_msp})
    \item \textbf{RQ4} How can a semantic parser be adapted to a new language when the annotation budget for manual translation is limited but an automatic MT service is available for the new language? 
    (\cref{chap:comb_human_auto})
    \end{itemize}
\textbf{Part III: Limited Computational Resources for Semantic Parsing}
\begin{itemize}
    \item \textbf{RQ5} How can a semantic parser be adapted rapidly to a new task, without sacrificing its performance on previous tasks, when the training time and computational memory are limited?  (\cref{chap:continual})
\end{itemize}

\section{Contributions}
Four areas of research are now being proposed to improve the performance of machine learning models under limited data and computational resource conditions: automatic data curation, knowledge transfer, active learning (AL), and CL. The first, automatic data curation generates a sizable amount of synthetic data that can be used to train a model directly without human involvement. In contrast, the second, knowledge transfer techniques, exploit existing knowledge from data-rich tasks in order to enable the parser to adapt quickly to the new task by using only limited training data. AL algorithms select the most representative unlabeled instances to be annotated in order to maximize model performance while incurring the least annotation cost. CL aims to lower computational costs while adapting the machine learning model to sequentially arriving tasks.

In light of these approaches, this thesis intends to use automatic data curation, knowledge transfer methods, AL, and CL to solve the issues regarding semantic parsing in low-resource conditions.

\textbf{Part I: Limited Parallel Data for Semantic Parsing}

\begin{description}
    \item[Automatic Data Curation] To answer RQ1, this thesis proposes a method that automatically generates a large amount of training data in a novel domain from zero training examples, given only the structured data stored in a database in this domain. 
    The approach here follows a data curation procedure similar to that of Overnight~\citep{wang2015overnight}, which firstly generates a large number of non-fluent canonical utterance-LF pairs given a set of seed lexicons extracted from database data, and then paraphrases the canonical utterances into fluent NLs.
    The primary contribution is to evaluate the efficacy of various paraphrasing techniques in automatic data curation procedures and to employ the most effective paraphrasing techniques to maximize data quality. As a result, more than 300,000 training examples could be rapidly collected by exerting minimal human effort. Furthermore, in terms of semantic parsing evaluation metrics, the neural semantic parsers trained on such a dataset can achieve desirable performance. The details of this contribution to the literature are presented in~\cref{chap:auto}.
        \item[Knowledge Transfer] To answer RQ2, this thesis addresses the few-shot learning (FSL) problems in semantic parsing. In the FSL context, there are only a few examples for the new task but a large amount of data for the source task. This thesis investigates whether the few-shot semantic parsing problems can be resolved by transferring information from related tasks. The main contributions are i) a meta-learning technique for pre-training the semantic parser in the source task so that the parser can quickly adapt to the new task; ii) a regularization technique that employs prior knowledge to help the neural model learn the alignment between the NL and LF. The experiments demonstrated that this method improves the performance of the semantic parser in few-shot settings to the point where the parser outperforms other low-resource semantic parsing baselines on three benchmarks. The details of these contributions are provided in~\cref{chap:meta}.
    \end{description}    
\textbf{Part II: Limited Annotation Budget for Multilingual Semantic Parsing}
    \begin{description}
        \item[Active Learning] To answer RQ3, this thesis presents the \textit{first} AL approach that addresses issues while adapting the semantic parsing to a new language and requires a limited annotation budget. The primary contribution is an AL framework that incrementally samples utterances from an existing dataset in a high-resource language for manual translation. Another contribution is an innovative sampling method that, given a limited budget for annotation, samples the most representative examples. The experiments demonstrated that this sampling technique consistently selects samples that are more representative than those selected by other sampling techniques. Using the proposed AL framework and sampling strategy, annotating 32\% of the pool yields a multilingual parser that performs comparably on multilingual benchmarks to the one trained on the full human-translated set. The details of the contribution can be found in~\cref{chap:al_msp}.
        
        \item[Automatic Data Curation, Active Learning] To answer RQ4, this thesis proposes combining the advantages of automatic data curation and AL to boost the performance of multilingual semantic parsing (MSP). The primary contribution is a novel AL framework and a sampling technique to maximize the parser performance in a target language given an annotation budget. Nonetheless, the AL and sampling approach is tailored to a scenario in which automatic MT could be used to translate all the examples. Experiments demonstrated that the proposed AL technique can mitigate the shortcomings of MT effectively and improve the performance in the target language of a multilingual parser trained solely on machine-translated data. By translating only 16\% of all examples, the performance of the multilingual parser would be comparable to that of the parser trained on the full human-translated set. The details of the contribution are discussed in~\cref{chap:comb_human_auto}.

\end{description}
\textbf{Part III: Limited Computational Resources for Semantic Parsing}
\begin{description}
\item[Continual Learning] 
To answer RQ5, this thesis presents a framework for CL in semantic parsing. In addition, it proposes a hybrid method that combines two novel CL methods customized for semantic parsing to address catastrophic forgetting and facilitate knowledge transfer in CL. Experiments indicated that the training time of a semantic parser can be accelerated by up to six times while maintaining performance comparable to an Oracle parser retrained on all seen tasks from scratch on three CL benchmarks. The contribution is detailed in~\cref{chap:continual}.  
\end{description}
\section{Thesis Outline}


This section outlines the organization of this thesis. The key contributions are discussed in Chapters~\ref{chap:auto},~\ref{chap:meta},~\ref{chap:al_msp},~\ref{chap:comb_human_auto} and~\ref{chap:continual}: Chapters~\ref{chap:auto} and~\ref{chap:meta} address the zero/few-shot semantic parsing, Chapters~\ref{chap:al_msp} and~\ref{chap:comb_human_auto} address the problems in MSP because of the limited annotation budgets and Chapters~\ref{chap:continual} presents solutions to the issues in computational resource-constrained semantic parsing.

\paragraph*{Chapter 2: Background}

This chapter provides background information on semantic parsing and a literature review. First, it introduces the fundamentals of semantic parsing, and then, it discusses the current state of low-resource semantic parsing.

\subsection*{Part I: Limited Parallel Data for Semantic Parsing}
\paragraph*{Chapter 3: Automatic Data Curation for Zero-shot Semantic Parsing}

This chapter addresses the zero-shot problem for semantic parsing by automatically generating a large amount of training data in a domain from zero training examples.

\paragraph*{Chapter 4: Knowledge Transfer for Few-shot Semantic Parsing}

This chapter addresses the FSL problems in semantic parsing through a meta-learning approach and a regularization method that utilize prior knowledge from a source task.

\subsection*{Part II: Limited Annotation Budget for Multilingual Semantic Parsing}

\paragraph*{Chapter 5: Active Learning for Multilingual Semantic Parsing}

This chapter presents the \textit{first} AL approach that maximizes the MSP performance in the new language under a limited annotation budget.

\paragraph*{Chapter 6: Machine Translation and Active Learning for Multilingual Semantic Parsing}
This chapter aims to combine the advantages of automatic data curation and AL in order to maximize the MSP performance in the new language given a limited annotation budget and a MT service.

\subsection*{Part III: Limited Computational Resources for Semantic Parsing}
\paragraph*{Chapter 7: Continual Learning for Semantic Parsing}

This chapter proposes a CL method that improves the semantic parser that has limited computational resources.

\paragraph*{Chapter 8: Summary}

This chapter concludes the thesis and discusses the future directions of semantic parsing that has limited resources.


\chapter{Background}
\label{chap:back}
\epigraph{The science of today is the technology of tomorrow.}{\textit{Edward Teller}}
\fbox
{
\begin{minipage}{0.96\textwidth}
     This chapter is based on\\
     Z. Li, L. Qu, G. Haffari. ``Context Dependent Semantic Parsing:A Survey'' The 28th International Conference on Computational Linguistics (COLING). 2020.
 
\end{minipage}
}

This chapter mainly focuses on providing the necessary context for readers to comprehend the studies conducted in this thesis.
~\sref{sec:prelim} briefly describes the fundamentals of semantic parsing, including the basic semantic parsing framework, two semantic parsing scenarios, and the corresponding models used in these scenarios.
~\sref{sec:limited} examines the current state of studies that tackle the low-resource problems in semantic parsing, including automatic data curation, knowledge transfer, AL, and CL.

\section{Preliminaries in Semantic Parsing}
\label{sec:prelim}
This section will first describe the semantic parsing framework, along with its problem definition, salient components, and evaluation approaches. Then, this section presents two semantic parsing scenarios, namely the monolingual and multilingual semantic parsing. 

\subsection{Semantic Parsing Framework}
\label{sec:parser_framework}
\paragraph*{Problem Setting.} As in~\fref{fig:sp_diagram}, semantic parsing aims to learn a mapping function $\pi_{\bm{\theta}} : \mathcal{X} \rightarrow \mathcal{Y}$, which translates an NL utterance $\vx \in \mathcal{X}$ into a machine-readable semantic representation $\vy \in \mathcal{Y}$, which can be referred to as an LF\footnote{In this thesis, all the semantic representations are referred to as LFs by default.}, a MR or a program~\citep{kamath2018survey} interchangeably. Specific types of LFs $\vy$ are designed to be executable within a programming environment (e.g., databases and knowledge graphs) to yield a result $\vz$, namely denotation. The LF structure takes the form of either a tree or a graph, depending on its underlying formal language. The LF languages are categorized into three types of formalism: logic-based, graph-based, and programming languages~\citep{kamath2018survey}. Some semantic parsers explicitly apply a production grammar to yield LFs from utterances. This grammar consists of a set of production rules, which define a list of candidate derivations for each NL utterance. Each derivation deterministically produces a grammatically valid LF. 
\paragraph*{Logical Forms.} The three types of LF languages are logic-based, graph-based and programming languages~\citep{kamath2018survey}. A logic-based language such as Lambda DCS~\citep{liang2013lambda} utilizes quantifiers, entities, or predicates with first-order logic. A graph-based language such as Abstract Meaning Representation (AMR)~\citep{banarescu2013abstract} represents the semantics of the NL as a semantic graph, in which nodes represent entities and edges represent their relationships. Various programming languages are used, such as Java, Python, and SQL. Typically, an LF takes the form of a tree or a graph; the subtrees and sub-graphs are termed \textit{compounds}~\citep{keysers2019measuring} or \textit{code idioms}~\citep{iyer2019learningIdiomsSP}, and the nodes are termed \textit{atoms}~\citep{keysers2019measuring}. 

\paragraph*{Grammar.} In semantic parsing, grammar rules constrain the space of the valid LFs given the utterances $\vx$ and the environment. A typical type of grammar is context-free grammar (CFG), in which each production rule in a grammar set is defined as:
\begin{equation}
    \alpha \rightarrow \beta
\end{equation}
where $\alpha$ is a non-terminal symbol, $\beta$ is a string of terminal or non-terminal symbols, and the arrow $\rightarrow$ means string $\beta$ is derived from $\alpha$. Another typical grammar formalism is combinatory categorial grammar (CCG), which is an extension of context-free grammar. CCG is usually used to parse the NL utterances into valid $\lambda$-calculus expressions~\citep{zettlemoyer2005learning}. Repetitively deriving the grammar rules could yield a grammatically valid derivation tree, $\vd \in G(\vx)$, where $G(\vx)$ denotes the set of all candidate derivations based on the input $\vx$.

\paragraph*{Environment.} Typically, the environment consists of relational databases or knowledge bases (KBs) containing structured data that the LFs can query. For example, a KB includes a knowledge graph constructed by a collection of facts where each fact is represented in a triple format, i.e., \textit{(subject entity, predicate, object entity)}, showing that there is a relation between two entities. Each triple represents a piece of world knowledge. For instance, `\textit{VIC is rainy}' is represented as `\textit{(VIC, weather, rainy)}'. The environment can also consist of multimedia content, such as images that are parsed into structured knowledge graphs~\citep{hudson2018gqa}. 
 
\paragraph*{Executor.} The executor executes the LF against the environment to obtain the denotation. For example, an executor could execute a SQL query on the relational database to extract a column value. 

\paragraph*{Parsers.} The parser is a parametric model $P_{\bm{\theta}}(\vy|\vx)$ or $P_{\bm{\theta}}(\vy|\vx,C)$ that assigns the probabilities to the possible LFs conditioned on the input utterance and, optionally, a context depending on whether the parser explicitly models the context. There are two types of contexts, as defined by~\citet{li2020context}, namely, local and global context. The \textit{local} context for an utterance is the text and multimedia content surrounding it, which is meaningful only for this utterance. For example, the local context for each utterance can be its history utterances or LFs in a dialogue. In contrast, the \textit{global} context is the information accessible to more than one utterance, including databases and external text corpora, images or class environment~\citep{iyer2018mapping}. The content of local context varies for each NL utterance while the global context is always static. Context provides additional information to resolve ambiguity and vagueness in current utterances. In addition, some types of parsers, namely multilingual semantic parsers, could parse utterances in different languages. Such parsers will be introduced at \sref{sec2:msp} in detail. 


\paragraph*{Learning.} Semantic parsing has various learning settings, including transfer learning, meta-learning, CL and AL, which will be introduced in later sections. This section provides a brief overview of the most common learning method, supervised learning. In a fully supervised setting for semantic parsing, the objective of learning is to minimize the negative conditional log-likelihood over the training set, which consists of utterance-LF pairs, $\mathcal{D}=\{(\vx_i, \vy_i)\}^N_{i=1}$:
\begin{equation}
\label{eq:sup_obj}
    \mathcal{L}_{\mathcal{D}}(\bm{\theta}) = \sum_{(\vx,\vy) \in \mathcal{D}} -\log P_{\bm{\theta}}(\vy|\vx) + \Omega
\end{equation}
where $\bm{\theta}$ represents the model parameters, and $\Omega$ is the regularization term. A common regularization term is the L2 norm, $\Omega = \lambda_1 ||\bm{\theta}||_{2}$, but it can also be in other forms. The gradient descent algorithm is often adopted to determine the optimal model parameters. There will be numerous iterations of learning before model convergence is achieved. At each step $i$ of the learning, the parameters $\bm{\theta}$ are updated as:
\begin{equation}
    \bm{\theta}_{i+1} = \bm{\theta}_{i} - \lambda_2 \nabla_{\bm{\theta}}\mathcal{L}_{\mathcal{D}}(\bm{\theta})
\end{equation}
where $\lambda_2$ is the learning rate that determines the size of each learning step.

\paragraph*{Evaluation.} 
In semantic parsing, \textit{exact match accuracy} is the most commonly used evaluation metric. With \textit{exact match accuracy}, the parsing results are considered correct only when the output LF/denotations exactly match the string of the ground truth LF/denotations. One flaw of the evaluation metric is that some types of LFs (e.g. SQL) do not hold order constraints across clauses. \citet{yu2018spider} proposed a metric \textit{set match accuracy} to evaluate the semantic parsing performance over SQL, which treats each SQL statement as a set of clauses and ignores their order.

Currently, there is a major research problem, compositional generalization. NLs can be highly compositional. For example, `What is the weather like today in the state closest to VIC during the period, not between 3 and 6 p.m.?' can be decomposed into several simple phrases, such as `what is', `weather like', `VIC', `in the state closest to', and `today'. The phrases can be composed with the phrases in other utterances, such as `where to buy', `pet food', and `NSW' in the utterance `Where to buy pet food in NSW' to form new utterances such as `where to buy pet food in the state closest to NSW'. The new utterances will be aligned with LF with new compositions. In order to evaluate whether the parser could be generalized to unseen utterance and LF compositions,~\citet{finegan2018improving}, and~\citet{keysers2019measuring} proposed two novel data splitting methods, \textit{Query Split} and \textit{Maximal Compound Divergence}, respectively, in comparison to the traditional random splitting of the dataset. The \textit{Query Split} split the dataset, so there are no template overlaps between training and test sets. The templates are LFs with entities and attribute values anonymized. For example, ``\textit{ask\_weather(state\_id(`VIC'), date(`today'), var)}'' will be ``\textit{ask\_weather(state\_id(var), date(var), var)}'' after anonymization. \textit{Maximal Compound Divergence} maximizes the difference of compounds among the training and test sets, where the \textit{compounds} are the subtrees of the LFs trees. On using these two approaches to split the data, the training and test sets share the same set of simple LF segments but include very different utterances and LF compositions. 


\subsection{Monolingual Semantic Parser}
\label{sec2:no_context}
In current semantic parsing research, the monolingual semantic parsers are the ones most frequently encountered. Within this section, such parsers are referred to as semantic parsers for simplicity. Formally, given an utterance $\vx \in \mathcal{X}$ and its paired LF $\vy \in \mathcal{Y}$, a semantic parsing model can form a \textit{conditional} distribution $P(\vy | \vx)$ or $P(\vy | \vx, C)$ depending on whether the parser models the context $C$.
The model learning can be supervised by either utterance-LF pairs or merely utterance-denotation pairs. If only denotations are available, a widely used approach~\citep{kamath2018survey} is to marginalize over all possible LFs for a denotation $\vz$, which leads to a \textit{marginal} distribution $P(\vz | \vx) = \sum_{\vy} P(\vz, \vy| \vx)$. A parsing algorithm aims to find the optimal LF in the combinatorially large search space.
The existing models are coarsely categorized into symbolic, neural and neural–symbolic approaches according to the category of machine learning methodology and whether any production grammars are explicitly used in the models.
\subsubsection*{Symbolic Approaches}
\label{sec:pipe}
A symbolic semantic parser employs production grammars to generate candidate derivations and find the most probable one via a scoring model. The scoring model is a statistical or machine learning model. 
 Each derivation is represented by hand-crafted features extracted from utterances or partial LFs. Let $\Phi(\vx, \vd)$ denote the features of a pair of utterance and derivation, and $G(\vx)$ be the set of candidate derivations based on $\vx$. A widely used scoring model is the log-linear model~\citep{zettlemoyer2012graphPCCG,kamath2018survey}. 
\begin{equation}
    P_{\bm{\theta}}( \vd | \vx) = \frac{\exp(\bm{\theta}\Phi(\vx, \vd))}{\sum_{\vd' \in G(\vx)} \exp(\bm{\theta}\Phi(\vx, \vd'))}
\end{equation}
where $\bm{\theta}$ denotes the model parameters. If only utterance-denotation pairs are provided at training time, a model marginalizes over all possible derivations yielding the same denotations by $P(\vz | \vx) = \sum_{\vd} P(\vz, \vd| \vx)$~\citep{krishnamurthy2012weakly,liang2016learning}. Those corresponding parsers further differentiate between graph-based parsers~\citep{flanigan2014graphSP,zettlemoyer2012graphPCCG} and shift-reduce parsers~\citep{zhao2014shiftReduceSP} due to the adopted parsing algorithms and the ways to generate derivations. From a machine learning perspective, these approaches are also linked to a structured prediction problem.


\subsubsection*{Neural Approaches}
\label{para:nerual}

\begin{figure}
    \centering
    \includegraphics[width=1\textwidth]{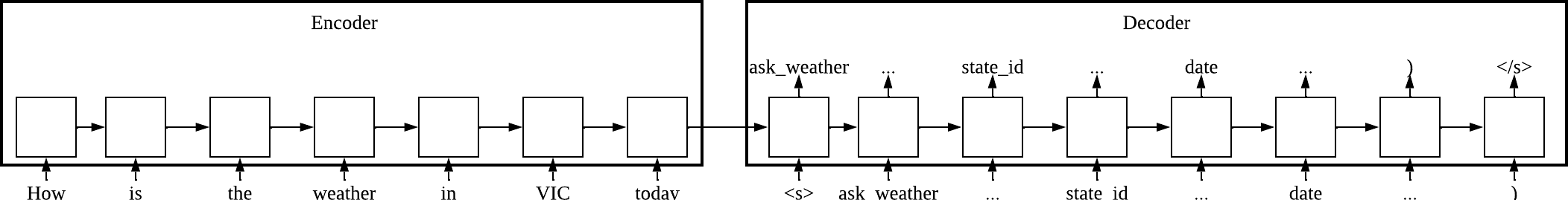}
    \caption{A common Seq2Seq semantic parsing model that generates the token sequence in LF for the utterance `How is the weather in VIC today'. Several token decoding steps have been omitted owing to space constraints.}
    \label{fig:seq2seq_figure}
\end{figure}

Neural approaches apply neural networks to translate NL utterances into LFs without using production grammars. These approaches formulate semantic parsing as a MT problem by viewing NL as the source language and the formal language of LFs as the target language. 

Most research in this category have adopted \Seq~\citep{sutskever2014sequence} as the backbone architecture, which consists of an encoder and a decoder. The NL utterance $\vx$ and linearized LF $\vy$ are regarded as two sequences of tokens, $\vx = \{x_{0},...,x_{|\vx|}\}$ and $\vy = \{y_{0},...,y_{|\vy|}\}$, respectively. The encoder projects NL utterances into hidden representations, whereas the decoder generates linearized LFs sequentially. 
 Both encoders and decoders employ either recurrent neural networks (RNNs)~\citep{goodfellow2016deepbook} or Transformers~\citep{vaswani2017attention}. Note that, these methods do not apply any production grammars to filter out syntactically invalid LFs.

The variants of the \Seq-based models also explore the structural information of LFs. Seq2Tree \citep{dong2016language} utilizes a tree-structured RNN as the decoder, which constrains generated LFs to take syntactically valid tree structures. The Coarse2Fine model~\citep{dong2018coarse} adopts a two-stage generation for the task. In the first stage, a \Seq model is applied to generate LF templates, which replace entities in LFs by slot variables for a high-level generalization. In the second stage, another \Seq model is applied to fill the slot variables with the corresponding entities. The CodeT5~\citep{wang2021codet5} model adopts Transformers instead of RNNs for its encoder and decoder. The main benefit of using Transformer is that the semantic parsers can be trained on large-scale training data. For example, CodeT5 is trained on more than 8.35 million functions across multiple programming languages, whereas common semantic parsing models that utilize RNNs are typically trained on just thousands of examples.

However, certain neural semantic parsers do not use \Seq architecture. For example, Codex~\citep{chen2021evaluating} uses only a Transformer decoder trained on billions of training data scraped from GitHub\footnote{https://github.com/}. As a result, it can work with most commonly encountered LFs, such as various programming languages, including Python, Java, Prolog, and so on. Such a model obtains the LF by feeding the input NL utterance to the decoder and inferring the subsequent tokens to the NL utterance as the LF tokens. Some prior studies on Codex~\citep{shin2021few} also concatenated the \textit{prompt} with the utterance to be parsed as the input to Codex. The prompt is typically a token string consisting of example NL-LF pairs that demonstrate how to perform parsing for Codex, thereby improving parsing performance. 

Some parsers would incorporate information from the local and global context to resolve ambiguity in current utterances. For example, \citet{suhr2018learning,suhr2018situated,zhang2019editing,he2019pointer,pascanu2013difficultyRNN,khandelwal2018sharpRNNContext} built context-aware encoders to encode historical utterances or LFs into neural representations. \citet{dong2018coarse,lin2020bridging} used encoders to encode the linearized tokens in the structured schemas of the database or knowledgebase into representations. Then, the decoders take the contextual representations with the ones from current utterances as the input. The parsers, such as Codex, could utilize context as well by including the history utterance-LF pairs or table schema in the prompt.


\paragraph*{Attentional Seq2Seq.}
Owing to the significance of \Seq to the neural semantic parser, we provide a comprehensive explanation of a common attentional Seq2Seq architecture, adopted as the baselines in Chapters~\ref{chap:auto},~\ref{chap:meta}, and~\ref{chap:continual}.
In this example, the encoder and decoder for this Seq2Seq are RNNs, although other networks, such as Transformers, can also be applicable. Such an architecture would learn the soft alignment between input words and the LF tokens. The \Seq estimates the token probabilities as follows:

\begin{equation}
    P(\vy | \vx ) = \prod_{t = 1}^{|\vy|} P(y_t | \vy_{<t},\vx)
\end{equation}

\textit{Word Embedding} The inputs for the \Seq are the discrete words and LF tokens, which will be converted into distributed word embeddings by the embedding tables. For semantic parsing, there are two learned embedding tables, one for input words $\mathbf{C}_{x} \in \mathbb{R}^{H_{x}\times|\mathcal{V}_{x}|} $, and the other for output LF tokens $\mathbf{C}_{y} \in \mathbb{R}^{H_{y}\times|\mathcal{V}_{y}|} $, where $H_{x}$ and $H_{y}$ are the dimension sizes of the representations of the words and LF tokens, respectively, while $\mathcal{V}_{x}$ and $\mathcal{V}_{xy}$ are the respective vocabularies for the words and LF tokens. The word embeddings are typically pre-trained on a large corpus using an unsupervised method, whereas the LF token embeddings are randomly initialized.

\textit{Encoder} A bidirectional RNN maps a sequence of $|\vx|$ words into a sequence of contextual word representations $\{\mathbf{e}\}_{i=1}^{|\vx|}$, where $\mathbf{e}_{i}$ is obtained by concatenating the RNN outputs from two directions, $\mathbf{e}_{i} = [\overrightarrow{\text{RNN}}(\mathbf{C}_{x}[x_{i}], \overrightarrow{\mathbf{h}}_{i-1});\overleftarrow{\text{RNN}}(\mathbf{C}_{x}[x_{i}], \overleftarrow{\mathbf{h}}_{i+1})]$.

\textit{Decoder} The decoder applies an RNN to generate token sequences.

At time $0$, the initial hidden state for the RNN is the last state from the encoder $\mathbf{e}_{|\vx|}$.

At time $t$, the RNN produces a hidden state $\mathbf{h}^d_{t}$ by $\mathbf{h}^d_t = \text{RNN}(\mu_{t}, \mathbf{h}^d_{t-1})$, where $\mu_{t}$ is a concatenation of the embedding of the LF token estimated at time $t$, which is $\mathbf{c}_{y_{t}}=\mathbf{C}_{y}[y_{t}]$, and a representation $\mathbf{h}_{t-1}$ estimated at time $t-1$. 

As a common practice, $\mathbf{h}^d_{t}$ is concatenated with an attended representation $\mathbf{h}_t^a$ over encoder hidden states to yield $\mathbf{h}_t$, with
\begin{equation}
    \mathbf{h}_t = \mathbf{W} \begin{bmatrix}
    \mathbf{h}_t^d\\
    \mathbf{h}_t^a
    \end{bmatrix}
\end{equation}
where $\mathbf{W}$ is a weight matrix and $\mathbf{h}^a_t$ is created by soft attention,
\begin{equation}
    \mathbf{h}^a_t = \sum_{i=1}^{|\vx|} P(\mathbf{e}_i| \mathbf{h}^d_t)\mathbf{e}_i
\end{equation}
The dot product attention~\citep{luong2015effective} computes the normalized attention scores $P(\mathbf{e}_i| \mathbf{h}^d_t)$ with

\begin{equation}
\label{eq:attention_score}
    P(\mathbf{e}_i| \mathbf{h}^d_t) = \frac{\exp(\mathbf{e}_i^{\intercal} \mathbf{h}^{d}_t)}{\sum^{|\vx|}_{i=1}\exp(\mathbf{e}_{i}^{\intercal} \mathbf{h}^{d}_t)}
\end{equation}

Given $\mathbf{h}_t$, the probability of a token is estimated by:
\begin{equation}
\label{eq:action_prob}
P(y_t | \vy_{<t},\vx) = \frac{\exp(\mathbf{c}_{y_t}^{\intercal} \mathbf{h}_t)}{\sum_{y_{t}' \in \mathcal{V}_{y_{t}}} \exp(\mathbf{c}_{y_{t}'}^{\intercal} \mathbf{h}_t)}
\end{equation}
where $\mathbf{c}_{y}$ denotes the embedding of token $y$, and $\mathcal{V}_{y_{t}}$ denotes the set of applicable LF tokens at time $t$. 

\textit{Supervised Attention}~\citep{rabinovich2017semanticParsingSupervisedAttention,yin2018tranx} is sometimes applied to facilitate learning attention weights when training the semantic parsing model, whereby it will assign a prior attention score $P'(\mathbf{e}_i| \mathbf{h}^d_t)$ to regularize the learning of $P(\mathbf{e}_i| \mathbf{h}^d_t)$. The regularization term in the training loss function is:
\begin{equation}
\label{eq:supervised_att}
    \Omega = ||P'(\mathbf{e}_i| \mathbf{h}^d_t)-P(\mathbf{e}_i| \mathbf{h}^d_t)||_{2}^2
\end{equation}
where $P'(\mathbf{e}_i| \mathbf{h}^d_t)$ is either $0$ or $1$ given whether the string of the LF token at time $t$ matches the word at position $i$. Supervised attention is applicable when the strings of LF tokens have semantic meanings. For instance, the LF predicate `\textit{ask\_weather}' directly conveys the meaning of `asking about the weather', which can be aligned with the tokens `weather' in the NL utterance. Otherwise, it is necessary to manually construct a lexicon in order to map LF tokens to meaningful NL words, which requires a substantial amount of manual labor.

\subsubsection*{Neural-Symbolic Approaches}

\begin{figure}
    \centering
    \includegraphics[width=1\textwidth]{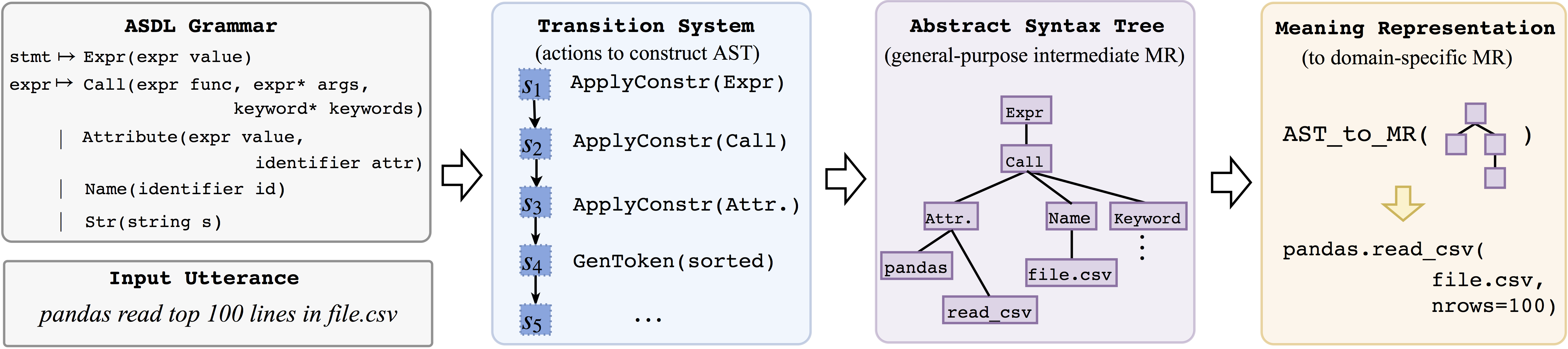}
    \caption{The parsing process of TranX~\citep{yin2018tranx}. An utterance is firstly converted into an action sequence, which is later mapped into an LF/MR. They use a grammar described by Abstract Syntax Description Language (ASDL)~\citep{wang1997zephyr} instead of the common CFG to constrain the decoding process. Figure obtained from~\citet{yin2018tranx}.}
    \label{fig:tranx_exp}
\end{figure}

In order to ensure the generated LFs are syntactically valid without compromising the generalization power of neural networks, neural-symbolic approaches fuse both symbolic and neural approaches by applying production grammars to the generated LFs; then the derivations are scored by neural networks. 


The majority of these methods linearize derivations such that they are able to leverage \Seq~\citep{liang2016neural,yin2018tranx,guo2019towards}. At each time step, the decoder of these methods emits either a parse action or a production rule. At the end of the decoding steps, the sequence of actions $\va = \{a_{0},...,a_{|\va|}\}$ can be executed to obtain a abstract syntax tree (AST), $\tau_{\vy} = \text{Action\_to\_AST}(\va)$. A user-defined function can further deterministically map the tree $\tau_{\vy}$ into a grammatically valid LF or MR, $\vy = \text{AST\_to\_LF/MR}(\tau_{\vy})$. 
\fref{fig:tranx_exp} depicts the entire process of parsing an utterance by a neural-symbolic semantic parser, TranX~\citep{yin2018tranx}. 

A neural-symbolic semantic parser produces derivations by varying grammars. NSM \citep{liang2016neural} uses a subset of Lisp syntax. TranX~\citep{yin2018tranx} defines the grammars in Abstract Syntax Description Language, while IRNet \citep{guo2019towards} considers the context-free grammar of a language called SemQL. The ProtoParser~\citep{li2021few} automatically extracts production rules from the training set and uses the rules to constrain the decoding process. The neural transition system consists of an encoder and a decoder for estimating action probabilities:
\begin{small}
\begin{equation}
    P(\mathbf{a} | \vx ) = \prod_{t = 1}^{|\mathbf{a}|} P(a_t | \va_{<t},\vx)
\end{equation}
\end{small}

In addition, some neural-symbolic approaches adopt neural architectures other than \Seq. One such example is that proposed by~\citet{andreas2016learning}, which adopts a dynamic neural module network to generate LFs.

To utilize local context, neural-symbolic methods normally take the same methods as the neural approaches to encode the contextual information. What differentiate them is the neural-symbolic could handle local context by i) designing specific actions, and ii) utilizing symbolic context representations. For instance, the context specific actions proposed in~\citet{iyyer2017search,sun2019knowledge,liu2020FarAwayContextModelingSP} adopt \textit{copy} mechanism to reuse the previous LFs. CAMP~\citep{sun2019knowledge} includes three actions to copy three different SQL clauses from precedent queries. Dialog2Action~\citep{guo2018dialog} incorporates a dialogue memory, which maintains symbolic context representations of entities, predicates and action subsequences from a dialogue history.

Schema linking is an increasingly popular neural-symbolic technique for text-to-SQL tasks that utilize global context. It is an auxiliary task that learns the alignment between table columns of a database schema and their mentions in the utterances. Then, the neural-symbolic methods, such as IRNet~\citep{guo2019towards} and RAT-SQL~\citep{wang2020rat}, devise specific actions to determine which columns should be selected. The analysis by~\citet{lei2020re} shows that these parsers have good generalization abilities across tasks or domains, which includes different table schemas, due to the fact that schema linking can aid in learning good column representations.

\subsection{Multilingual Semantic Parser}
\label{sec2:msp}

\begin{figure}
    \centering
    \includegraphics[width=1\textwidth]{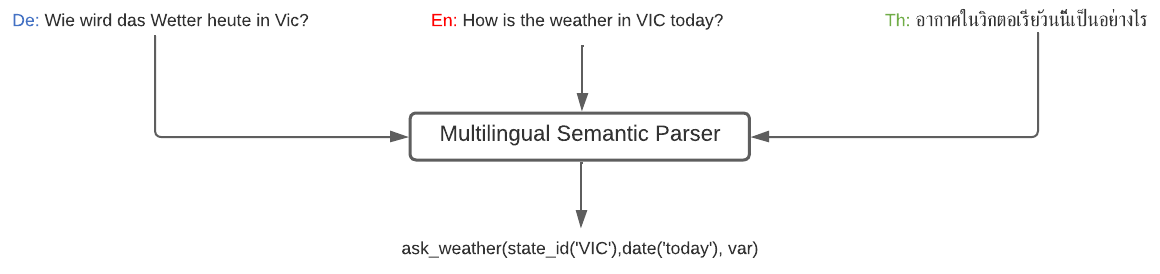}
    \caption{A multilingual semantic parser is capable of correctly parsing an utterance in multiple languages (e.g. English, Thai, German) into an LF.}
    \label{fig:multilingual}
\end{figure}

MSP is an natural extension of monolingual semantic parsing. As in~\fref{fig:multilingual}, an MSP model is a parametric model $P_{\bm{\theta}}(\vy|\vx)$ that assigns the probability to an LF $\vy \in \mathcal{Y}$ given the NL utterance $\vx \in \mathcal{X}$, where the input space $\mathcal{X} = \bigcup_{l \in L} \mathcal{X}_l$ includes utterances in different languages $L$.   
MSP is now widely used in the real world. For example, numerous commercial virtual assistants now serve customers from diverse cultural and linguistic backgrounds. In 2021, Amazon Alexa supported eight languages, whereas Google Assistant supported 12\footnote{https://summalinguae.com/language-technology/language-support-voice-assistants-compared/}. In such applications, multilingual semantic parsers can be useful NLU components. 

Moreover, a multilingual semantic parser is a cost-effective solution for applications whose users have diverse linguistic backgrounds, for three main reasons. First, many languages share similar linguistic phenomena. A multilingual parser could utilize supervision from related languages to improve the performance of parsing across all languages~\citep{ammar-etal-2016-many}. Second, code-switching is a challenging issue for most monolingual semantic parsers~\citep{duong2017multilingual}. A multilingual semantic parser can effectively handle situations in which a single sentence contains terms in different languages. Third, deploying one model per language is burdensome in real-world systems because it would require maintaining an enormous number of models, and this issue can be addressed by using a multilingual semantic parser.


\subsubsection*{Symbolic Approaches}
Hybrid Tree~\citep{jie2014multilingual} is the only symbolic approach that employs hand-crafted multilingual features such as a bigram or unigram for a log-linear model to predict the optimal semantic tree of an LF given an utterance.
\subsubsection*{Neural Approaches} The majority of multilingual parsers are currently neural architectures that focus on how to parse multilingual utterances with the \Seq. \citet{susanto2017neural} proposed a \Seq architecture that employs a separate RNN for each language. This strategy enables the parser to learn the mixing knowledge of several languages and simultaneously parse multiple utterances in different languages. \citet{duong2017multilingual} trained a \Seq model directly on a multilingual dataset using a single RNN encoder. The performance of multilingual semantic parsers has also been enhanced following the development of large multilingual pre-trained language models, which are pre-trained on large-scale multilingual corpora. For example,~\citet{li2021mtop} employed mBART~\citep{liu2020multilingual} as the multilingual semantic parser. BERT-LSTM~\citep{moradshahi2020localizing} as well as the multilingual parsers proposed by~\citet{xia2021multilingual} and \citet{zhu2020don} all employ the mBERT~\citep{devlin2018bert} as the encoders, whereas BERT-LSTM uses an LSTM, and the parsers of \citet{xia2021multilingual} and \citet{zhu2020don} use a randomly initialized Transformer as the decoders, respectively. Currently, the multilingual semantic parsers that apply pre-trained language models have achieved SOTA performance on different benchmarks~\citep{xia2021multilingual,li2021mtop}.
\subsubsection*{Neural-Symbolic Approaches} 
\citet{zou2018learning} employed the same model architecture as Hybird Tree~\citep{jie2014multilingual}, with the difference that their model learns low-dimensional distributed cross-lingual embeddings of LF nodes, which are considered the features of Hybrid Tree.

\section{Semantic Parsing with Limited Resources}
\label{sec:limited}
As noted in~\cref{intro}, constraints on data availability and computational resources significantly impact the performance of semantic parsers for several reasons. Therefore, different lines of prior studies are proposed to tackle the two issues. This section introduces works in automatic data curation, knowledge transfer, and AL that address the problems of limited data, as well as works in CL that address the problems of limited computational resources. 

\subsection{Automatic Data Curation for Semantic Parsing}
Automatic data curation solves data scarcity by generating training examples directly. This section introduces the two types of automatic data curation approaches categorized by whether the information from the existing datasets is utilized during the data curation procedure.
\subsubsection*{Data Curation from Scratch} 
\label{sec:cura_scratch}
When building a machine learning system, often, annotated data are unavailable. It is proposed that in such situations, the automatic data curation method \textit{Overnight}~\citep{wang2015overnight} be used to rapidly collect adequate training data in a new task/domain starting with zero training examples. Overnight and its extended works could significantly reduce human efforts while collecting data from scratch. In general, this line of methods~\citep{wang2015overnight, xu2020schema2qa,herzig2019don,xu2020autoqa,yin2021ingredients} includes two steps:
\begin{itemize}
    \item 
    
    \textbf{Step1:} Given a database schema or a set of seed lexicons, which specify the alignments between canonical NL phrases and the predicate or entities in LFs, human experts write a set of synchronous context-free grammar (SCFG) rules or canonical templates, which can be used to generate a large number of clunky canonical utterance-LF pairs automatically.
    The term ``clunky'' means the generated utterances are less fluent than NL utterances, while human crowd-workers with no technical background on the target LFs can still easily understand the semantics of utterances.

    \item \textbf{Step2:} The clunky canonical utterances are paraphrased into fluent NL utterances through different paraphrasing methods.
\end{itemize}

\begin{figure}
    \centering
    \includegraphics[width=0.9\textwidth]{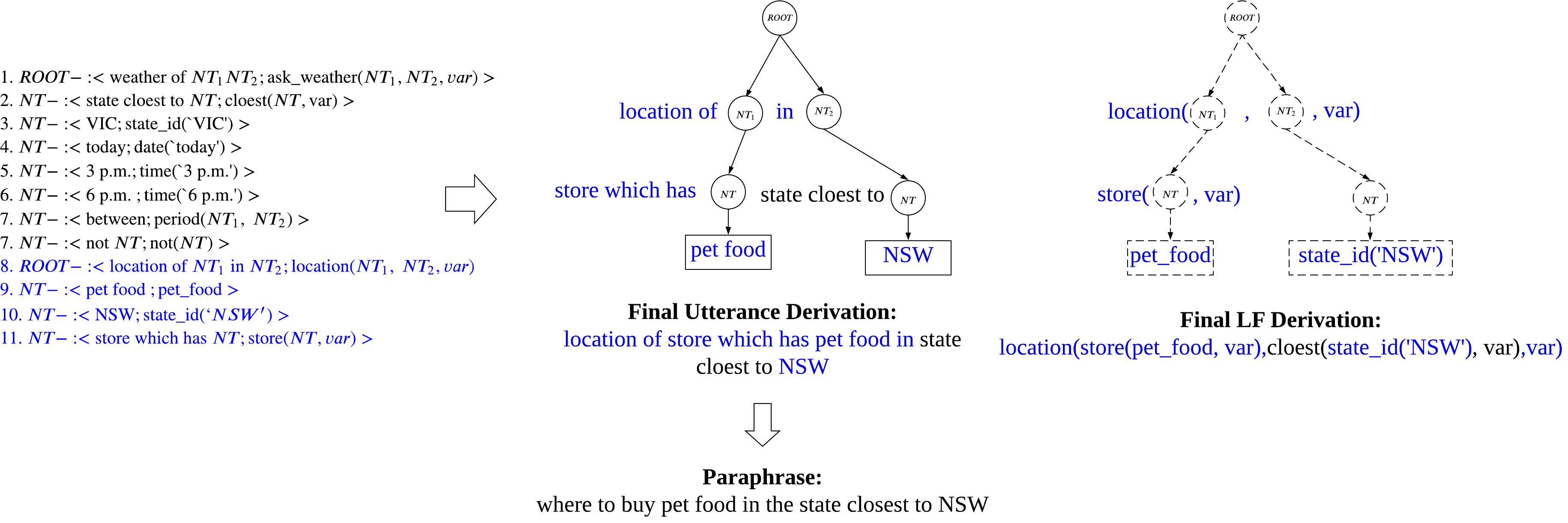}
    \caption{The derivation process using the SCFG rules and the paraphrase for the generated clunky canonical utterance. The SCFG rules are extracted from two utterance-LF pairs where the utterances are `What is the weather like today in the state closest to VIC during the period, not between 3 and 6 p.m.?' and `Where to buy pet food in NSW.' The phrases in the first utterance are colored black, while the ones in the second are colored blue.}
    \label{fig:scfg}
\end{figure}

These approaches reduce the annotation cost by minimizing the expert knowledge of the LFs required by the annotators. In the framework of Overnight, only a few experts are required to write the SCFG rules. In contrast, the traditional data collection approaches require all the annotators to be experts; thus, it is difficult to scale up the dataset size.

The two steps involve two important concepts, SCFG and paraphrasing, which are explained in detail next.

\paragraph*{SCFG.} It is among the most important concepts in automatic data generation. The idea of SCFG originated from phrase-based MT~\citep{chiang2007hierarchical}. Using SCFG, MT models can generate bilingual sentences simultaneously. In semantic parsing, SCFG rules comprise two context-free grammar rules, one to describe the structures of utterances, and the other to specify the LF structures. Formally, a SCFG rule is defined as:
\begin{equation}
    NT \rightarrow (\alpha, \beta, \sim)
\end{equation}
where $NT$ is the symbol of non-terminal, $\alpha$ and $\beta$ are strings that include terminals and non-terminals, $\sim$ indicates the one-to-one alignment between non-terminals in $\alpha$ and the non-terminals in $\beta$, and $\rightarrow$ indicates that the two non-terminal $NT$s in $\alpha$ and $\beta$ would be replaced with the strings $\alpha$ and $\beta$, respectively, during the derivations. As in Fig.~\ref{fig:scfg},
sampling the SCFG rules and recursively applying them to the non-terminals, including $ROOT$ or $NT$, could yield clunky utterance-LF pairs with new utterance and LF compositions at the end of the derivations.


\paragraph*{Paraphrasing Techniques.} SCFG can generate only one clunky utterance paired with each LF. However, due to the rich linguistic complexity, usually, multiple utterances can be mapped to one LF. In addition, clunky utterances are not fluent natural utterances. Therefore, the parsers trained on the clunky utterance-LF pairs would not achieve optimal parsing performance on the test sets with natural utterances. To solve this issue, multiple approaches have been proposed to paraphrase each clunky utterance into a set of natural utterances. Overnight~\citep{wang2015overnight} and Schema2QA~\citep{xu2020schema2qa} hire human crowd-workers from crowd-sourcing platforms to manually write paraphrases. Though this method avoids the most expensive step of teaching workers about target LFs, manually writing paraphrases is still costly and time-consuming. Thus, AutoQA~\citep{xu2020autoqa,yin2021ingredients} automatically paraphrase the clunky utterances using a paraphrasing model, a pre-trained \Seq language model, BART~\citep{lewis2020bart}, fine-tuned on the paraphrasing dataset. The quality of machine-generated paraphrases is typically inferior to that of human-written ones, but these cost much less than the latter.
\citet{herzig2019dontParaphraseDetect} found a mismatch between the induced distribution of SCFG-generated examples and the true distribution. Thus, they assumed to access an existing utterance pool, where the utterances are collected from real users. Then they used paraphrase detection models and crowd-workers to detect paraphrases of the clunky utterances from the pool. Last, they discarded the clunky utterance-LF pairs without corresponding paraphrases. This approach ensures that the distribution of examples in the generated dataset is close to the true example distribution. This method, however, assumes an utterance pool, which is not always feasible. Hence, given the importance of paraphrasing approaches in the automatic data duration for semantic parsing, this thesis examines various paraphrase techniques in Chapter~\ref{chap:auto}. 


\subsubsection*{Data Curation using Existing Data} 


The Overnight approach creates the dataset from scratch using prior knowledge from experts. Other types of works~\citep{wang2021learning,jia2016data,qiu2021improving,yang2022subs,yu2020grappa} use domain-specific knowledge from the existing data to generate new examples. Such methods normally cost less than the Overnight approach. For example, inducing the SCFG rules from the existing data is much easier than inducing the rules given only the database schemas. Conversely, the Overnight approach leads to inferior parsing performance due to the mismatch between the true distribution and the distribution induced by the prior assumption. Sampling the synthetic examples from the true distribution estimated over the existing dataset could mitigate this issue. This section introduces the studies that utilize the existing data. This line of research is closely related to data augmentation.


Traditional data augmentation techniques, such as flipping words or paraphrasing the utterances~\citep{wei2019eda,sennrich2015improving}, can be directly applied to the semantic parsing datasets. However, such techniques do not augment the LF compositions and thus cannot improve the compositional generalization ability of semantic parsers. In contrast, Data Recombination~\citep{jia2016data} generates new utterances by recombining the fragments of the utterances and LFs given the SCFG rules. GraPPa~\citep{yu2020grappa} generates examples by sampling the SCFG rules and replacing its non-terminals with the terminals, such as column mentions and table values, extracted from the database tables. Human experts induce the SCFG rules by observing the patterns of utterances and LFs in the original dataset. 

The aforementioned techniques generate new compositions of LFs, but sample SCFG rules from the uniform distribution, which is far from the true distribution. Therefore,~\citet{wang2021learning} estimated a probabilistic context-free grammar (PCFG) and a translation model over the original data. Then they sampled the LFs from the PCFG and apply the translation model to translate the LFs into utterances. Further, CSL~\citep{qiu2021improving} automatically induces the SCFG rules and then samples SCFG rules from a probabilistic sampling distribution learned over the original dataset. This sampling distribution ensures that the synthetic examples are in the same distribution as the true examples in the original dataset. Therefore, semantic parsers trained using CSL examples reach SOTA performance on several compositional generalization benchmarks.



\paragraph*{Curation of Multilingual Data.} Almost all the current MSP datasets are obtained by translating the utterances in existing semantic parsing datasets in high-resource languages through the automatic translation services~\citep{moradshahi2020localizing,sherborne2020bootstrapping} or human translators~\citep{susanto2017neural,duong2017multilingual,li2021mtop}. 
In most cases, the translation methods~\citep{susanto2017neural,duong2017multilingual} require translators to have only bilingual knowledge and not expert knowledge of target LFs. However,~\citet{moradshahi2020localizing} asserted that the entities in the utterances and LFs should be localised regarding the languages of the utterances. For instance, the entities `Colosseo' and `Fontana di Trevi' in Rome are more likely than `Times Square' in New York to appear in Italian questions concerning restaurants. Therefore,~\citet{moradshahi2020localizing} used heuristic methods to localize the entities in utterance-LF pairs. For the LFs in TOP~\citep{gupta2018semantic} datasets, the slots in the LFs should also be translated. Accordingly, Translate \& Fill~\citep{nicosia2021translate} and~\citet{xia2021multilingual} proposed methods to align the translated slots in the LFs with the slots in the translated utterances.

\subsection{Knowledge Transfer for Semantic Parsing}
\label{sec:transfer_knowledge}
In comparison to learning a task in isolation, in learning across tasks, the knowledge from source tasks could be used to improve the ability of a machine learning model to generalize to a new target task~\citep{spieckermann2015multi}. Furthermore, using specially designed knowledge transfer methods, the machine learning model can generalize well to the new task with only limited training data~\citep{li-etal-2016-unsupervised,zhuang2020comprehensive}. Therefore, knowledge transfer is now a trending topic in studies that aim to solve data scarcity issues.

Formally, knowledge transfer problems have a source task $\mathcal{T}^{(s)}$ and a target task $\mathcal{T}^{(t)}$ with datasets $\mathcal{D}^{(s)}$ and $\mathcal{D}^{(t)}$ in their respective domains $D^{(s)}$ and $D^{(t)}$, each with training and test sets, $\mathcal{D}=\{\mathcal{D}_{train},\mathcal{D}_{test}\}$. Each dataset is a tuple of a collection of inputs and labels, $\mathcal{D} = (X, Y)$. A domain $D = \{\mathcal{X}, P(X) \}$ is defined as an input space and the probability over the input $X \in \mathcal{X}$, and a task $\mathcal{T} =
\{\mathcal{Y},P(Y|X)\}$ is defined as the label space and the conditional distribution of label $Y \in \mathcal{Y}$ in $\mathcal{D}_{train}$ given $X$. In some settings, the single source domain can also be extended to multiple domains. Knowledge transfer uses the source task in the source domain to improve a machine learning model $f_{\bm{\theta}} : \mathcal{X}_t \rightarrow \mathcal{Y}_t$ that predicts the label in the target label space, $\vy_t \in \mathcal{Y}_t$, given the input in the target feature space, $\vx_t \in \mathcal{X}_t$. The source and target domains or tasks may differ or be the same depending on the problem settings.


Learning the model in low-data conditions is sometimes formulated as a FSL problem. In FSL, the target task includes only a handful of labeled training examples, $\mathcal{D}^{(t)}_{train} = \{(\vx_i,\vy_i)\}^{I}_{i=1}$, where $I$ is a small number.
In a $N$-way, $K$-shot classification setting, the training set $\mathcal{D}^{(t)}_{train}$ includes $N$ labels with $K$ examples associated with each label such that $I=N\times K$. However, the few-shot settings vary for semantic parsing. Most semantic parsing studies~\citep{shin2021few,drozdov2022compositional,yang2022seqzero} have considered a few-shot setting because their training set was extremely small (e.g. with less than 200 examples~\citep{shin2021few}). However, \citet{lee2019oneshot} formulated FSL since there is one example per LF template in the target task training set. Knowledge transfer has now become a standard way to solve FSL problems, for it can reduce the demand for semantic parsing training data for the target domain by using the knowledge in the source domain.


Currently, there are many research areas in knowledge transfer. However, since this thesis is more focused on semantic parsing than knowledge transfer, this section will mainly introduce the knowledge transfer techniques that have been applied to semantic parsing in low-data settings, including transfer learning, in-context learning, and meta-learning.

\subsubsection*{Transfer Learning}

There are multiple problem formulations for transfer learning which adopt different techniques. This section provides a brief overview of the prior semantic parsing studies that employ various transfer learning techniques, including joint training and sequential transfer learning, which both fall under inductive transfer learning, domain adaptation, and cross-lingual transfer learning.

\paragraph*{Joint Training.}

Joint training belongs to inductive transfer learning, which assumes the source and target tasks are different in terms of the label spaces $\mathcal{Y}_{s}\neq\mathcal{Y}_{t}$ and conditional distribution $P(Y_s|X_s)\neq P(Y_t|X_t)$. Joint training methods are similar to multi-task learning ones, which simultaneously train models for the source and target tasks. Multi-task learning focuses more on the overall performance across all tasks, whereas the objective of joint training is to maximize the performance of the model for the target tasks, even if the model must compensate for the source tasks.

Recent studies on joint training have developed auxiliary source tasks from which knowledge can be easily transferred to the target semantic parsing task with no or minimal training data. For example, \citet{rongali2022training} proposed Mask Prediction and Denoising as the auxiliary tasks. Mask Prediction masks out the spans in abundant unlabeled in-domain utterances and allows the semantic parser's encoder to predict the span. Denoising generates a noisy version of synthetic utterances sampled from SCFGs and reconstructs the synthetic utterances from the noisy utterances using the \Seq semantic parser. When training the semantic parser, \citet{shao2019weakly} introduced two auxiliary tasks: classification of question types and mention detection, which train a classifier to classify questions into pre-defined types and a detector to identify entity mentions in questions. Experiments in these works demonstrated that well-designed auxiliary source tasks can significantly improve semantic parsing performance in low-data settings. 

\paragraph*{Sequential Transfer Learning.}
\label{sec:background_transfer}
\begin{figure}
    \centering
    \includegraphics[width=0.95\textwidth]{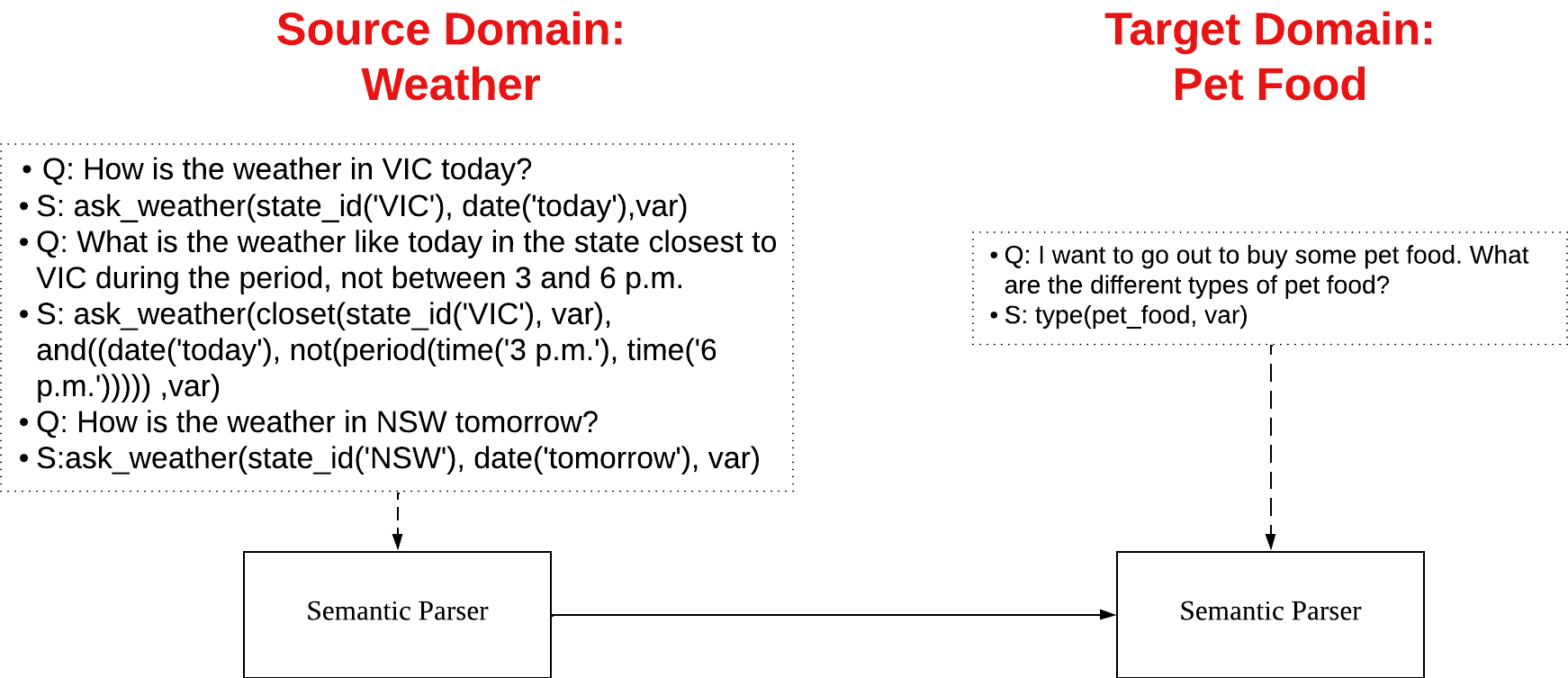}
    \caption{A transfer learning process. A semantic parser is first pre-trained on a large amount of data in a source domain and then fine-tuned on a limited number of data in a target domain.}
    \label{fig:seq_transfer}
\end{figure}

Sequential transfer learning also belongs to inductive transfer learning. In sequential transfer learning, as opposed to joint learning, the model is pre-trained on the source tasks prior to learning to generalize to the target task with limited data. The most prevalent sequential transfer learning paradigm is \textit{Pre-training and Fine-tuning}. Fine-tuning updates a portion or all of the pre-trained model parameters based on data from the target tasks. The goal of fine-tuning is to leverage the useful information from the pre-trained model, such as representations and model parameters, and distributional logits~\citep{traphd2022}.

\citet{shin2021constrained} employed the pre-trained GPT-2~\citep{radford2019language} as the backbone model to develop a few-shot semantic parsing model. The challenging part is that GPT-2 is a language model pre-trained on the general text, which is not related to the target of semantic parsing. Therefore, 
\citet{shin2021constrained} fine-tuned the GPT-2 to paraphrase the NL utterance into the clunky form, which is then converted into an LF via the SCFG rules. Similar to \citet{lewis2020bart}, \citet{yang2022seqzero} fine-tuned a pre-trained BART to generalize to a semantic parsing task with limited data. However, \citet{yang2022seqzero} decomposed the parsing problem into sub-problems, such that the model predicts the sub-clauses of each LF rather than the entire LF. Instead of fine-tuning all of the model parameters, \citet{schucher2022power} fine-tuned only the parameters of the prefix layer on top of GPT-2 while freezing the rest GPT-2 parameters, resulting in superior performance in low-data settings compared to updating all the parameters. 

\paragraph*{Domain Adaptation.} Domain adaption, unlike inductive transfer learning, assumes that the source and target domains are distinct, where $\mathcal{X}_s \neq \mathcal{X}_t$ or the conditional distribution $P(Y_s|X_s) \neq P(Y_t|X_t)$.

Both source and target tasks in DA~\citep{li2021domain} are semantic parsing tasks, while the domains of the tasks are different. Specifically, DA~\citep{li2021domain} jointly trains the semantic parser on the data of two domains, using a domain discriminator and domain relevance attention to improve knowledge transfer by separating domain-invariant from domain-specific features. \citet{su2017cross} reformulated the semantic parsing problem into a paraphrase problem because \citet{su2017cross} could utilize pre-trained word embeddings to boost the cross-domain paraphrase. In particular, \citet{su2017cross} pre-trained a \Seq model using the source domain data to generate clunky paraphrases of utterances and then fine-tuned the model using the data in the target domain. The clunky utterances would then be converted into LFs, according to the pre-defined SCFG rules. The other domain adaptation approaches usually update all model parameters while \citet{ray2019fast} fine-tuned only the encoder of the \Seq, which requires much less time to adapt the parser to the new domain.

\paragraph*{Cross-lingual Transfer Learning.} 
Cross-lingual transfer learning is a subset of transfer learning that involves the transfer of knowledge from one language to another. In cross-lingual transfer learning, the input space and distributions are distinct: $\mathcal{X}_s\neq\mathcal{X}_t$ and $P(X_s)=P(X_t)$, respectively.

\citet{duong2017multilingual} trained the \Seq parser with semantic parsing data from two languages, relying on cross-lingual word embeddings to enhance cross-lingual transfer. Currently, the majority of multilingual semantic parsers~\citep{xia2021multilingual,moradshahi2020localizing,li2021mtop} use contextual multilingual word representations generated by language models such as mBART~\citep{liu2020multilingual} and mBERT~\citep{devlin2018bert} to improve cross-lingual transfer. Specifically, the pre-trained language models of the parsers are fine-tuned on the target-language data. Consequently, the performance of these parsers is vastly superior to that of parsers employing word embeddings. To improve the performance of the multilingual semantic parser in target languages, \citet{sherborne2022zero} devised auxiliary tasks, which are to reconstruct the input utterance, translate the input utterance to the target language, and classify the input language in order to improve the language-specific features for semantic parsing. Training the parser for both the parsing and auxiliary tasks could improve the parser performance in the target language.


\subsubsection*{In-context Learning}
\begin{figure}
    \centering
    \includegraphics[width=1\textwidth]{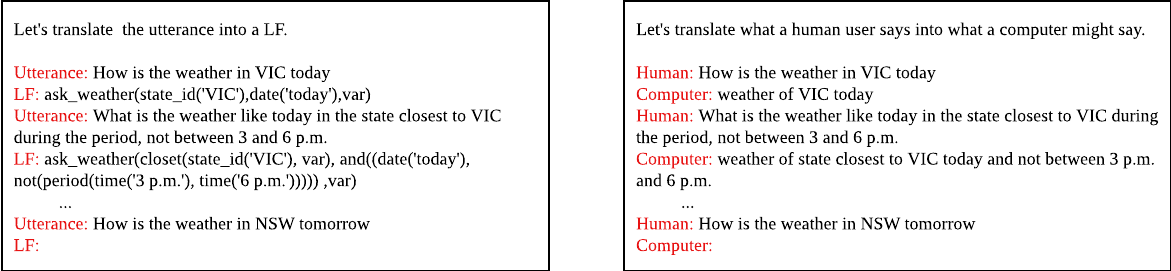}
    \caption{(Left) The prompt is used to predict the LF given the utterance. (Right) The prompt is used to paraphrase the NL utterance into the clunky canonical utterance.}
    \label{fig:prompt}
\end{figure}

In-context learning~\citep{xie2021explanation} is another way to utilize prior knowledge, which does not require optimizing any model parameters. Therefore, when using in-context learning, the performance of the parser is highly dependent on the prior knowledge acquired by the large pre-trained language models. Specifically, \citet{shin2021few} and \citet{drozdov2022compositional} used few utterance-LF pairs and \citet{shin2021few} also used table schema in the prompt to leverage prior knowledge from a pre-trained language model. 
As in~\fref{fig:prompt}, the pre-trained language models (e.g., GPT-3~\citep{brown2020language}, Codex) then infer the LFs as the tokens following the end token of the prompt (i.e. `\textit{LF:}'). The form of the prompt can vary for different purposes. As in \fref{fig:prompt}, the prompt can also include paraphrase pairs, which guide the model to paraphrase the NLs into clunky utterances.  

\subsubsection*{Meta-Learning}
In contrast, to transfer learning which focuses on how to transfer knowledge from the source task, meta-learning, a.k.a learning to learn, focuses on how to learn the meta-knowledge from the source task. For example, a typical meta-learning algorithm, Model-Agnostic Meta-Learning (MAML)~\citep{finn2017model}, explores an optimal initialization of the model parameters $\bm{\theta}$ given what is learned from the source task so that the model could be rapidly adapted to the target task with few-shot examples with the initial parameters. The Prototypical Network~\citep{snell2017prototypical} and Matching Network~\citep{vinyals2016matching} learn how to separate the embeddings of the distinct classes in the source task so that the models can distinguish unseen classes in the target task. In meta-learning, the training set of the source task is usually named the training set $\mathcal{D}_{train}$. In contrast, the training set with only few-shot examples and the test set in the target task are named as support set $\mathcal{D}_{s}$ and test set $\mathcal{D}_{test}$, respectively. The performance of the meta-learning algorithm is usually evaluated by whether the model using meta-learning could generalize well to the target task.

As aforementioned, \citet{lee2019oneshot} formulated FSL as there is one example per LF template in the training set of the target task. \citet{lee2019oneshot} then reformulated the semantic parsing problem into the template classification problem so that it could apply Matching Network to train the classifier. PT-MAML~\citep{huang2018natural}, DG-MAML~\citep{wang2021meta}, XG-REPTILE~\citep{lapata2022meta} and ~\citet{sun2020neural} all apply MAML to pre-train the model on the source task. In MAML, as outlined in~\algoref{algo:maml}, there are distinct phases for meta-training and meta-testing to emulate the cross-task, cross-domain, and cross-language generalization processes that occur during both pre-training and fine-tuning. The key difference between these methods lies in their approach to batch sampling for meta-training and meta-testing, which addresses specific generalization challenges. For instance, PT-MAML strategically samples batches in both the meta-training and meta-testing sets based on the similarity of the LFs within each batch. This is designed to enhance the model's cross-task generalization capabilities. In contrast, DG-MAML and XG-REPTILE use meta-training and meta-testing sets from different domains and languages to enhance the models' capabilities in cross-domain and cross-lingual generalization, respectively.

\begin{algorithm}[ht]
{\small
\SetKwData{Left}{left}\SetKwData{This}{this}\SetKwData{Up}{up}
\SetKwFunction{Union}{Union}\SetKwFunction{FindCompress}{FindCompress}
\SetKwInOut{Input}{Input}\SetKwInOut{Output}{Output}
\SetAlgoLined
\Input{Distribution over tasks $P(\mathcal{T})$,  step size hyperparameters, $\alpha$ and $\beta$}
\Output{A well-initialized set of model parameters \(\bm{\theta}\)}
Randomly initialize model parameters \(\bm{\theta}\)\\
\While{\text{not done}} 
{
\textcolor{blue}{Task Sampling}\\
Sample a batch of tasks \(\mathcal{T}_{i} \sim P(\mathcal{T})\) \\
\For{\text{\textbf{all}} $\mathcal{T}_{i}$
}
{
\textcolor{blue}{Meta-Train Step}\\
Compute the gradient \(\nabla_{\bm{\theta}}\mathcal{L}_{\mathcal{T}_{i}}(\bm{\theta})\) using \(K\) examples from \(\mathcal{T}_{i}\) \\
Compute adapted parameters with gradient descent: $\bm{\theta}'_{i} = \bm{\theta} - \alpha\nabla_{\bm{\theta}}\mathcal{L}_{\mathcal{T}_{i}}(\bm{\theta})$
}
\textcolor{blue}{Meta-Test Step}\\
Update $\bm{\theta} = \bm{\theta} - \beta\nabla_{\bm{\theta}}\sum_{\mathcal{T}_{i} \sim P(\mathcal{T})}\mathcal{L}_{\mathcal{T}_{i}}(\bm{\theta}'_{i})$
}
\caption{ Model-Agnostic Meta-Learning~\citep{finn2017model}
}
\label{algo:maml}
}
\end{algorithm}

\subsection{Active Learning for Semantic Parsing}
\label{sec:al_background}
The essence of AL is selecting the most valuable unlabeled instances to be annotated in order to maximize the model's performance, and hence reduce the annotation cost for data-hungry machine learning models. AL has been used in various natural language processing (NLP) fields, including MT~\citep{vu2019learning}, text classification~\citep{mccallum1998employing}, and semantic parsing~\citep{duong-etal-2018-active}. 

There are now two types of AL scenarios: stream-based and pool-based~\citep{liu2019weak}. In a stream-based environment, AL systems successively access unlabeled data points in a stream, with the AL algorithm determining the examples to retain and to discard. Pool-based AL, on the other hand, assumes the existence of a large unlabeled data pool from which the AL algorithm can select samples of data to be annotated. All current AL studies have focused on data annotation for monolingual semantic parsing, whereas the studies discussed in Chapters~\ref{chap:al_msp} and~\ref{chap:comb_human_auto} have focused on data annotation for MSP in the pool-based setting. Next, the two AL scenarios in the context of semantic parsing will be discussed.
\subsubsection*{Stream-based Active Learning} 
Stream-based AL is formed as feedback/interactive semantic parsing in semantic parsing. The majority of feedback semantic parsing systems~\citep{iyer2017learning,yao2019model,yao2019interactive,elgohary2020speak} follow these steps: 
\begin{enumerate}
    \item In a dialogue system, a semantic parser converts a given user utterance into an initial LF.
    \item Then the LF is interpreted in NL and sent to a user. 
    \item The user provides feedback, based on which the systems revise the initial LF.
    \item Update the parser given the updated data.
    \item This process is repeated until the model converges.
\end{enumerate}

\subsubsection*{Pool-based Active Learning}
\label{sec:pool_al}

\begin{figure}
    \centering
    \includegraphics[width=1\textwidth]{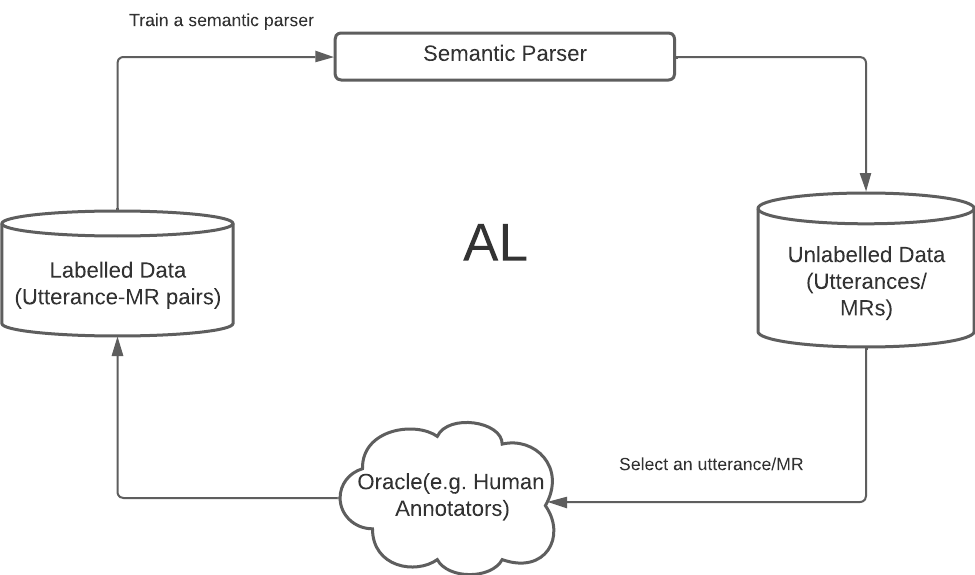}
    \caption{The pool-based AL procedure in~\citet{duong-etal-2018-active,sen-yilmaz-2020-uncertainty}}
    \label{fig:alsp}
\end{figure}

\begin{algorithm}[ht]
{\small
\SetKwData{Left}{left}\SetKwData{This}{this}\SetKwData{Up}{up}
\SetKwFunction{Union}{Union}\SetKwFunction{FindCompress}{FindCompress}
\SetKwInOut{Input}{Input}\SetKwInOut{Output}{Output}
\SetAlgoLined
\Input{Unlabeled data pool $\mathcal{D}_{u}$, labeled data $\mathcal{D}_{l}$, a test set $\mathcal{D}_{test}$, budget size $K$, number of the selection rounds $Q$}
\Output{A well-trained parser $P_{\bm{\theta}}(\vy|\vx)$}
Train the parser $P_{\bm{\theta}}(\vy|\vx)$ on the annotated data $\mathcal{D}_{l}$ \\
\For{$q \gets 1$ to $Q$} 
{
Estimate the acquisition $\phi(\cdot)$ \\
Select a subset $\mathcal{D}_{u}' \in \mathcal{D}_{u}$ of the size $K$ based on the acquisition function $\phi(\cdot)$ \\
Annotate the unlabeled data, $\mathcal{D}_{l}' = \text{Annotate}(\mathcal{D}_{u}')$\\
Combine the annotated data, $\mathcal{D}_{l}=\mathcal{D}_{l}' \cup \mathcal{D}_{l}$\\
Exclude the selected data from the unlabeled pool, $\mathcal{D}_{u} = \mathcal{D}_{u} \setminus \mathcal{D}_{u}'$ \\
Re-train the parser $P_{\bm{\theta}}(\vy|\vx)$ on all annotated data $\mathcal{D}_{l}$ \\
Evaluate parser performance on the test set $\mathcal{D}_{test}$
}
}
\caption{The active learning algorithm in~\cite{duong-etal-2018-active,sen-yilmaz-2020-uncertainty}
}
\label{algo:al}
\end{algorithm}


\fref{fig:alsp} illustrates a pool-based AL paradigm from~\citet{duong-etal-2018-active}. In pool-based AL, there are typically numerous iterations of sampling. At each iteration, AL will select examples with the highest acquisition scores from the unlabeled data pool. 
In the context of semantic parsing, unlabeled data can be LFs~\citep{duong-etal-2018-active} when the Overnight~\citep{wang2015overnight} collection method is used, or unlabeled utterances~\citep{duong-etal-2018-active,sen-yilmaz-2020-uncertainty,ni2020merging} when the standard data collection approaches are used. The samples will then be annotated, and the parser will be retrained on all annotated data up to the current iteration. The detailed algorithm is as depicted in~\algoref{algo:al}.

The acquisition function $\phi(\cdot)$ assigns a score to each example, and using these scores, AL algorithms determine the examples to annotate. There are three common types of acquisition functions that measure the uncertainty, diversity, and density of unlabeled data points: 
\paragraph*{Uncertainty.} This type of methods selects examples for which the model is most uncertain about how to label them. For example, the Least Confidence~\citep{duong-etal-2018-active,ni2020merging} selects the utterances, for which the parser has least confidence in the their most likely output LFs:
\begin{align}
    \phi(\vx) &=  1 - P_{\bm{\theta}}(\vy'|\vx) \\
    s.t. \vy' &= \argmax_{\vy \in \mathcal{Y}} P_{\bm{\theta}}(\vy|\vx)
    \label{eq:confidence}
\end{align}
where $\vy'$ is the most likely output LF of parser $P_{\bm{\theta}}(\vy|\vx)$ given the utterance $\vx$. Traffic-aware~\citep{sen-yilmaz-2020-uncertainty} selects the example with the highest language model perplexity per word in the utterance. N-best Sequence Entropy~\citep{settles-craven-2008-analysis} is another way to measure the parser's uncertainty regarding the utterance. It approximates the entropy of parser's outputs given the utterance:
\begin{align}
    \phi(\vx) &=  - \sum_{\vy \in \mathcal{N}} P'_{\bm{\theta}}(\vy|\vx) \log P'_{\bm{\theta}}(\vy|\vx)\\
    s.t. P'_{\bm{\theta}}(\vy|\vx) &= \frac{P_{\bm{\theta}}(\vy|\vx)}{\sum_{\vy \in \mathcal{N}}P_{\bm{\theta}}(\vy|\vx)}
    \label{eq:n_best_entropy}
\end{align}
where $\mathcal{N} = \{\vy'_0,...,\vy'_{|\mathcal{N}|}\}$ are the most likely $|\mathcal{N}|$ output LFs of the parser $P_{\bm{\theta}}(\vy|\vx)$. The higher entropy means that the parser has less confidence in the generated LF sequences. The \textit{N-best Sequence Entropy} is applied to the AL work for MSP in~\cref{chap:comb_human_auto} of this thesis.
\paragraph*{Diversity.} The Diversity-based methods diversify the selected examples. \citet{duong-etal-2018-active} used the K-means~\citep{han2022data} clustering algorithm to partition the examples in the pool into different clusters, then randomly select one example from each cluster. \citet{ni2020merging} select the utterance with the greatest number of words that results in an incorrect prediction of LFs. This approach requires access to the denotation label, and therefore LFs that cannot result in accurate denotations are regarded erroneous.
\paragraph*{Density.} Density-based acquisition functions score examples based on their distance to other examples. A higher density score indicates that this example is more representative. The general form for estimating the kernel density of an example is:
\begin{equation}
    \phi(\vx) = \frac{1}{k} \sum_{\vx' \in NN_{k}(\vx)} K(\vx,\vx')
\end{equation}
where $K(\cdot)$ is the kernel function estimating the distance between two examples, $NN_{k}(\vx)$ is the set of $k$ nearest neighbors of the example $\vx$. CSSE~\citep{hu2021phrase} is a variation of the density kernel estimation, which is utilized in MT and serves as the baseline for the work in Chapters~\ref{chap:al_msp} and~\ref{chap:comb_human_auto}. CSSE selects the utterances from the highest density regions and avoids selecting utterances similar to the labeled utterances. The acquisition function is as follows:
\begin{align}
\label{eq:csse}
    \phi(\vx) & = \min_{\vx' \in \mathcal{D}_{l}} \text{ratio}(\ve_{\vx},\ve_{\vx'}) \\
    \text{ratio}(\ve_{\vx},\ve_{\vx'}) & = \frac{\cos{(\ve_{\vx},\ve_{\vx'})}}{\sum\limits_{\vz \in NN_{k}(\vx)}\frac{\cos{(\ve_{\vx},\ve_{\vz})}}{2k} + \sum\limits_{\vz \in NN_{k}(\vx')}\frac{\cos{(\ve_{\vx'},\ve_{\vz})}}{2k}}
\end{align}
where $\cos{(\ve_{\vx},\ve_{\vx'})}$ is the cosine similarity between the representations of two utterances $\ve_{\vx}$ and $\ve_{\vx'}$. The representation of each utterance is extracted by the pre-trained language model, BERT~\citep{devlin2018bert}.

\subsection{Continual Learning for Semantic Parsing}
\label{sec:back_cl}

As described in~\sref{sec:al_background}, AL incrementally collects data. In real applications, this situation is also quite common. For example, data for semantic parsing are usually collected task by task. When new tasks and data arrive, usually, the semantic parser is retrained on all the seen tasks to ensure the parser performs well on all tasks. However, it is computationally costly to retrain the parser each time on all seen tasks. Then, CL is a cost-effective method for reducing the required computational resources. With CL, the model can be fine-tuned only on the new data in order to reduce the training time. CL focuses on both facilitating the forward transfer and reducing negative backward transfer. In the facilitating forward transfer process, the knowledge from previous tasks is used to boost the learning on current tasks. The techniques used in CL are similar to those in transfer learning. Please refer to~\cref{sec:transfer_knowledge} for the descriptions of transfer learning techniques. The negative backward transfer, also termed \textit{catastrophic forgetting}, means that fine-tuning the model on the new data will cause the model to forget what it learned from the old tasks. Currently, most studies on CL have aimed to mitigate catastrophic forgetting.

There are online and offline settings for CL. In online settings, training examples for the semantic parser are provided to the learner one at a time without an explicit task boundary identifier. In contrast, offline settings present the semantic parser with a batch of training examples each time, with distinct task boundary identifiers between batches. In semantic parsing, all current continual learning works, including the work outlined in~\sref{chap:continual}, are offline. This section therefore only provides a comprehensive overview of the efforts on offline CL for semantic parsing.

\begin{figure}
    \centering
    \includegraphics[width=0.95\textwidth]{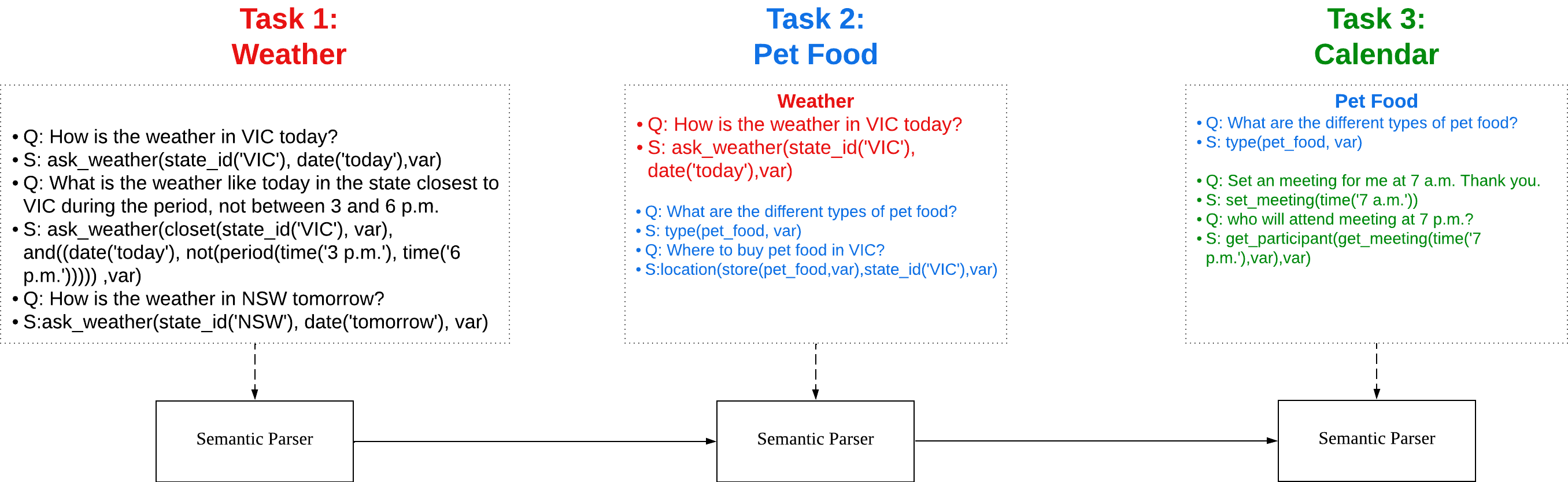}
    \caption{The offline CL procedure for semantic parsing.}
    \label{fig:sp_cl}
\end{figure}

As in~\fref{fig:sp_cl}, in offline CL~\citep{lopez2017gradient}, the parser $P_{\bm{\theta}}(\vy|\vx)$ is trained continually on a sequence of $K$ distinct tasks \{$\mathcal{T}^{(1)},...,\mathcal{T}^{(k)},\mathcal{T}^{(K)}$\}. Each task $\mathcal{T}^{(k)}$ includes its own training/validation/test set ($\mathcal{D}^{(k)}_{train}$,$\mathcal{D}^{(k)}_{valid}$,$\mathcal{D}^{(k)}_{test}$), each of which is a set of utterance-LF pairs $(X^{(k)}, Y^{(k)})$. The \textit{boundaries} between tasks are defined by \textit{task descriptors}~\citep{lopez2017gradient}, which provide contextual information about example pairs $(X^{(k)}, Y^{(k)})$ in each task. For example, the descriptors might describe that each task belongs to a distinct domain, or the question utterances $X^{(k)}$ in each task are in a unique question type that query about a distinct set of cities. A fixed-size memory $\mathcal{M}_{k}$ is also associated with each task $\mathcal{T}^{(k)}$ for storing a small amount of replay instances, as adopted in~\citep{rebuffi2017icarl,wang2019sentence}.

CL has found applications across a wide range of NLP tasks. For instance, it has been used in relation extraction~\citep{wang2019sentence,han2020continual}, natural language generation~\citep{mi2020continual}, language modeling~\citep{sun2019lamol}, and in adapting pre-trained language models to diverse NLP tasks~\citep{wang2020k,pfeiffer2020adapterfusion}.

The methods employed in CL can be broadly classified into four main categories:
\begin{itemize}
    \item \textbf{Regularization-Based Methods:} These techniques, as in~\citet{kirkpatrick2017overcoming,zenke2017continual,ritter2018online,li2017learning,zhao2020maintaining,schwarz2018progress}, use either knowledge distillation~\citep{hinton2015distilling} to penalize loss updates or apply regularization to parameters crucial for retaining the models with knowledge of earlier tasks.
    \item \textbf{Dynamic Architecture Methods:} As explored in~\citet{mallya2018packnet,serra2018overcoming,maltoni2019continuous,houlsby2019parameter,wang2020k,pfeiffer2020adapterfusion,rusu2016progressive}, these methods dynamically modify model architectures to mitigate catastrophic forgetting.
    \item \textbf{Memory-Based Methods:} These approaches, presented in works like~\citet{lopez2017gradient,wang2019sentence,han2020continual,aljundi2019gradient,chrysakis2020online,kim2020imbalanced}, store instances from past tasks and continue to learn from them alongside new tasks.
    \item \textbf{Hybrid Methods:} There are hybrid methods that integrate features from multiple categories~\citep{mi2020continual,liu2020mnemonics,rebuffi2017icarl}.
\end{itemize}

Despite the range of methods available, most of the existing research specifically in the realm of continual semantic parsing~\citep{lialin2020clNeuralSemanticParsing,sun2019lamol,zhang2022continual} tends to apply general CL techniques, which are not specially designed for semantic parsing tasks, to address issues related to catastrophic forgetting.




Next, several important CL approaches are discussed.
\paragraph*{Regularization-based Approaches.} These methods overcome catastrophic forgetting by preserving network parameters close to those of networks learned from previous tasks. The learning objective in a regularization method, namely Elastic Weight Consolidation (EWC)~\citep{kirkpatrick2017overcoming}, is:

\begin{equation}
    \mathcal{L}(\bm{\theta}) = \mathcal{L}_{\mathcal{D}^{(k)}_{train}}(\bm{\theta}_{k}) + \lambda \sum^{N}_{j=1} F_{j}(\bm{\theta}_{k,j}-\bm{\theta}_{k-1,j})^{2}
\end{equation}
where $\mathcal{L}_{\mathcal{D}^{(k)}_{train}}(\bm{\theta}_{k})$ is the loss for task $k$ only, $N$ is the number of model parameters, $\bm{\theta}_{k-1,j}$ is the model parameters learned until $\mathcal{T}^{(k-1)}$, $\lambda$ is the coefficient and $F_{j}$ is the second-order gradient, $F_{j} = \nabla^{2}\mathcal{L}(\bm{\theta}_{k-1,j})$ given the instances stored in $\mathcal{M}$.

\paragraph*{Dynamic Architecture Approaches.} These approaches alleviate forgetting by isolating a different set of model parameters for each task. For example, Hard Attention to the Task (HAT)~\citep{serra2018overcoming} applies different attention masks on the layers of the neural networks while learning the model for each task. Likewise, the adapter-based methods~\citep{wang2020k,pfeiffer2021adapterfusion,zhang2022continual} dynamically add an adapter that includes isolated parameters between layers of the Transformers for each task.
\paragraph*{Memory-based Approaches.} These approaches memorize what has learned by replaying the model on the memory instances. Currently, two approaches are popular. The Episodic Memory Replay (EMR)~\citep{wang2019sentence,chaudhry2019tiny} trains the model on the data of the current task as well as the data from previous tasks stored in memory $\mathcal{M}$. The other one, Gradient Episodic Memory (GEM)~\citep{lopez2017gradient}, minimize an objective:
\begin{align}
    \mathcal{L}(\bm{\theta}) &= \mathcal{L}_{\mathcal{D}^{(k)}_{train}}(\bm{\theta}_{k})\\
    s.t.  \mathcal{L}_{\mathcal{M}}(\bm{\theta}_{k}) &\leq \mathcal{L}_{\mathcal{M}}(\bm{\theta}_{k-1})
\end{align}
Such an objective ensures that training the model on the current task would not lower the performance of the model on the sampled data. 

The technique used to sample the examples in memory $\mathcal{M}$ is also important. Replaying good examples in $\mathcal{M}$ can largely reduce the catastrophic forgetting issues. Most sampling techniques are highly relevant to the sampling techniques in AL. For example, \textbf{FSS}~\citep{aljundi2019gradient,wang2019sentence,mi2020continual}, \textbf{GSS}~\citep{aljundi2019gradient} and \textbf{LFS} diversify the examples in the $\mathcal{M}$, which partition the instances into clusters w.r.t. the spaces of utterance encoding features, instance gradients, and LFs, respectively, and then select the instances that are closest to the centroids. Other sampling techniques, such as uncertainty- or density-based sampling, can also be directly applied to CL sampling.
\paragraph*{Hybrid Approaches} The hybrid methods integrate more than one category of methods. For example, \citet{lialin2020clNeuralSemanticParsing} combine EMR, and EWC while~\citet{chen2022learn} combine EMR and adapters to reduce negative backward transfer for continual semantic parsing.
\paragraph*{Evaluation of Continual Learning.}

A suitable CL method facilitates forward transfer and reduces catastrophic forgetting, enabling the model to perform well on all tasks. Therefore, two evaluation settings in CL are commonly used to evaluate the overall model performance across all tasks and have been adopted as the evaluation metric for this thesis in~\cref{chap:continual}. 

One commonly used evaluation metric, as employed by~\citet{lopez2017gradient,wang2019sentence}, calculates performance by taking the average of parser accuracies across the test sets for all tasks observed up to task $\mathcal{T}^{(k)}$. Mathematically, this is expressed as:
\begin{equation}
    \text{ACC}_{\text{avg}} = \frac{1}{k} \sum^{k}_{i=1} acc_{i,k}
\end{equation}
where $acc_{i,k}$ denotes the accuracy of the parser on the test set $\mathcal{D}_{test}^{(i)}$ after training the parser on task $\mathcal{T}^{(k)}$.

An alternative evaluation approach is proposed by~\citet{wang2019sentence,han2020continual}. In this setting, the accuracy is evaluated on the test sets of all observed tasks combined, following the completion of training on the $\mathcal{T}^{(k)}$. The corresponding equation for this measure is:
\begin{equation}
    \text{ACC}_{\text{whole}} = acc_{\mathcal{D}^{(1:k)}_{test}}
\end{equation}
\section{Summary}
This chapter delves into the fundamental concepts of semantic parsing, with a special focus on prior studies concerning low-resource semantic parsing. In particular,~\sref{sec:prelim} introduces the semantic parsing framework and four types of semantic parsing models. Subsequently,~\sref{sec:limited} reviews prior related studies that have employed techniques such as automatic data curation, knowledge transfer, AL, and CL to improve the performance of semantic parsers in low-resource conditions. In the chapters that follow, this thesis aims to explore and extend these studies to address the challenges associated with low-resource semantic parsing.

\part{Limited Parallel Data for Semantic Parsing}
\label{part1}
\chapter{Automatic Data Curation for Zero-shot Semantic Parsing}
\epigraph{The art and science of asking questions is the source of all knowledge.}{\textit{Thomas Berger}}
\label{chap:auto}

\fbox
{
\begin{minipage}{0.96\textwidth}
     This chapter is based on:\\
      F. Shiri$^{*}$, T. Y. Zhuo$^{*}$, Z. Li$^{*}$, V. Nguyen, S. Pan, W. Wang, G. Haffari, Y. F. Li. ``Paraphrase Techniques for Maritime QA System'' Authors marked with $^{*}$ have contributed equally and share first authorship. 25th International Conference on Information Fusion (FUSION). 2022.
\end{minipage}
}

This chapter answers RQ1: `How can a semantic parser be rapidly adapted to a new task when no parallel data is available but just structured data in a database exists?' Following the prior studies described in~\sref{sec:cura_scratch}, we propose an automatic data curation approach to automatically generate massive training data in a new domain for the semantic parser.

Previous approaches~\citep{xu2020autoqa,yin2021ingredients} synthesized parallel training data by paraphrasing the clunky canonical utterances automatically generated by the SCFG. The SCFG is written given the structured data in the database. In this chapter, we employ a similar strategy to create synthetic data. In addition, we improve upon previous data generation work by examining how to use the most effective paraphrasing techniques for the automated generation of large-scale training datasets. The contributions of this chapter are i) a well-trained semantic parser in the maritime domain with zero manually annotated data and ii) the thorough evaluation and analysis of techniques for the paraphrasing techniques used in automatic semantic parsing data curation.  



\section{Introduction}

Traditional methods require humans to interact with a computer using a machine-readable language, such as SQL, Prolog, or SPARQL. However, such interaction is difficult because it requires normal users to be proficient in machine language. Semantic parsing, on the other hand, is the process of automatically converting NL into machine-readable LFs via a semantic parser. With semantic parsing, users can interact with machines without knowing about machine languages. Therefore, in recent years, there has been a growing demand for semantic parsing in various human-computer interaction scenarios, such as question-answering (QA) and dialogue systems.

However, training a semantic parser requires a substantial amount of training data. The conventional method collects data by annotating utterances with LFs, which is expensive and usually requires a collection of utterances. However, in many situations, neither the utterance pool nor sufficient resources exist to annotate LFs. For instance, in a small company that has just launched its QA system, it is impractical to collect user utterances because there are no users yet, and it is expensive to collect user utterances and LFs to launch the business. To resolve this issue, recent studies~\citep{xu2020autoqa,yin2021ingredients} proposed approaches to automate the data generation process. As described in~\sref{sec:cura_scratch}, these approaches rely on SCFG grammar rules to generate a large amount of clunk utterance-LF pairs and a paraphraser to paraphrase the clunky utterances into the fluent NL utterances. However, these approaches only evaluate one technique for paraphrasing based on BART~\citep{lewis2020bart}, and thus have not yet explored the full potential of all techniques. 

In this chapter, we investigate the application of semantic parsing toward enhancing human-system interaction.
Particularly in the maritime domain, where there is no utterance pool or zero manually annotated utterances, we learn a semantic parser on over 300,000 domain-specific synthesized examples to answer the maritime questions. Moreover, we explore different paraphrasing techniques to reduce the gap between synthesized and NL utterances.

Our contributions are: 
\begin{itemize}
\item An efficient and effective approach for generating large-scale domain-specific training datasets, which are used to build a semantic parser for the maritime question-answering systems.



\item Evaluation and analyses of various paraphrasing techniques to produce diverse and high-quality natural language questions in the synthetic QA dataset.

\end{itemize}

\section{Automatic Data Curation with Paraphrase Generation}\label{sec:approach}

\begin{figure}
    \centering
    \resizebox{0.9\textwidth}{!}{
    \includegraphics{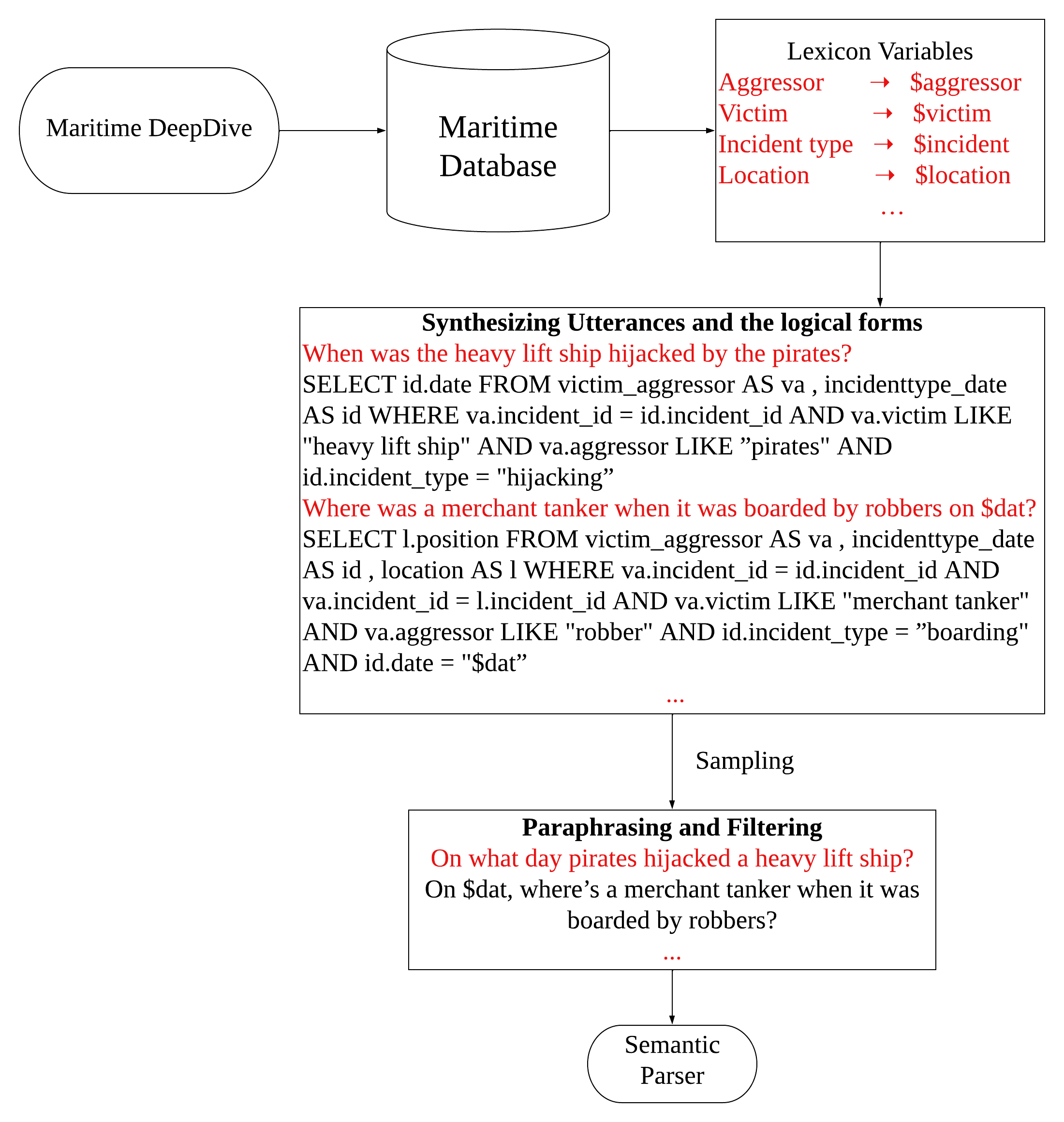}}
    \caption{An illustration of our end-to-end framework.}
    \label{fig:pipline}
\end{figure}


As shown in \fref{fig:pipline}, our proposed approach improves on the existing Overnight~\citep{wang2015overnight} approach for bootstrapping semantic parsers by requiring no manually annotated training data. Instead, large-scale training datasets can be generated automatically using existing automatic paraphrasing techniques. Given a knowledgebase, we design a compact set of SCFG rules to generate \textit{seed} examples as pairs of clunky canonical utterances and corresponding LFs. Following this, we select a representative subset from the initial seed examples. Utilizing filtering techniques and various paraphrasing methods, we expand this selected subset into a significantly larger and more diverse collection of examples. 
Please refer to~\sref{sec:cura_scratch} for the detailed description of Overnight.


\subsection{Knowledgebase}
We employ Maritime DeepDive \citep{shiri2021toward}, a probabilistic knowledge graph in the maritime domain, as the query environment for the LFs generated by our semantic parser. This knowledge graph is automatically constructed from NL data collected from two main sources: i) Worldwide Threats To Shipping (WWTTS)\footnote{\url{https://msi.nga.mil/Piracy}}, and ii) the Regional Cooperation Agreement on Combating Piracy and Armed Robbery against Ships in Asia (ReCAAP)\footnote{\url{https://www.recaap.org/}}. Both datasets include 1,452 piracy reports in total. Utilizing this dataset gives us complete control over the underlying database. Therefore, we can extract relevant entities, concepts, and their associated semantic relationships from the knowledgebase. This curated maritime knowledge graph serves as the primary database for developing our QA system. 

The constructed knowledge graph contains a set of entities such as {\it{victim}}, {\it{aggressor}}, and triples like {\it{($e_1$, r, $e_2$)}}, where {\it{$e_1$}} and {\it{$e_2$}} are entities and {\it{r}} is a relationship (e.g., {\it{victim\_aggressor}}). This database can be queried using SQL, a type of LF.
\subsection{Synchronous Context-free Grammar}

\begin{figure*}
    \centering
    \resizebox{.9\textwidth}{!}{
    \includegraphics{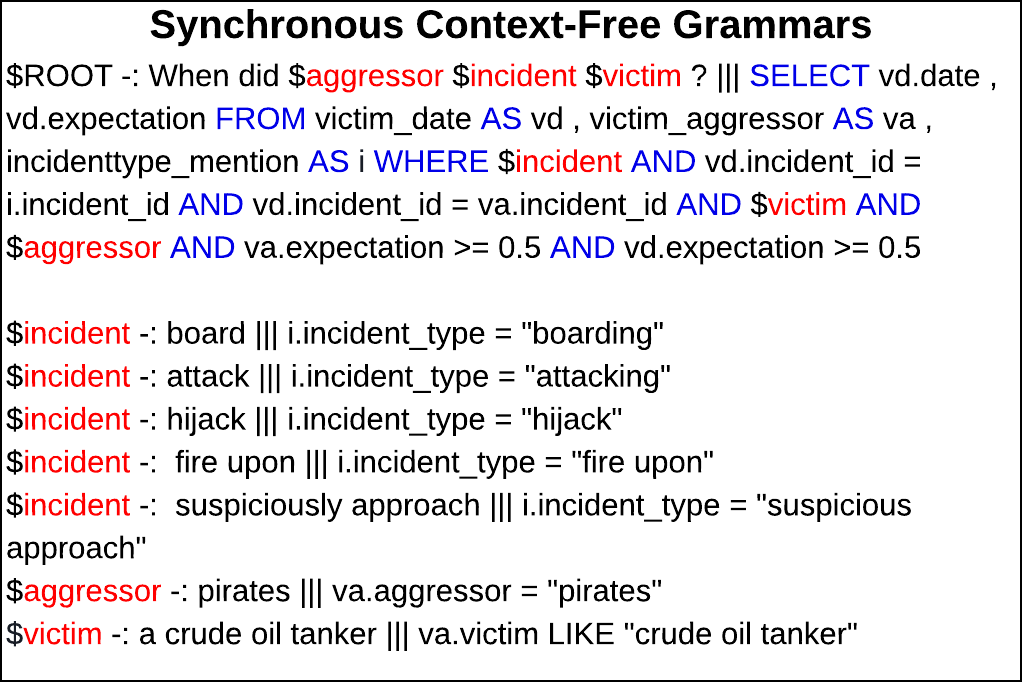}}
    \caption{A snippet of the synchronous grammar rules.}
    \label{fig:grammar}
\end{figure*}

We propose to design compact grammars with only 31 SCFG rules that simultaneously generate both LFs and clunky canonical utterances that are understandable by a human. First, we define variables specifying a canonical phrase such as {\it{victim}},  {\it{aggressor}}, {\it{incident\_type}},  {\it{date}}, {\it{position}},  {\it{location}}. Second, we develop grammar rules for different SQL structures (\fref{fig:grammar}). Finally, we build a framework using the grammar rules and the list of variables and their possible values ($G$ , $L$) to automatically generate canonical utterances paired with their SQLs, $\{(\vx_{i},\vy_{i})\}^{N}_{i=1}$, exhaustively (\fref{fig:grammar}). The SCFG rules yield many canonical examples to train a semantic parser. Then, we assign real values to the domain-specific variables including {\it{victim}} (victimized ships and individuals, e.g. oil tanker),  {\it{aggressor}} (e.g., pirates), {\it{incident\_type}} (e.g., robbery and hijacking). However, we replace all the domain-general variables including {\it{location}} (approximate place, e.g. country), {\it{position}} (place with longitude and latitude), and {\it{date}} with abstract variables  $\$loc$,  $\$pos$ and $\$dat$, which can be later replaced with the actual entities in the SQL queries using a named entity recognition (NER) model. 

With 31 grammar rules, the list of variables, and the Maritime database, we automatically synthesize 341,381 canonical utterances paired with SQL queries. However, semantic and grammatical mismatches exist between the synthetic canonical examples and real-world user-issued ones. To close these gaps, paraphrasing the synthesized utterances can produce linguistically diverse questions that are better aligned with user-generated queries.
\subsection{Question Collection Sampling}
\citet{oren-etal-2021-finding} demonstrated that, with a proper sampling method, the semantic parser can also achieve decent parsing performance with much less synthetic training data. Following this insight, we only select a subset of synthetic canonical examples (10\%) to reduce our paraphrase and training costs. We use an approach similar to Uniform Abstract Template Sampling (UAT)~\citep{oren-etal-2021-finding} to obtain utterances that correspond to diversified SQL structures for paraphrasing. UAT aims to diversify the atoms and idioms within the selected SQL queries. Please see~\sref{sec:parser_framework} for the descriptions of atoms and compounds.

There are 35,050 canonical examples in our sampled dataset, with 49\% of samples containing abstract variables.
\subsection{Paraphrase Generation}
\begin{table*}[t!]
\centering
\resizebox{\textwidth}{!}{
\begin{tabular}{ll}
\hline
\textbf{Canonical   Utterance:}          & which weapon did pirates use to rob the offshore supply vessel on $\$dat$ in $\$loc$?       \\ \hline
\multicolumn{1}{l|}{\textbf{Techniques}} & \textbf{Paraphrased  Utterances}                              \\ \hline
\multicolumn{1}{l|}{Back-translation (Spanish)}    &  what weapon did the pirates used to steal the offshore supply ship in $\$dat$ at $\$loc$?   \\
\multicolumn{1}{l|}{Back-translation (Telugu)}    &  which weapon is used to rob the offshore supply vessel in $\$dat$ by pirates in $\$loc$?  \\
\multicolumn{1}{l|}{Back-translation (Chinese)}    &  which weapon is used in $\$loc$? it is used to rob the offshore supply boat? \\
\multicolumn{1}{l|}{GPT-3}               & what was the weapon used by pirates to rob the offshore supply vessel on $\$dat$ in $\$loc?$ \\
\multicolumn{1}{l|}{BART}                & what gun did the pirates use to rob an offshore supply vessel in $\$loc$ on $\$dat$? \\
\multicolumn{1}{l|}{Quillbot}            & when pirates robbed an offshore supply vessel on $\$dat$ in $\$loc$, what weapon did they use?                         \\ \hline
\end{tabular}
\centering
}
\caption{Examples of utterances paraphrased using various techniques.}
\label{tab:examples}
\end{table*}
The paraphrase generation model enhances the diversity of canonical utterances by rewriting them into various alternative forms. These rewritten and canonical utterances, when paired with SQL queries, can be combined to train the semantic parser. However, previous studies have not fully explored the potential of current paraphrasing methodologies, often relying solely on a single approach such as BART for paraphrasing.

To advance beyond existing research, we investigate diverse augmentation strategies specifically designed for \textit{paraphrase generation} in domain-specific contexts. Our approach incorporates techniques such as \textbf{back-translation}, \textbf{prompt-based language models}, \textbf{fine-tuned autoregressive language models}, and \textbf{commercial paraphrasing models}.


\subsubsection*{Paraphrasing using Back-Translation}
Back-translation is a data augmentation technique widely used in several studies \citep{beddiar2021data, romaissa2021data, prabhumoye2018style}. Given a sentence, we aim to translate it into another language and then translate it back. The back-translated sentence will often differ slightly from the original. We thoroughly evaluated the performance of the Google Translation API~\citep{wu2016google}, a well-known tool commonly used for everyday translations and capable of translating 109 languages. The strength of the Google Translate API lies in its use of powerful neural MT methods, which are enhanced by advanced techniques within a well-designed pipeline architecture. 

To ensure the diversity of paraphrased data, we adopt the procedure as defined in \citet{corbeil2020bet}:
\begin{itemize}
    \item Cluster languages into different family branches via the language ontology from Wikipedia info-boxes.
    \item Select the most used languages from each family and translate them using an appropriate translation system.
    
    
\end{itemize}

We select Chinese (medium-resource), Spanish (high-resource), and Telugu (low-resource) as bridge languages for back-translation. 



\subsubsection*{Paraphrasing using Prompt-based Language Model}

Several studies \citep{shin2021few, shin2021constrained, schucher2021power} have explored few-shot large pre-trained language models such as GPT-3~\citep{brown2020language} in the context of semantic parsing. Traditional methods often fine-tune these pre-trained models for specific tasks, but the current approach uses GPT-3 via in-context learning. The preliminary concept of in-context learning has been described in Section~\ref{sec:background_transfer}. This approach allows us to transform specialized tasks into general language modeling challenges. By crafting a well-designed input text sequence, known as a \textit{prompt}, we can guide the model to generate the desired output, such as a paraphrased utterance, without updating the model's parameters. Our empirical analyses show that in-context learning is highly effective for various tasks in scenarios with limited data availability, regardless of whether the prompts are hand-crafted~\citep{goodfellow2016deepbook} or automatically generated~\citep{banarescu2013abstract}. None of the existing methods reports the ability of a pre-trained language model to generate paraphrased utterances in a zero-shot setting. Our chapter addresses this gap by providing an instructive prompt specifically designed for GPT-3\footnote{We employ the GPT-3 Davinci variant.}, thus enabling it to produce paraphrases in a zero-shot context. As illustrated in~\tref{tab:gpt3_prompt}, our prompt consists solely of instruction and does not include paired examples of utterances and their paraphrases for in-context learning.

\begin{table}[ht]
\noindent\rule{.5\textwidth}{1pt}\\
\\
Prompt: I'm a professional and creative paraphraser. I'm required to paraphrase the Original sentence in other words. The output sentence must not be the same as the given sentence. \\

Original: How many heavy lift vessels have been approached in \$loc ? \\
Paraphrased: \textcolor{cyan}{How many heavy lift vessels have been contacted in \$loc?} \\

\noindent\rule{.5\textwidth}{1pt}\\
\\
Prompt: I'm a professional and creative paraphraser.\\

Original: How many heavy lift vessels have been approached in \$loc ? \\
Paraphrased: \textcolor{purple}{How many heavy lift vessels have been approached  in \$loc ?} \\
\noindent\rule{.5\textwidth}{1pt}\\
\\
Prompt: I'm a professional and creative paraphraser. I'm required to paraphrase the Original sentence in other words. \\

Original: How many heavy lift vessels have been approached in \$loc ? \\
Paraphrased: \textcolor{purple}{How many heavy lift vessels have been contacted in your area?} \\

\noindent\rule{.5\textwidth}{1pt}\\
\\
\caption{Different GPT-3 prompts produce different paraphrases of the same original sentence. The paraphrased sentence in \textcolor{cyan}{blue} is correct, while the paraphrased sentences in \textcolor{purple}{purple} are incorrect.}
\label{tab:gpt3_prompt}
\end{table}

\subsubsection*{Paraphrasing using Fine-tuned Language Model}
We employ a pre-trained language model, BART-Large~\citep{lewis2020bart}, to rewrite a canonical utterance to a more diverse alternative.
The BART-Large model\footnote{We use the official implementation in fairseq, https://github.com/pytorch/fairseq.} is fine-tuned on a corpus of high-quality, diverse paraphrases sampled from PARANMT \citep{wieting2018paranmt} released by \citet{krishna2020reformulating}. Therefore, the model learns to generate paraphrases with diverse linguistic patterns.
\subsubsection*{Paraphrasing using a Commercial System} 
We also experiment with paraphrase generation using Quillbot\footnote{https://quillbot.com}. Quillbot is a commercial tool that provides a scalable and robust paraphraser by controling synonyms and generation styles. 

\subsection{Paraphrase Filtering}
\label{sec:para_filter}
Since the automatically generated paraphrases may have varying quality, we filter out low-quality ones. In our work, we adopt the filtering method discussed in~\citet{xu2020autoqa} in the spirit of self-training. The process consists of the following steps:
\begin{enumerate}
    \item Evaluate the parser on the generated paraphrases and keep those for which the parser correctly generates the corresponding SQL queries.
    \item Add the paraphrase-SQL pairs into the training data and retrain the parser.
    \item Repeat steps 1--2 for several rounds. 
\end{enumerate}

This method is based on three assumptions~\citep{xu2020autoqa}: i) the parser could generalize well to the unseen paraphrases which share the same semantics with the original questions, ii) the seed synthetic dataset generated by the SCFGs is good enough to train an initial model, and iii) it is very unlikely for a poor parser to generate correct SQL queries by chance. 

In the following section, we will provide an experimental analysis of the performance of our proposed approach.


\section{Experiments}

\subsection{Paraphrase Analysis}
\begin{table*}[t!]

\centering
\resizebox{.85\textwidth}{!}{
\begin{tabular}{l|llllll}
\hline
\textbf{\#Rounds}  & BT (Spanish) & BT (Telugu) &  BT (Chinese) & GPT-3 & BART & Quillbot  \\ \hline
Round 1 &  25.15 &  18.52 &\textcolor{cyan}{6.67} &\textcolor{purple}{44.96} & 20.22 & 42.20   \\
Round 2 &    29.63    &   37.85    &  \textcolor{cyan}{13.78}  & 51.81    &   22.34    &    \textcolor{purple}{51.91}     \\
Round 3 &  31.23 &  38.56   &  \textcolor{cyan}{15.42}  &53.22 & 23.35 & \textcolor{purple}{54.25}   \\
\hline
Total &  30.21 &  40.91   &  \textcolor{cyan}{16.37}  & 53.37 & 19.68 & \textcolor{purple}{54.92}   \\
\hline

\end{tabular}
}
\caption{\% of examples kept after each filtering round. The lowest \% in each round is highlighted in \textcolor{cyan}{blue}. The highest \% in each round is highlighted in \textcolor{purple}{purple}.}
\centering
\label{tab:filtering}
\end{table*}
\subsubsection*{Paraphrase Filtering}
\label{sec:filtering}


\tref{tab:filtering} shows the filtering results of several paraphrasing techniques, including back-translation, fine-tuned BART, GPT-3, and Quillbot. In general, we see that the GPT-3 from OpenAI and Quillbot significantly outperform the other approaches. In contrast, after the first filtering round, the fine-tuned BART achieves only 20.22\% retention. Furthermore, we observe that GPT-3 has the highest semantic preservation ability in the first round of filtering, retaining 44.96\% of the paraphrased sentences, whereas Quillbot performs best in the second and third rounds, retaining 51.91\% and 54.25\%, respectively. This indicates that prompt-based and commercial methods for paraphrasing have great potential.
The back-translation method using the Chinese language performs the worst. We hypothesize that the linguistic structure of Chinese is dissimilar to English. Therefore, this would lead to translation errors in the back-translation process. With more retraining rounds, the parser's generalization ability is improved, so more examples are kept. However, the improvement becomes marginal after the second round of retraining. Therefore, we only adopt three rounds since retraining is time-consuming.

\begin{table}[t!]
\centering
\begin{tabular}{l|cccc|c}
\hline 
\textbf{Techniques} & BT & GPT-3 & \textbf{*}BART & Quillbot & All \\ \hline
Time (minutes) & 8.67  & 33.3 & 0.43 & 60 & 102.4\\
\hline
\end{tabular}
\caption{Time comparison per 1000 samples. \\ \textbf{*}The measurement of BART was run on a single RTX 8000 NVIDIA GPU.}
\label{tab:practical}

\end{table}
\begin{table*}[t!]
\centering
\resizebox{.85\textwidth}{!}{
\begin{tabular}{l|llllll}
\toprule
\textbf{Metrics}  & BT (Spanish) & BT (Telugu) &  BT (Chinese) & GPT-3 & BART & Quillbot  \\ \hline
Self-BLEU-4$\downarrow$ & 12.88  & 10.71 & \textcolor{cyan}{6.67} & 7.28 & \textcolor{purple}{36.54} & 7.04    \\
MTLD$\uparrow$ & 18.92  & 18.87 & 18.96 & 19.23 & \textcolor{purple}{16.99} & \textcolor{cyan}{28.80}    \\
TTR$\uparrow$ & 0.49  & 0.53 & \textcolor{cyan}{1.63} & 0.72 & \textcolor{purple}{0.05} & 0.21   \\
\bottomrule

\end{tabular}
}
\caption{Diversity evaluation on paraphrase methods. Symbols $\uparrow$ and $\downarrow$ indicate that higher or lower values, respectively, correlate with greater diversity. The highest diversity for each metric is highlighted in \textcolor{cyan}{blue}, and the lowest diversity for each metric is highlighted in \textcolor{purple}{purple}.}
\centering
\label{tab:diversity}
\end{table*}
\begin{table*}[t!]
\centering
\resizebox{.85\textwidth}{!}{
\begin{tabular}{l|llllll}
\hline
\textbf{Metrics}  & BT (Spanish) & BT (Telugu) &  BT (Chinese) & GPT-3 & BART & Quillbot  \\ \hline
BLEU-1 & 59.05  & 48.97 & \textcolor{cyan}{36.95} & \textcolor{purple}{74.81} & 53.39 & 59.74    \\
BLEU-2 & 43.65  & 36.26 & \textcolor{cyan}{23.89} & \textcolor{purple}{68.42} & 40.32 & 51.22    \\
BLEU-3 & 33.22  & 26.60 & \textcolor{cyan}{15.98} & \textcolor{purple}{63.48} & 31.12 & 44.63   \\
BLEU-4 & 25.63  & 19.33 & \textcolor{cyan}{10.30} & \textcolor{purple}{59.46} & 24.17 & 39.55   \\
\hline

\end{tabular}
}
\caption{Dissimilarity levels between paraphrases and their corresponding clunky forms, measured by BLEU-$N$. The lowest score in each BLEU metric is highlighted in \textcolor{cyan}{blue}. The highest score in each BLEU metric is highlighted in \textcolor{purple}{purple}.}
\centering
\label{tab:disim}
\end{table*}
\subsubsection*{Paraphrase Diversity} 
\label{chap3:para_diversity_metrics}

\paragraph*{Metrics.} 
To begin our evaluation of dataset diversity, we first employ the BLEU-$N$ metric to quantify the level of similarity between each generated paraphrase and its original utterance. BLEU-$N$ is traditionally employed in MT to quantify the similarity between generated output and the references using $N$-grams. Higher BLEU scores generally suggest a greater similarity.

To comprehensively evaluate paraphrase diversity, we employ three metrics: Self-BLEU-$N$~\citep{zhu2018texygen}, MTLD~\citep{mccarthy2005assessment}, and TTR~\citep{templin1957certain}. Self-BLEU-$N$ measures the internal diversity of the dataset by averaging BLEU-$N$~\citep{papineni-etal-2002-bleu} scores, which are generated by comparing pairs of paraphrases within the dataset. MTLD reports the mean length of word strings in the utterances in the dataset that maintain a given Type-token ratio value (TTR). TTR calculates the ratio of unique tokens to the total token count in the dataset.

\paragraph*{Analysis.} 
According to \tref{tab:diversity}, BART generates the least diverse sentences. In contrast, back-translation (Chinese) and Quillbot produce the most diversified paraphrases. Notably, a larger portion of Quillbot's paraphrases are retained, suggesting it excels at both semantic preservation and diversity. An interesting observation, as indicated in ~\tref{tab:disim}, is that BART-generated paraphrases are more dissimilar to the original utterances than those produced via back-translation despite scoring lowest on diversity metrics. This suggests that while BART paraphrases may deviate from the original utterance, they often adhere to similar linguistic patterns. Additionally, paraphrases generated through back-translation (Chinese) demonstrate the lowest retention rates. This might imply that paraphrases with more lexical overlap with the original questions are easier for the parser to generalize.


\subsubsection*{Time-efficiency}
Table \ref{tab:practical} presents the time requirements for each technique to paraphrase 1,000 canonical utterances. Among the methods evaluated, the pre-trained BART model emerges as the most time-efficient approach for paraphrasing, whereas Quillbot ranks as the least time-efficient.






\subsection{Semantic Parser Evaluation}
\subsubsection*{Semantic Parsers}
\label{chap3:sec:parser}
We experiment with two common semantic parsers in the data generation framework: i) the attentional \Seq~\citep{luong2015effective} framework, which uses LSTMs~\citep{hochreiter1997long} as the encoder and decoder, and ii) \bertlstm~\citep{xu2020schema2qa}, a \Seq framework with a copy mechanism~\citep{gu2016incorporating} which uses RoBERTa~\citep{liu2019roberta} as the encoder and LSTM as the decoder. The input to the \Seq model is the NL question, and the output is a sequence of linearized LF tokens.
\subsubsection*{Evaluation Dataset}
We evaluate our semantic parser with real-world crowd-sourced questions. In particular, we collect 231 questions containing both simple and complex compositions regarding a given set of piracy reports. 
We manually write SQL queries for all collected questions and randomly divide them into a validation set with 76 examples and a test with 154.
\begin{table*}[ht]

\centering
\resizebox{1\textwidth}{!}{
\begin{tabular}{l|ccc|ccc}
\hline
\multirow{2}{*}{\textbf{Training Data}} & \multicolumn{3}{c|}{\textbf{SEQ2SEQ}}       & \multicolumn{3}{c}{\textbf{RoBERTa-base}}  \\ \cline{2-7} 
                                        & \textbf{\begin{tabular}[c]{@{}c@{}}Exact match\\ (acc)\end{tabular}} & \textbf{\begin{tabular}[c]{@{}c@{}}Exact match\\ no order (acc)\end{tabular}} & \textbf{\begin{tabular}[c]{@{}c@{}}Compo\_match\\  (F1)\end{tabular}} & \textbf{\begin{tabular}[c]{@{}c@{}}Exact match\\  (acc)\end{tabular}} & \textbf{\begin{tabular}[c]{@{}c@{}}Exact match\\ no order (acc)\end{tabular}} & \textbf{\begin{tabular}[c]{@{}c@{}}Compo\_match\\  (F1)\end{tabular}} \\ \hline
Original dataset  & 36.77  & 48.39  & 70.71 & 43.07   & 57.79 & 77.26 \\
BT (Spanish)  & 37.42 & 50.97  & 75.09  & 42.86  & 57.14 & 77.11 \\
BT (Chinese) & 36.77  & 54.84  & 74.61  & 45.44  & 59.44 & 78.17 \\
BT (Telugu)  & 34.84  & 49.68 & 73.19 & 42.86 & 57.14 & 76.72  \\
BART  & 36.77  & 54.84 & 75.60  & 45.02 & 59.93 & 78.93  \\
GPT-3 & \textbf{43.87}  & \textbf{58.86} & \textbf{76.79} & 47.05 & 61.64  & 79.04 \\
Quilbot & 41.29 & 55.48 & 75.53  & \textbf{50.65} & \textbf{62.99}  & \textbf{80.58} \\
Full Filtered Data  & 40.00  & 56.13 & 76.19 & 46.75 & 60.17 & 78.55 \\ \hline
\end{tabular}
}
\caption{Evaluation of the semantic parser baselines based on additional paraphrased data using different techniques. }
\label{tab:experiment_1}
\centering
\end{table*}
\subsubsection*{Evaluation Metrics}
Our evaluation metrics include Exact Matching as in~\citet{dong2018coarse,li2021few}, Exact Matching (no-order), and Component Matching as in~\citet{yu2018spider}

\paragraph*{Component Matching.} To conduct a detailed
analysis of model performance, we report F1 between the prediction and ground truth in terms of different SQL components. For each of the following components: {\fontfamily{qcr}\selectfont • SELECT • FROM • WHERE • GROUP BY • ORDER BY}.
We decompose each component in the prediction and the ground truth as bags of several sub-components and check whether or not these two sets of components match exactly. In our evaluation, we treat each component as a set. For example, two components {\fontfamily{qcr}\selectfont WHERE va.aggressor = "pirates" AND va.victim = "container ship"} and {\fontfamily{qcr} \selectfont WHERE va.victim = "container ship" AND va.ag-} {{\fontfamily{qcr} \selectfont gressor = "pirates"} would be treated as the same query although the two sub-components in each component are in different order.

\paragraph*{Exact Match.} As described in~\sref{sec:prelim}, Exact Match evaluates whether the string of the predicted query is identical to the one of the ground truth query. 

\paragraph*{Exact Match (no-order).} Exact Match (no-order) is an extension of Exact Match. We first evaluate the SQL component and ignore the order in each component. The predicted query is correct only if all of the components are correct.

\subsubsection*{Analysis} \tref{tab:experiment_1} demonstrates the performance of parsers on various training datasets. For each paraphrasing technique, we have combined the filtered data with the original training dataset to produce a new merged dataset. The semantic parsers are then trained on the merged training dataset. Without any human annotation, the \Seq and \bertlstm can achieve Exact Match (no-order) accuracies of approximately 60\% with the best paraphrasing technique, indicating that our approach could generate a high-quality synthesized dataset. We observe that various paraphrasing techniques affect parsing performance differently, showing that evaluating various paraphrasing techniques in this thesis is necessary. Overall, back-translation techniques even degrade parsing performance, suggesting that the generated paraphrases are of poor quality. The model trained on filtered paraphrases generated by back-translation (Telugu) performs the worst among these back-translation methods. This may result from limited training data when training Google Translation on Telugu, as Telugu is a language with limited resources. We have also discovered that paraphrasing only 10\% of the original training dataset with GPT-3 and Quillbot can consistently improve semantic parsing performance. In terms of Exact Match accuracies, the parsers trained on GPT-3 and Quillbot-paraphrased data outperform those using BART by 2\% to 7\%, indicating that \citet{yin2021ingredients}, who use BART as their paraphrase model, may not make the optimal choice. Our evaluation would be valuable to those using similar data generation approaches. We also demonstrate in \tref{tab:qb_paraphrasing} that the semantic parser's performance improves as more Quillbot paraphrase data is added. In addition, based on the results presented in \tref{tab:filtering} and \tref{tab:diversity}, we conclude that paraphrase filtering and the diversity of the paraphrases are directly related to semantic parsing performance. In general, the paraphrase techniques that generate a more diverse dataset and retain more paraphrases after filtering yield superior parsing performance.

\begin{table}[ht]
\centering
\resizebox{0.7\textwidth}{!}{
\begin{tabular}{l|cc}
\hline
\textbf{Training   Data}  & \multicolumn{1}{c}{\textbf{Filtering}} & \multicolumn{1}{c}{\textbf{Compo\_match (F1)}} \\ \hline
Original training dataset &  -  &  77.26 \\
With Quillbot (10\%) & 54.25   &  80.58  \\
With Quillbot (20\%) & 64.73  & 81.73   \\ \hline
\end{tabular}
}
\caption{The effect of data paraphrasing portion on filtering and parser (RoBERTa-base) LF matching accuracy}
\label{tab:qb_paraphrasing}
\end{table}

\section{Summary}
In this chapter, we develop a semantic parser to improve human-system interaction in the maritime domain when no parallel data are available and only structured data stored in the database exist. By developing a compact set of SCFG rules given the database data and multiple automatic paraphrasing techniques, it is possible to generate massive amounts of synthetic data at a very low cost. The parser trained on such data can also achieve a promising result on the manually annotated test set. This demonstrates that our approach could be useful in many real-world situations where parallel training data is scarce or unavailable.
\chapter{Knowledge Transfer for Few-shot Semantic Parsing}
\epigraph{Knowledge can only be volunteered it cannot be conscripted.}{\textit{Dave Snowden}}

\fbox
{
\begin{minipage}{0.95\textwidth}
     This chapter is based on:\\
     Z. Li, L. Qu, S. Huang, G. Haffari ``Few-shot Semantic Parsing for New Predicates'' The 16th conference of the European Chapter of the Association for Computational Linguistics (EACL). 2021.
\end{minipage}
}


Following the last chapter, this chapter addresses issues for semantic parsers incurred by limited parallel data of the target task, but now assumes the availability of an old task with abundant parallel data.

Specifically, we investigate the problems of semantic parsing in an FSL setting. In this setting, the parser is provided with $k$ utterance-LF pairs per new predicate. The SOTA neural semantic parsers achieve less than 25\% accuracy on benchmark datasets when $k=1$. To tackle this problem, we propose to i) apply a designated meta-learning method to train the model, ii) regularize attention scores with alignment statistics, and iii) apply a smoothing technique in pre-training. As a result, our method consistently outperforms all the baselines in both one and two-shot settings.

\label{chap:meta}
\section{Introduction}
\label{sec:intro}
%
%
Semantic parsing is the task of mapping NL utterances to structured meaning representations, such as LFs. One key obstacle preventing the wide application of semantic parsing is the lack of task-specific training data. New tasks often require new predicates of LFs. Suppose a personal assistant (e.g. Alexa) is capable of booking flights. Due to new business requirements, it needs to book ground transport as well. A user could ask the assistant `\textit{How much does it cost to go from Atlanta downtown to airport?}'. The corresponding LF is as follows:

{\small
\begin{center}
    \begin{tabular}{c r}
    \emph{(lambda \$0 e (exists \$1 (and ( ground\_transport \$1 )  }\\ 
     \emph{(to\_city \$1 atlanta:ci )(from\_airport \$1 atlanta:ci)}\\
     \emph{( =(ground\_fare \$1 ) \$0 ))))} 
    \end{tabular}
\end{center}
}
\noindent where both \textit{ground\_transport} and \textit{ground\_fare} are new predicates while the other predicates are used in flight booking, such as \textit{to\_city}, \textit{from\_airport}.
As manual construction of large parallel training data is expensive and time-consuming, we consider the \textit{few-shot} formulation of the problem, which requires only a handful of utterance-LF training pairs for each new predicate. The cost of preparing few-shot training examples is low. Thus, the corresponding techniques permit significantly faster prototyping and development than supervised approaches for business expansions.

%

%
 An LF template is derived by \emph{normalizing} the entities and attribute values of an LF into typed variable names~\citep{finegan2018improvingtexttosql}. The semantic parsers are poor at cross-template generalization. Current SOTA parsers achieve less than 32\% of accuracy on five widely used corpora when the LFs in the test sets do not share LF \emph{templates} in the training sets~\citep{finegan2018improvingtexttosql}. However, semantic parsing in the few-shot setting is even more challenging. In our experiments, the accuracy of the SOTA semantic parsers drops to less than \textit{25\%}, when there is only one example per new predicate in training data.

The few-shot setting imposes two major challenges for SOTA neural semantic parsers.  
First, it lacks sufficient data to learn effective representations for new predicates in a supervised manner. 
%
%
Second, new predicates bring in new LF templates, which are mixtures of known and new predicates. In contrast, the tasks (e.g. image classification) studied by the prior work on FSL~\citep{snell2017prototypical,finn2017maml} consider an instance exclusively belonging to either a known class or a new class. Thus, it is non-trivial to apply conventional FSL algorithms to generate LFs with mixed types of predicates.

To address above challenges, we present \parser, a transition-based neural semantic parser, which applies a sequence of parse actions to transduce an utterance into an LF template and fills the corresponding slots. The parser is pre-trained on a training set with known predicates, followed by fine-tuning on a \textit{support set} that contains few-shot examples of new predicates. It extends the attention-based sequence-to-sequence architecture~\citep{sutskever2014sequence} with the following novel techniques to alleviate the specific problems in the few-shot setting:
\begin{itemize}
    \item \textit{Predicate-droput}. Predicate-dropout is a meta-learning technique to improve representation learning for both known and new predicates. We empirically found that known predicates are better represented with supervisely learned embeddings, while new predicates are better initialized by a metric-based FSL algorithm~\citep{snell2017prototypical}. In order to let the two types of embeddings work together in a single model, we devised a training procedure called \textit{predicate-dropout} to simulate the testing scenario in pre-training.
    \item \textit{Attention Regularization}. In the few-shot setting, new predicates appear approximately once or twice during training. Thus, it is insufficient to learn reliable attention scores in the Seq2Seq architecture for those predicates. In the spirit of supervised attention~\citep{liu2016nmtSupervisedAttention}, we propose to regularize them with alignment scores estimated by using co-occurrence statistics and string similarity between words and predicates. The prior work on supervised attention\footnote{The description of supervised attention can be found at~\eqref{eq:supervised_att} in~\sref{para:nerual}.} is not applicable because it requires either large parallel data~\citep{liu2016nmtSupervisedAttention}, significant manual effort~\citep{bao2018supervisedAttentionRationales,rabinovich2017semanticParsingSupervisedAttention}, or it is designed only for applications other than semantic parsing~\citep{liu2017eventDetectionSupervisedAttention,kamigaito2017supervisedAttentionConstituencyParsing}.
    \item \textit{Pre-training Smoothing}. The vocabulary of predicates in fine-tuning is higher than that in pre-training, which leads to a distribution discrepancy between the two training stages. Inspired by Laplace smoothing~\citep{manning2008introductionIR}, we achieve significant performance gain by applying a smoothing technique during pre-training to alleviate the discrepancy.
\end{itemize}
Our extensive experiments on three benchmark corpora show that \parser outperforms the competitive baselines by a significant margin. The ablation study demonstrates the effectiveness of each individual proposed technique. The results are statistically significant with p$\le$0.05 according to the Wilcoxon signed-rank test~\citep{wilcoxon1992individual}.

\section{Semantic Parser}
\label{sec:template-general}

\parser follows the SOTA neural semantic parsers~\citep{dong2018coarse,guo2019irnet} to map an utterance into an LF in two steps: \textit{template generation} and \textit{slot filling}\footnote{Code and datasets can be found in this repository: \url{https://github.com/zhuang-li/few-shot-semantic-parsing}}. It implements a designated transition system to generate templates, followed by filling the slot variables with values extracted from utterances. To address the challenges in the few-shot setting, we proposed three training methods, detailed in~\sref{sec:training}.

Many LFs differ only in mentioned atoms, such as entities and attribute values. An LF template is created by replacing the atoms in LFs with typed slot variables. As an example, the LF template of our example in \sref{sec:intro} is created by substituting i) a typed atom variable $v_e$ for the entity ``atlanta:ci''; ii) a shared variable name $v_a$ for all variables ``$\$0$`` and ``$\$1$``.

{\small
\begin{center}
    \begin{tabular}{c r}
    \label{example_template}
    \emph{(lambda $v_a$ e (exists $v_a$ (and ( ground\_transport $v_a$ )  }\\ 
     \emph{(to\_city $v_a$ $v_e$ )(from\_airport $v_a$ $v_e$) ( =(ground\_fare $v_a$ ) $v_a$ ))))} 
    \end{tabular}
\end{center}
}



\begin{table}[t]
{\small
  \begin{center}
\begin{tabular}{l l} 
\hline\hline 
\textbf{t} & \textbf{Actions} \\ 
\hline 
$t_1$ & \gen{(ground\_transport $v_a$)}  \\
$t_2$ &  \gen{(to\_city $v_a$ $v_e$)}  \\
$t_3$ &  \gen{(from\_airport $v_a$ $v_e$)}\\
$t_4$ &  \gen{(= (ground\_fare $v_a$) $v_a$)}  \\
$t_5$ &  \reduce{and :- NT NT NT NT}  \\
$t_6$ &  \reduce{exists :- $v_a$ NT}\\
$t_7$ &  \reduce{lambda :- $v_a$ e NT}\\
\hline 
\end{tabular}
\caption{An example action sequence.}
\label{fig:tree_example}
  \end{center}
 }
\end{table}

\noindent Formally, let $\vx = \{x_1, ... , x_n\}$ denote an NL utterance, and its LF is represented as a semantic tree $\vy = (\mathcal{V}, \mathcal{E})$, where $\mathcal{V} = \{v_1, ... , v_m\}$ denotes the node set with $v_i \in \mathcal{V}$, and $\mathcal{E} \subseteq \mathcal{V} \times \mathcal{V} $ is its edge set. 
The node set $\mathcal{V} = \mathcal{V}_p \cup \mathcal{V}_v$ is further divided into a template predicate set $\mathcal{V}_p$, and a slot value set $\mathcal{V}_v$. A template predicate node represents a predicate symbol or a term, while a slot value node represents an atom mentioned in utterances. Thus, a semantic tree $\vy$ is composed of an abstract tree $\tau_{\vy}$ representing a template and a set of slot value nodes $\mathcal{V}_{v,\vy}$ attaching to the abstract tree.

\begin{figure}
    \centering
    \includegraphics[width=\textwidth]{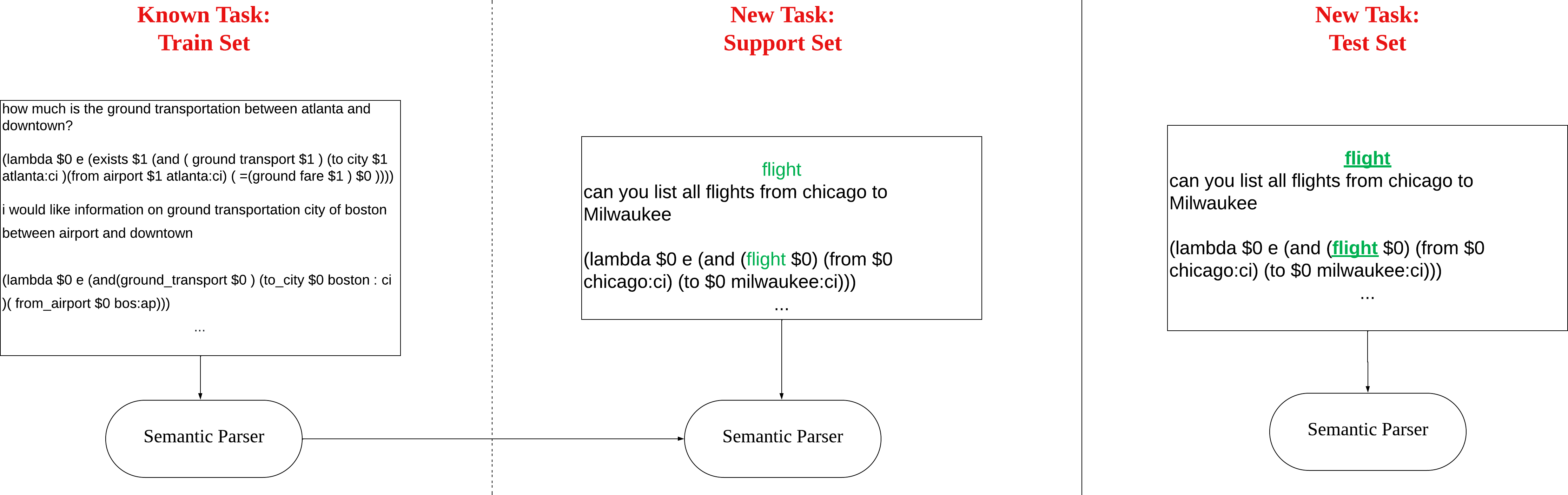}
    \label{fig:semantic_tree}
    \caption{The few-shot setting.}
    \label{fig:few_shot_meta}
\end{figure}

As in~\fref{fig:few_shot_meta}, in the few-shot setting, we are provided with a train set $\mathcal{D}_{train}$, a support set $\mathcal{D}_s$, and a test set $\mathcal{D}_{test}$. Each example in either of those sets is an utterance-LF pair $(\vx_i, \vy_i)$. The new predicates appear only in $\mathcal{D}_s$ and $\mathcal{D}_{test}$ but \textit{not} in $\mathcal{D}_{train}$. For $K$-shot learning, there are $K$ $(\vx_i, \vy_i)$ per each new predicate $p$ in $\mathcal{D}_s$. Each new predicate appears also in the test set. The goal is to maximize the accuracy of estimating LFs given utterances in $\mathcal{D}_{test}$ by using a parser trained on $\mathcal{D}_{train} \cup \mathcal{D}_{s}$. 

\subsection{Transition System}
We apply the transition system~\citep{cheng2019executableParser} to perform a sequence of transition actions to generate the template of a semantic tree. The transition system maintains partially constructed outputs using a \textit{stack}. The parser starts with an empty stack. At each step, it performs one of the following transition actions to update the parsing state and generate a tree node. The process repeats until the stack contains a complete tree. 

\begin{itemize}
    \item \textbf{\gen{$y$}} creates a new leaf node $y$ and pushes it on top of the stack.
    \item \textbf{\reduce{$r$}}. The reduce action identifies an implication rule \textit{ $\text{head} :- \text{body}$}. The rule body is first popped from the stack. A new subtree is formed by attaching the rule head as a new parent node to the rule body. Then, the whole subtree is pushed back to the stack. 
\end{itemize}
\tref{fig:tree_example} shows an action sequence for generating the above LF template. Each action produces \textit{known} or \textit{new} predicates. 

\subsection{Base Parser}
\label{sec:nn_model}


\parser generates an LF in two steps: i) template generation, ii) slot filling. The base architecture largely resembles~\citep{cheng2019executableParser}.

\paragraph*{Template Generation.} Given an utterance, the task is to generate a sequence of actions $\mathbf{a} = a_1, ..., a_k$ to build an abstract tree $\tau_{\vy}$.

We found out that LFs often contain idioms, which are frequent subtrees shared across LF templates. Thus, we apply a \textit{template normalization} procedure in a similar manner as~\citet{iyer2019learningIdiomsSP} to preprocess all LF templates. As in~\fref{fig:template_normalization}, template normalization collapses idioms into single units such that all LF templates are converted into a compact form. Template normalization can enhance the semantic parser's ability to generalize across templates. Please refer to \sref{appendix:meta} for more details on template normalization.
\begin{figure}
    \centering
    \includegraphics[width=1\textwidth]{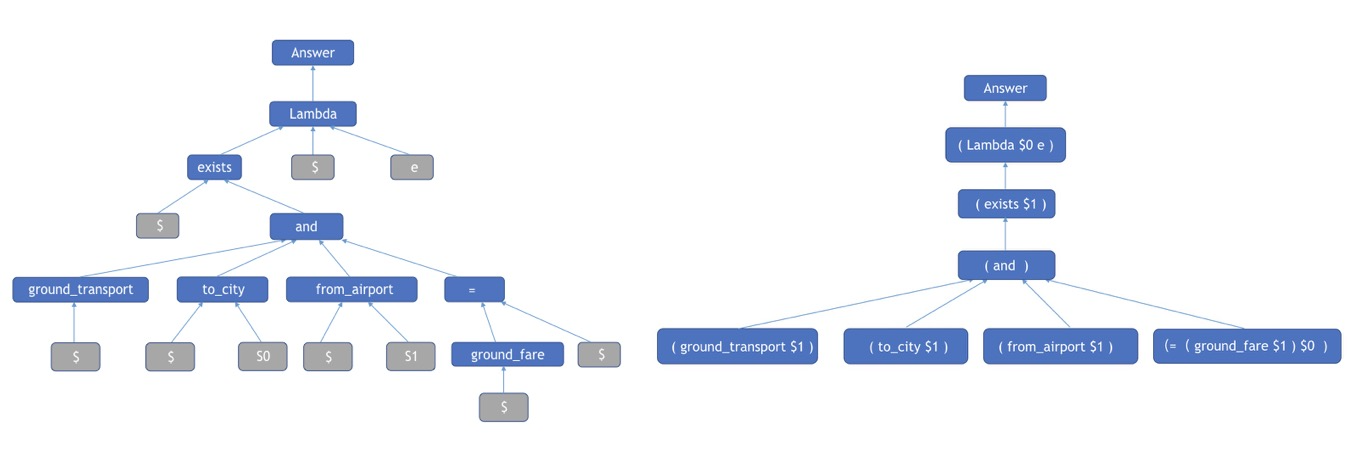}
    \caption{A LF tree (Left) is normalized to its compact form (Right).}
    \label{fig:template_normalization}
\end{figure}

The neural transition system consists of an encoder and a decoder for estimating action probabilities.

\begin{equation}
    P(\mathbf{a} | \vx ) = \prod_{t = 1}^{|\mathbf{a}|} P(a_t | \va_{<t},\vx)
\end{equation}

\textit{Encoder} We apply a bidirectional Long Short-term Memory (LSTM) network~\citep{gers1999LSTM} to map a sequence of $n$ words into a sequence of contextual word representations $\{\mathbf{e}\}_{i=1}^n$. 

\textit{Template Decoder} The decoder applies a stack-LSTM~\citep{dyer2015transition} to generate action sequences.
A stack-LSTM is an unidirectional LSTM augmented with a pointer. The pointer points to a particular hidden state of the LSTM, which represents a particular state of the stack. It moves to a different hidden state to indicate a different state of the stack.

At time $t$, the stack-LSTM produces a hidden state $\mathbf{h}^d_{t}$ by $\mathbf{h}^d_t = \text{LSTM}(\mu_{t}, \mathbf{h}^d_{t-1})$, where $\mu_{t}$ is a concatenation of the embedding of the action $\mathbf{c}_{a_{t-1}}$ estimated at time $t-1$ and the representation $\mathbf{h}_{y_{t-1}}$ of the partial tree generated by history actions at time $t-1$. 

As a common practice, $\mathbf{h}^d_{t}$ is concatenated with an attended representation $\mathbf{h}_t^a$ over encoder hidden states to yield $\mathbf{h}_t$, with
$
    \mathbf{h}_t = \mathbf{W} \begin{bmatrix}
    \mathbf{h}_t^d\\
    \mathbf{h}_t^a
    \end{bmatrix}
$, where $\mathbf{W}$ is a weight matrix and $\mathbf{h}^a_t$ is created by soft attention,
\begin{equation}
\label{eq:attented_rep}
    \mathbf{h}^a_t = \sum_{i=1}^n P(\mathbf{e}_i| \mathbf{h}^d_t)\mathbf{e}_i
\end{equation}
We apply dot product to compute the normalized attention scores $P(\mathbf{e}_i| \mathbf{h}^d_t)$ 
~\citep{luong2015effective}. The \textit{supervised attention}~\citep{rabinovich2017semanticParsingSupervisedAttention,yin2018tranx} is also applied to facilitate the learning of attention weights. Given $\mathbf{h}_t$, the probability of an action is estimated by:
\begin{equation}
\label{eq:action_prob_meta}
P(a_t | \mathbf{h}_t) = \frac{\exp(\mathbf{c}_{a_t}^{\intercal} \mathbf{h}_t)}{\sum_{a' \in \mathcal{A}_t}\exp(\mathbf{c}_{a'}^{\intercal} \mathbf{h}_t)}
\end{equation}
where $\mathbf{c}_{a}$ denotes the embedding of action $a$, and $\mathcal{A}_t$ denotes the set of applicable actions at time $t$. 
The initialization of those embeddings will be explained in the following section.

\paragraph*{Slot Filling.}
As in~\fref{fig:template_normalization}, a tree node in a semantic tree may contain more than one slot variable due to template normalization. Since there are two types of slot variables, given a tree node with slot variables, we employ an LSTM-based decoder with the same architecture for each type of slot variable, respectively. The output of such a decoder is a value sequence of the same length as the number of slot variables of that type in the given tree node.

For each tree node to predict, we set the initial state of the LSTM as the hidden state $\mathbf{h}_t$, which is used to generate the tree node (Eq. \eqref{eq:action_prob}). The first input of the LSTM is always fixed to \textit{$<$START$>$}. In the subsequential steps, an LSTM takes the slot value predicted at the previous time step as input. At each time step $t$, we estimate the probability of the slot value $P(v_t | \mathbf{h}_t, v_{1:t-1})$ by $\text{softmax}(\mathbf{W}_v \mathbf{h}^v_t)$, where $\mathbf{h}_t^v$ denotes the hidden state of the LSTM at time $t$ and $\mathbf{W}_v$ is a embedding matrix for that type of slot variables. The vocabulary of such a decoder is a pre-defined lexicon collected from training datasets. Since we know the number of each slot type in a tree node, the prediction stops when $t$ reaches $T$. Note that the decoding complexity does not exceed the complexity for decoding the entire LF, as the number of variables is always smaller than the length of the LF.

\section{Few-Shot Model Training}
\label{sec:training}
The few-shot setting differs from the supervised setting by having a support set in testing in addition to train/test sets. The support set contains $k$ utterance-LF pairs per new predicate, while the training set contains only known predicates. To evaluate model performance on new predicates, the test set contains LFs with both known and new predicates. Given the support set, we can tell if a predicate is known or new by checking if it only exists in the train set. 

We take two steps to train our model: i) pre-training on the training set, ii) fine-tuning on the support set. Its predictive performance is measured on the test set. 
We take the two-steps approach because i) our experiments show that this approach performs better than training on the union of the train set and the support set; ii) for any new support sets, it is computationally more time efficient than training from scratch on the union of the train set and the support set.


There is a distribution discrepancy between the train set and the support set due to new predicates. The meta-learning algorithms~\citep{snell2017prototypical,finn2017maml} suggest simulating the testing scenario in pre-training by splitting each batch into a meta-support set and a meta-test set. The models utilize the information (e.g. prototype vectors) acquired from the meta-support set to minimize errors on the meta-test set. In this way, the meta-support and meta-test sets simulate the support and test sets sharing new predicates. 

However, we cannot directly apply such a training procedure due to the following two reasons. First, each LF in the support and test sets is a mixture of both known predicates and new predicates. To simulate the support and test sets, the meta-support and meta-test sets should include both types of predicates as well. We cannot assume that there is only one type of predicate.
Second, our preliminary experiments show that if there is sufficient training data, it is better off training action embeddings of known predicates $\mathbf{c}$ (Eq. \eqref{eq:action_prob_meta}) in a supervised way, while action embeddings initialized by a metric-based meta-learning algorithm~\citep{snell2017prototypical} perform better for rarely occurred new predicates. Therefore, we cope with the differences between known and new predicates by using a customized initialization method in fine-tuning and a designated pre-training procedure to mimic fine-tuning on the train set. In the following, we introduce fine-tuning first because it helps understand our pre-training procedure.


\subsection{Fine-tuning}
During fine-tuning, the model parameters and the action embeddings in Eq. \eqref{eq:action_prob_meta} for known predicates are obtained from the pre-trained model. The embedding of actions that produce new predicates  $\mathbf{c}_{a_t}$ are initialized using prototype vectors as in prototypical networks~\citep{snell2017prototypical}.
The prototype representations act as a type of regularization, which shares the similar idea as the deep learning techniques using pre-trained models.

A prototype vector of an action $a_t$ is constructed by using the hidden states of the \textit{template} decoder collected at the time of predicting $a_t$ on a support set. Following~\citet{snell2017prototypical}, a prototype vector is built by taking the mean of such a set of hidden states $\mathbf{h}_t$.
\begin{equation}
    \mathbf{c}_{a_t} = \frac{1}{|M|}\sum_{\mathbf{h}_t \in M} \mathbf{h}_t
    \label{eq:action_embed_proto}
\end{equation}
where $M$ denotes the set of all hidden states at the time of applying the action $a_t$. After initialization, the whole model parameters and the action embeddings are further improved by fine-tuning the model on the support set with a supervised training objective $\mathcal{L}_f$.
\begin{equation}
    \mathcal{L}_f = \mathcal{L}_s + \lambda \Omega
\end{equation}
where $\mathcal{L}_s$ is the cross-entropy loss and $\Omega$ is an attention regularization term explained below. The degree of regularization is adjusted by $\lambda \in \mathbb{R}^+$.

\paragraph*{Attention Regularization.} We address the poorly learned attention scores $P(\mathbf{e}_i| \mathbf{h}^d_t)$ of infrequent actions by introducing a novel attention regularization. We observe that the probability $P(a_j | x_i) = \frac{\text{count}(a_j, x_i)}{\text{count}(x_i)}$ and the character similarity between the predicates generated by action $a_j$ and the token $x_i$ are often strong indicators of their alignment. The indicators can be further strengthened by manually annotating the predicates with their corresponding natural language tokens. In our work, we adopt $1 - dist(a_j, x_i)$ as the character similarity, where $dist(a_j, x_j)$ is normalized Levenshtein distance~\citep{levenshtein1966stringDistance}. Both measures are in the range $[0,1]$, thus we apply $g(a_j, x_i) = \sigma(\cdot)P(a_j | x_i) + (1 - \sigma(\cdot)char\_sim(a_j, x_i)$ to compute alignment scores, where the sigmoid function $\sigma(\mathbf{w}^\intercal_p \mathbf{h}^d_t)$ combines two constant measures into a single score. The corresponding normalized attention scores is given by
\begin{equation}
    P'(x_i | a_k) = \frac{g(a_k, x_i)}{\sum_{j=1}^n g(a_k, x_j)}
\end{equation}
The attention scores $P(x_i | a_k)$ should be similar to $P'(x_i | a_k)$. Thus, we define the regularization term as $\Omega = \sum_{ij} | P(x_i | a_j) - P'(x_i | a_j)|$ during training. 

\subsection{Pre-training}
The pre-training objectives are two-fold: i) learn action embeddings for known predicates in a supervised way, and ii) ensure our model can quickly adapt to the actions of new predicates, whose embeddings are initialized by prototype vectors before fine-tuning.
 
\paragraph*{Predicate-dropout.} Starting with randomly initialized model parameters, we alternately use one batch for the meta-loss $\mathcal{L}_m$ and one batch for optimizing the supervised loss $\mathcal{L}_s$. 

In a batch for $\mathcal{L}_m$, we split the data into a meta-support set and a meta-test set. In order to simulate the existence of new predicates, we randomly select a subset of predicates as "new" thus their action embeddings $\mathbf{c}$ are replaced by prototype vectors constructed by applying Eq. \eqref{eq:action_embed_proto} over the meta-support set. The actions of remaining predicates keep their embeddings learned from previous batches. The resulting action embedding matrix $\mathbf{C}$ is the combination of both.
\begin{equation}
\label{eq:action_embed_meta}
    \mathbf{C} = (1 - \mathbf{m}^\intercal) \mathbf{C}_s + \mathbf{m}^\intercal \mathbf{C}_m
\end{equation}
where $\mathbf{C}_s $ is the embedding matrix learned in a supervised way, and $\mathbf{C}_m$ is constructed by using prototype vectors on the meta-support set. The mask vector $\mathbf{m}$ is generated by setting the indices of actions of the "new" predicates to ones and the other to zeros. We refer to this operation as \textit{predicate-dropout}. The training algorithm for the meta-loss is summarized in \algoref{algo:sup}.

In a batch for $\mathcal{L}_s$, we update the model parameters and all action embeddings with a cross-entropy loss $\mathcal{L}_s$, together with the attention regularization. Thus, the overall training objective becomes
\begin{equation}
   \mathcal{L}_p = \mathcal{L}_m + \mathcal{L}_s + \lambda \Omega
\end{equation}
\paragraph*{Pre-training smoothing.} Due to the new predicates, the number of candidate actions during the prediction of fine-tuning and testing is larger than the one during pre-training. That leads to distribution discrepancy between pre-training and testing. To minimize the differences, we assume a prior knowledge on the number of actions for new predicates by adding a constant $k$ to the denominator of Eq. \eqref{eq:action_prob_meta} when estimating the action probability $P(a_t | \mathbf{h}_t)$ during pre-training. 
\begin{equation}
\label{eq:smooth}
P(a_t | \mathbf{h}_t) = \frac{\exp(\mathbf{c}_{a_t}^{\intercal} \mathbf{h}_t)}{\sum_{a' \in \mathcal{A}_t}\exp(\mathbf{c}_{a'}^{\intercal} \mathbf{h}_t) + k}
\end{equation}
We do not consider this smoothing technique during fine-tuning and testing.  
Despite its simplicity, the experimental results show a significant performance gain on benchmark datasets.
\begin{algorithm}[t]
{\small
\SetKwData{Left}{left}\SetKwData{This}{this}\SetKwData{Up}{up}
\SetKwFunction{Union}{Union}\SetKwFunction{FindCompress}{FindCompress}
\SetKwInOut{Input}{Input}\SetKwInOut{Output}{Output}
\SetAlgoLined
\Input{Training set $\mathcal{D}$, supervisely trained action embedding $\mathcal{C}_s$, number of meta-support examples $k$, number of meta-test examples $n$ per one support example, predicate-dropout ratio $r$}
\Output{The loss $\mathcal{L}_m$.}
    Extract a template set $\mathcal{T}$ from the training set $\mathcal{D}$\\
    Sample a subset $\mathcal{T}_i$ of size $k$ from $\mathcal{T}$\\ 
    $S$ := $\emptyset$ \textcolor{blue}{\# \textit{meta-support set}}\\ 
    $Q$ := $\emptyset$  \textcolor{blue}{\# \textit{meta-test set}}\\
    \For{t in  $\mathcal{T}_i$}{
    Sample a meta-support example $s'$ with template $t$ from $\mathcal{D}$ without replacement\\ 
    Sample a meta-test set $Q'$ of size $n$ with template $t$ from $\mathcal{D}$ \\
    $S = S \cup s'$\\
    $Q = Q \cup Q'$\\
    }
    Build a prototype matrix $\mathcal{C}_m$ on $S$\\
    Extract a predicate set $\mathcal{P}$ from $S$\\ 
    Sample a subset $\mathcal{P}_s$ of size $r\times |\mathcal{P}|$ from $\mathcal{P}$ as new predicates \\ 
    Build a mask $\mathbf{m}$ using $\mathcal{P}_s$\\
    With $\mathcal{C}_s$, $\mathcal{C}_m$ and $\mathbf{m}$, apply Eq. \eqref{eq:action_embed_meta} to compute $\mathbf{C}$\\
    Compute $\mathcal{L}_m$, the cross-entropy on $Q$ with $\mathbf{C}$\\

}
\caption{Predicate-Dropout
}
\label{algo:sup}
\end{algorithm}





\section{Experiments}
\paragraph*{Datasets.}
We use three semantic parsing datasets: \Jobs, \Geo, and \Atis. 
\Jobs contains 640 question-LF pairs in Prolog about job listings. \geoquery~\citep{zelle1996geoQuery} and \Atis~\citep{price1990atis} include 880 and 5,410 utterance-LF pairs in lambda calculus about US geography and flight booking, respectively. The number of predicates in \Jobs, \geoquery, \Atis is 15, 24, and 88, respectively. All atoms in the datasets are anonymized as in~\citet{dong2016language}.

For each dataset, we randomly selected $m$ predicates as the new predicates, which is 3 for \Jobs, and 5 for \Geo and \Atis. Then we split each dataset into a train set and an \textit{evaluation} set. And we removed the instances, the template of which is unique in each dataset. The number of such instances is around 100, 150 and 600 in \Jobs, \Geo, and \Atis. The ratios between the evaluation set and the train set are 1:4, 2:5, and 1:7 in \jobs, \geoquery, and \atis, respectively. Each LF in an evaluation set contains at least a new predicate, while an LF in a train set contains only known predicates. To evaluate $k$-shot learning, we build a support set by randomly sampling $k$ pairs per new predicate without replacement from an \textit{evaluation} set, and keep the remaining pairs as the test set. To avoid evaluation bias caused by randomness, we repeat the above process six times to build six different splits of support and test sets from each evaluation set. One for hyperparameter tuning and the rest for evaluation. We consider, at most, 2-shot learning due to the limited number of instances per new predicate in each evaluation set.

\paragraph*{Implementation Details.}
\label{para:training}
We pre-train our parser on the training sets for \{80, 100\} epochs with the Adam optimizer~\citep{kingma2014adam}. The batch size is fixed to 64. The initial learning rate is 0.0025, and the weights are decayed after 20 epochs with a decay rate of 0.985. The predicate dropout rate is 0.5. The smoothing term is set to \{3, 6\}. The number of meta-support examples is 30, and the number of meta-test examples per support example is 15.
The coefficient of attention regularization is set to 0.01 on \jobs and 1 on the other datasets. 
We employ the 200-dimensional GLOVE embedding~\citep{pennington2014glove} 
to initialize the word embeddings for utterances. The hidden state size of all LSTM models~\citep{hochreiter1997long} is 256. 
During fine-tuning, the batch size is 2, and the learning rates and the epochs are selected from \{0.001, 0.0005\} and \{20, 30, 40, 60, 120\}, respectively. 



\begin{table*}[ht]
\small
\begin{tabular}[t]{@{} |l|ccc||ccc|l|}
\toprule
         & \Jobs & \Geo & \Atis & \Jobs & \Geo & \Atis & p-values\\ 
\midrule
\hline         
\seqseq (\pt)  & 11.27  &  20.00      &   17.23  & 14.58   &  33.01     &   18.76 & 3.32e-04 \\
\seqseq (\comb)  & 11.70   &  7.64     &   2.25  & 21.49   &  14.36      &   7.91 & 6.65e-06 \\
\seqseq (\os)  & 14.18   &  11.38      &   4.45  & 30.46   &  33.59     &   10.17 & 5.30e-05\\
\hline
\coarsefine (\pt) & 10.91   & 24.07       &  17.44  & 13.83  & 35.63       &  21.08 &  1.48e-04 \\
\coarsefine (\comb) & 9.28  & 14.50     &  0.42  & 19.61   & 28.93       &  9.25 & 2.35e-06 \\
\coarsefine (\os) & 6.73   & 10.35        &  5.26  & 16.08   & 28.55      &  17.73 & 1.13e-05  \\
\hline
\irnet (\pt) & 16.00   & 20.00     &  17.12  & 19.06  & 35.05        &  20.11 & 2.86e-05 \\
\irnet (\comb) & 19.67   & 21.90     &  5.60  & 28.22   & 44.08      &  15.73 &2.76e-03 \\
\irnet (\os) & 14.91  & 18.78      &  4.95 & 30.84   & 40.97       &  18.05 & 2.47e-04 \\
\hline
\da     & 18.91  & 9.67      &  4.29    & 21.31 & 20.88  &   17.18 & 1.13e-06 \\
\hline
\ptmaml     & 11.64  & 9.76     &  6.83    & 17.76  & 22.52        &   12.28  & 1.73e-06 \\
\hline
\parser  & \textbf{27.09}  & \textbf{27.49}        & \textbf{19.27}    & \textbf{32.5} & \textbf{48.45}        & \textbf{22.48}  &  \\
\bottomrule

\end{tabular}%
\caption{Evaluation of learning results on three datasets. (Left) The one-shot results. (Right) The two-shot results.}

\label{table:meta}
\vspace{-3mm}
\end{table*}

\paragraph*{Baselines.}

We compared our methods with five competitive baselines, \seqseq with attention~\citep{luong2015effective}, 
\coarsefine~\citep{dong2018coarse}, \irnet~\citep{guo2019irnet}, \ptmaml~\citep{huang2018metaLearning} and \da~\citep{li2020domain} in the transfer learning and meta-learning settings. \coarsefine, as \proto, adopts a two-stage decoding process. 
\irnet is a strong semantic parser that can generalize to unseen database schemas. In our case, we consider a list of predicates in support sets as the columns of a new database schema and incorporate the schema encoding module of \irnet into the encoder of our base parser. We choose \irnet over \ratsql~\citep{wang2019rat} because \irnet achieves superior performance on our datasets. \ptmaml is an FSL semantic parser that adopts Model-Agnostic Meta-Learning~\citep{finn2017maml}. We adapt \ptmaml in our scenario by considering a group of instances that share the same template as a pseudo-task. \da is the neural semantic parser applying transfer learning techniques.


We consider three different learning settings for \seqseq with attention, 
\coarsefine, \irnet. First, we pre-train a model on a training set, followed by fine-tuning it on the corresponding support set, coined \textit{pt}, as the sequential transfer learning. Second, a model is trained on the combination of a train set and a support set, coined \textit{cb}, as the joint training. Third, the support set in \textit{cb} is oversampled by 10 times and 5 times for one-shot and two-shot, respectively, coined \textit{os}. 
\paragraph*{Evaluation Details.}
The same as prior work~\citep{dong2018coarse,li2020domain}, we report the accuracy of exactly matched LFs as the main evaluation metric. 

To investigate if the results are statistically significant, we conducted the Wilcoxon signed-rank test, which assesses whether 
our model consistently performs better than another baseline across \textit{all} evaluation sets. It is considered superior to the t-test in our case because it supports comparison across different support sets and does not assume normality in data~\citep{demvsar2006statistical}. We include the corresponding $p$-values in our result tables.

\subsection{Results and Discussion}

\tref{table:meta} shows the average accuracies and significance test results
of all parsers compared on all three datasets. Overall, \parser outperforms all baselines with at least 2\% on average in terms of accuracy in both one-shot and two-shot settings. The results are statistically significant w.r.t. the strongest baselines, \irnet(\comb) and \coarsefine(\pt). The corresponding p-values are 0.00276 and 0.000148, respectively. Given a one-shot example on \jobs,
our parser achieves 7\% higher accuracy than the best baseline, and the gap is 4\% 
on \Geo with two-shot examples. In addition, none of the SOTA baseline parsers can consistently outperform other SOTA parsers when there are few parallel data for new predicates. In a one-shot setting, the best baseline \irnet(\comb) can achieve the best results on \geoquery and \jobs among all baselines, and in a two-shot setting, it performs best only on \geoquery.
It is also difficult to achieve good performance by adapting the existing meta-learning or transfer learning algorithms to our problem, as evident by the moderate performance of \ptmaml and \da on all datasets.

The problems of FSL demonstrate the challenges imposed by infrequent predicates. There are significant proportions of infrequent predicates on the existing datasets. For example, on \geoquery, there are 10 predicates contributing to only 4\% of the total frequency of all 24 predicates, while the top two frequent predicates amount to 42\%. As a result, the SOTA parsers achieve merely less than 25\% and 44\% of accuracy with one-shot and two-shot examples, respectively. In contrast, those parsers achieve more than 84\% accuracy on the standard splits of the same datasets in the supervised setting.


Infrequent predicates in semantic parsing can also be viewed as a class imbalance problem, when support sets and train sets are combined in a certain manner. In this work, the ratio between the support set and the train set in \jobs, \geoquery, and \atis is 1:130, 1:100, and 1:1000,  respectively. Different models prefer different ways of using the train sets and support sets. The best option for \coarsefine and \seqseq is to pre-train on a train set followed by fine-tuning on the corresponding support set, while \irnet favors oversampling in a two-shot setting.

\paragraph*{Ablation Study.}
\begin{table*}[ht]
\small
\begin{tabular}{|@{} |l|ccc||ccc|l|}
\toprule
         & \Jobs & \Geo & \Atis  & \Jobs & \Geo & \Atis & p-values\\ 
\midrule
Ours  &27.09  & \textbf{27.49}         & \textbf{19.27}  & \textbf{32.50} & \textbf{48.45}       & \textbf{22.48}  &  \\
\hline
\quad - sup  & 23.63 & 18.86  & 12.91 & 26.91 & 39.51  & 14.89 & 1.44e-05 \\

\quad -  proto  & 22.91 & 18.77  & 13.24  & 29.16  & 38.93   & 16.81 & 1.77e-05 \\
\hline
\quad -  reg  & \textbf{29.27 } & 18.10 & 13.66  & 31.03 & 39.61  & 18.58 & 9.60e-04 \\

\quad -  strsim  & 22.18 & 19.62  & 10.14  & 28.41  & 47.09  & 19.98 & 9.27e-04 \\

\quad -  cond  & 23.27 & 19.05  & 9.63  & 27.66  & 40.97  & 17.50 & 4.37e-05 \\
\hline

\quad - smooth  & 24.36  & 23.60       & 15.23  & 30.84   & 44.95        & 18.71  & 3.27e-03  \\
\bottomrule
\end{tabular}%
\caption{Ablation study results. (Left) The one-shot learning results. (Right) The two-shot learning results.}
\label{table:ablation}
\end{table*}


We examine the effect of different components of our parser by removing each of them individually and reporting the corresponding average accuracy. As shown in \tref{table:ablation}, removing any of the components almost always leads to a statistically significant drop in performance. The corresponding p-values are all less than 0.00327.

To investigate predicate-dropout, we exclude either supervised-loss during pre-training (-sup) or initialization of new predicate embeddings by prototype vectors before fine-tuning (-proto). It is clear from \tref{table:ablation} that ablating either supervisely trained action embeddings or prototype vectors hurts performance severely.  

We further study the efficacy of attention regularization by removing it completely (-reg), removing only the string similarity feature (-strsim), or conditional probability feature (-cond). Removing the regularization completely degrades performance sharply except on \jobs in the one-shot setting. Our further inspection shows that model learning is easier on \jobs than on the other two datasets. Each predicate in \jobs almost always aligns to the same word across examples, while a predicate can align with different word/phrase in different examples in \geoquery and \atis. The performance drop with -strsim and -cond indicates that we cannot only reply on a single statistical measure for regularization. For instance, we cannot always find predicates take the same string form as the corresponding words in input utterances. In fact, the proportion of predicates present in input utterances is only 42\%, 38\% and 44\% on \jobs, \atis, and \geoquery, respectively.

Furthermore, without pre-training smoothing (-smooth), the accuracy drops at least 1.6\% in terms of mean accuracy on all datasets. Smoothing enables better model parameter training by more accurate modelling in pre-training.


\paragraph*{Support Set Analysis.}
\begin{figure}[t] 
\centering
\includegraphics[width=0.75\textwidth]{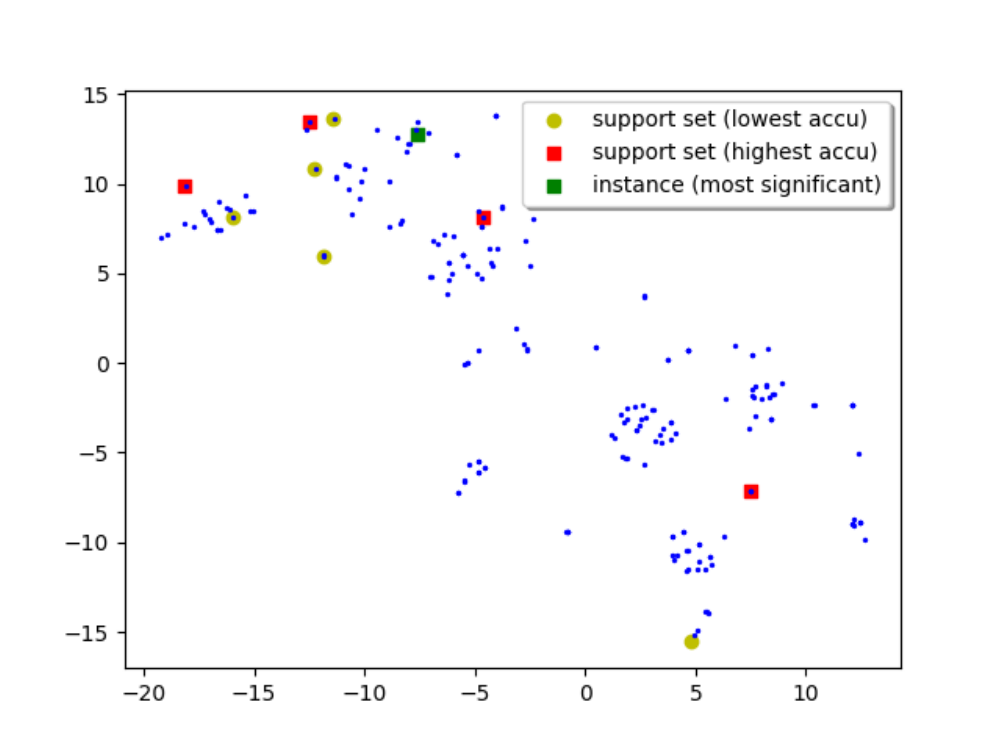} 
\caption{(Round) The support set with the lowest accuracy. (Box) The support set with the highest accuracy.}
\label{fig:utter_encode}
\vspace{-6mm}
\end{figure}
We observe that all models consistently achieve high accuracy on certain support sets of the same dataset, while obtaining low accuracies on the other ones. 
We illustrate the reasons of such effects by plotting the evaluation set of \geoquery. Each data point in \fref{fig:utter_encode} depicts an representation, which is generated by the encoder of our parser after pre-training. We applied T-SNE~\citep{maaten2008visualizing} for dimension reduction. We highlight two support sets used in the one-shot setting on \geoquery.
All examples in the highest-performing support set tend to scatter evenly and cover different dense regions in the feature space, while the examples in the lowest-performing support set are far from a significant number of dense regions. Thus, the examples in good support sets are more representative of the underlying distribution than the ones in poor support sets. When we leave out each example in the highest performing support set and re-evaluate our parser each time, we observe that the good ones (e.g. the green box in \fref{fig:utter_encode}) locate either in or close to some of the dense regions.

\section{Summary}
This chapter presents a novel FSL-based semantic parser, coined \parser, to cope with new predicates in LFs. To address the challenges in FSL, we propose to train the parser with a pre-training procedure involving predicate-dropout, attention regularization, and pre-training smoothing. The resulting model achieves superior results over competitive baselines on three benchmark datasets.
\part{Limited Annotation Budget for Multilingual Semantic Parsing}
\label{part2}
\chapter{Active Learning for Multilingual Semantic Parsing}
\epigraph{Knowledge travels via language.}{\textit{Verna Allee}}

\fbox
{
\begin{minipage}{0.95\textwidth}
     Subsequent to PhD enrollment, research findings from~\cref{chap:al_msp}, along with additional new results, have been submitted for publication and appear in the paper:\\
     Z. Li, G. Haffari ``Active Learning for Multilingual Semantic Parsing'' Findings of the 17th conference of the European Chapter of the Association for Computational Linguistics (EACL). 2023.
\end{minipage}
}

\label{chap:al_msp}

Part~\ref{part1} answered the questions related to how to adapt a parser to a new task/domain with limited parallel task/domain-specific data. Part ~\ref{part2} will mainly address the problem incurred by the low annotation budget while adapting a parser to a new language. Specifically, this chapter addresses RQ3: `How can a semantic parser be adapted to a new language when the annotation budget for manual translation is limited?'

A dataset in the target language is typically required to adapt a parser to the target language. Current MSP datasets are almost all collected by translating the utterances in the existing datasets from the resource-rich language to the target language. However, manual translation is costly. As described in~\sref{sec:pool_al}, AL is a wise choice to reduce the manual translation cost while there are no current AL methods for MSP. We propose the first active learning procedure for MSP (AL-MSP) to reduce the translation effort. AL-MSP selects only a subset from the existing datasets to be translated. We also propose a novel selection method that prioritizes the examples, diversifying the LF structures with more lexical choices, and a novel hyperparameter tuning method that needs no extra annotation cost. Our experiments show that AL-MSP significantly reduces translation costs with ideal selection methods. Our selection method with proper hyperparameters yields better parsing performance than the other baselines on two multilingual datasets.

\section{Introduction}

MSP converts multilingual NL utterances into LFs using a single model. However, there is a severe data imbalance among the MSP datasets. Currently, most semantic parsing datasets are in English, while only a few non-English datasets exist.
To adapt the parser to new languages, almost all current efforts build MSP datasets by translating utterances in the existing datasets from the resource-rich language (e.g. English) into other languages~\citep{duong2017multilingual,li2021mtop}. However, manual translation is slow and laborious. AL is an excellent solution to lower the translation cost in such cases.

AL is a family of methods that collects training data when the annotation budgets are limited~\citep{lewis1994heterogeneous}. This chapter proposes the \textit{first} AL approach for MSP. Compared to translating the full dataset, AL-MSP aims to select only a subset from the existing dataset to be translated, which significantly reduces the translation cost.

We further study which examples AL-MSP should select to optimize multilingual parsing performance in the target languages. \citet{oren2021finding} demonstrated that a training set with diverse LF structures significantly enhances the compositional generalization of the parsers. Furthermore, our experiments show that the examples with LFs aligned with more diversified lexical variants in the training set considerably improve the performance of multilingual parsing during AL. Motivated by both, we propose a novel strategy for selecting the instances that include diversified LF structures with more lexical choices. Our selection method yields better parsing performance than the other baselines. By translating just 32\% of all examples, the parser achieves comparable performance on multilingual \geo and \nlmap as translating full datasets.


The prior study~\citep{duong2018active} applies AL methods to monolingual semantic parsing, which is similar to our setting. They determine the hyperparameters of the AL methods by either i) copying configurations from comparable settings or ii) tuning the hyperparameters using seed evaluation data. However, the first approach is unsuitable for our approach because our AL setting is designed for a multilingual context, which is distinct from theirs. Meanwhile, the second approach incurs additional annotation costs, as the seed data also requires manual annotation. This chapter introduces a cost-free method for obtaining optimal hyperparameters in our AL scenario.

Our contributions are:
\begin{itemize}
    \item The first AL procedure for MSP that reduces the manual translation effort.
    \item An approach that selects examples for superior parsing performance in the target languages.
    \item A hyperparameter tuning method for the selection that does not incur any extra annotation costs.
\end{itemize}

\begin{figure}
    \centering
    \includegraphics[width=1\textwidth]{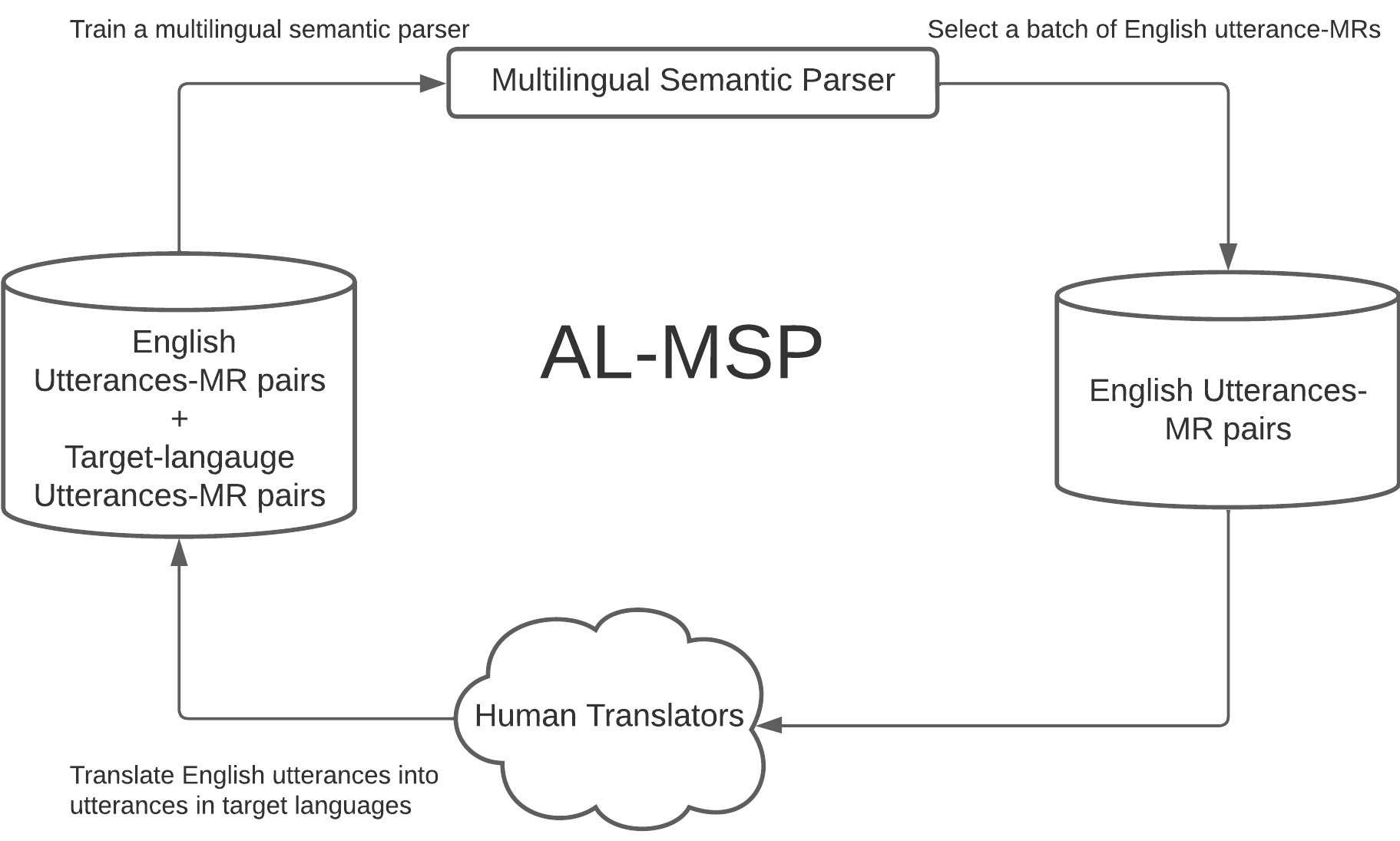}
    \caption{The AL procedure for MSP (AL-MSP).}
    \label{fig:seq2seq_exp}
\end{figure}

\section{Active Learning Methods}
AL-MSP considers only a bilingual scenario for the proof of concept while extending our AL method to more target languages is easy. The goal of AL-MSP is to minimize the human effort in translating utterances while the semantic parser can still achieve a certain level of performance on the bilingual test sets. Starting from a semantic parser initially trained on the dataset $\mathcal{D}_s = \{(\vx^{s}_i, \vy_i)\}^{N}_{i=1}$ whose utterances are in the high-resource language $s$, AL-MSP selects $K_q$ examples $\tilde{\mathcal{D}}_s = \{(\vx^{s}_i, \vy_i)\}^{K_q}_{i=1}$ from $\mathcal{D}_s$, followed by manually translating the utterances in $\tilde{\mathcal{D}}_s$ into a target language $t$, denoted by $\tilde{\mathcal{D}}_t = t_{s\rightarrow t}(\tilde{\mathcal{D}}_s)$, where $\tilde{\mathcal{D}}_t = \{(\vx^{t}_i, \vy_i)\}^{K_q}_{i=1}$. The selection criterion is based on our proposed \textit{acquisition} function $\phi(\ve_s)$ scoring each example, $\ve_s = (\vx_s, \vy)$. The parser is retrained on the union of $\tilde{\mathcal{D}}_t$ and $\mathcal{D}_s$. There will be $Q$ iterations of selection and retraining until the retrained parser reaches a good performance on the bilingual test sets $\mathcal{D}^{test}_{s}$ and $\mathcal{D}^{test}_{t}$. \algoref{algo:al_msp} describes our experimental settings in detail.

\begin{algorithm}[ht]
{\small
\SetKwData{Left}{left}\SetKwData{This}{this}\SetKwData{Up}{up}
\SetKwFunction{Union}{Union}\SetKwFunction{FindCompress}{FindCompress}
\SetKwInOut{Input}{Input}\SetKwInOut{Output}{Output}
\SetAlgoLined
\Input{Initial training set $\mathcal{D}^{0} = \mathcal{D}^{0}_s$, budget size $K_q$, number of the selection rounds $Q$}
\Output{A well-learned multilingual parser $P_{\bm{\theta}}(\vy|\vx)$}
Train the parser $P_{\bm{\theta}}(\vy|\vx)$ on the training set $\mathcal{D}^{0}$ \\
Evaluate parser performance on test sets $\mathcal{D}^{test}_{s}$, $\mathcal{D}^{test}_{t}$ \\
\For{$q \gets 1$ to $Q$} 
{
Estimate the acquisition $\phi(\cdot)$ \\
Select a subset $\tilde{\mathcal{D}}^{q}_s \in \mathcal{D}^{q-1}_s$ of the size $K_q$ based on the acquisition function $\phi(\cdot)$ \\
Translate the utterances in $\tilde{\mathcal{D}}^{q}_s$ into the target language, $\tilde{\mathcal{D}}^{q}_t = t_{s\rightarrow t}(\tilde{\mathcal{D}}^{q}_s)$.\\
Combine the training sets, $\mathcal{D}^{q}=\mathcal{D}^{q-1} \cup \tilde{\mathcal{D}}^{q}_t$\\
Exclude the selected examples $\tilde{\mathcal{D}}^{q}_s$ from $\mathcal{D}^{q}_s = \mathcal{D}^{q-1}_s \setminus \tilde{\mathcal{\mathcal{D}}}^{q}_s$ \\
Re-train the parser $P_{\bm{\theta}}(\vy|\vx)$ on $\mathcal{D}^{q}$ \\
Evaluate parser performance on test sets $\mathcal{D}^{test}_{s}$, $\mathcal{D}^{test}_{t}$ 
}
}
\caption{AL-MSP
}
\label{algo:al_msp}
\end{algorithm}
\vspace{-3mm}
\subsection{Selection Acquisition}

Our selection strategy selects the untranslated examples, which maximize the acquisition scores. The acquisition comprises two individual terms, LF Structure Diversity and Lexical Choice Diversity.

\paragraph*{LF Structure Diversity (LFSD).}
\citet{oren2021finding,li2021total} showed that increasing the diversity of the LF sub-structures, atoms and compounds\footnote{For the definitions of atoms and compounds, please refer to the description of LFs in~\sref{sec:parser_framework}.}, in the training set improves the parser's compositional generalization ability. In light of this, we give a simple technique to diversify the atoms and compounds in the instances. At $q$th iteration, let $\mathcal{D}^l_{s}=\bigcup^{q-1}_{i=1} \tilde{\mathcal{D}}^{i}_{s}$ denotes all the translated examples and $\mathcal{D}^u_{s}=\mathcal{D}^{q-1}_{s}$ be the untranslated ones. We partition their union $\mathcal{D}^u_{s} \bigcup \mathcal{D}^l_{s}$ into $|\mathcal{D}^l_{s}| + K_q$ clusters with Incremental K-means~\citep{liu2020active}. Each example $\ve_s = (\vx_s, \vy)$ is featurized by extracting all the atoms and compounds in the LF tree $\tau_{\vy}$, followed by calculating the TF-IDF~\citep{salton1986introduction} value for each atom and compound. Incremental K-means considers each example of $\mathcal{D}^l_{s}$ as a fixed clustering centroids and estimates $K_q$ new cluster centroids. For each of the $K_q$ new clusters, we select one example closest to the centroid. 


Such selection strategy is reformulated as selecting $K_q$ examples with the highest acquisition scores one by one at each iteration:

\begin{small}
\begin{align}
\label{acq:lcs}
    \phi_{s}(\ve_s) =     
    \begin{cases}
      -||f(\vy) - \vc_{m(\vy)}||^2 & \text{if $m(\vy) \notin \bigcup\limits_{\ve_s \in D^{l}_s} m(\vy)$}\\
      -\infty & \text{Otherwise}
    \end{cases}
\end{align}
\end{small}

\noindent where $f(\cdot)$ is the feature function, $m(\cdot)$ maps each LF into its cluster id and $\vc_i$ is the center embedding of the cluster $i$. Since we use batch-wise AL, we estimate the clusters once per iteration to save the estimation cost. As in Eq.~\eqref{acq:lcs}, when a new example is chosen, none of its cluster mates will be selected again. The incremental mechanism guarantees the newly selected examples are structurally different from those chosen in previous iterations. 

To illustrate, consider the selected LF ``\textit{( lambda \$0 e ( and ( state:t \$0 ) ( next\_to:t \$0 s0 ) ) )}''. . This LF includes the atoms such as ``\textit{lambda}'', ``\textit{\$0}'', ``\textit{e}'', as well as compounds such as ``\textit{( state:t \$0 )}'', ``\textit{( next\_to:t \$0 s0 )}'', ``\textit{( and state:t next\_to:t )}'', and ``\textit{( lambda \$0 e and )}''. Another LF, ``\textit{( lambda \$0 e ( state:t \$0 ) )}'', shares several atoms and compounds with the first LF. The second LF will not be selected in subsequent iterations because they are highly similar and belong to the same cluster. 

\paragraph*{Lexical Choice Diversity (LCD).} \lcd aims to select examples whose LFs are aligned with the most diversified lexicons. We achieve this goal by choosing the example maximizing the average entropy of the conditional probability $p(x_s|a)$:

\begin{small}
\begin{align}
     \phi_{c}(\ve_s) &= 
     -\frac{1}{|\mathcal{A}_{\vy}|}\sum_{a \in \mathcal{A}_{\vy}}\lambda_{a}\sum_{x_s \in \mathcal{V}_{s}} p(x_{s}|a) \log p(x_{s}|a) \\
          \lambda_{a} & =     
    \begin{cases}
      1 & \text{if $a \in \mathcal{A}_{l}$}\\
      \beta & \text{Otherwise, $0\leq\beta<1$}
    \end{cases}
\end{align}
\end{small}

\noindent {where $a$ is the atom/compound, $\mathcal{A}_{\vy}$ is the set of all atoms/compounds extracted from $\vy$, $\mathcal{V}_s$ is the vocabulary of the source language,  $\mathcal{A}_{l}$ is the set of atoms/compounds in all selected examples until now, and $p(x_{s}|a)$ is constructed by counting the co-occurrence of $a$ and $x_{s}$ in the source-language training set, $p(x_{s}|a) = \frac{\text{count}(x_s, a)}{\text{count}(a)}$.
To prevent selecting structurally similar LFs, the score of each selected atom or compound is penalized by a decay weight $\beta$.} 

{Our intuition has two premises. First, the parser trained on example pairs whose LFs have more lexical choices generalizes better. Second, LFs with more source-language lexical choices will have more target-language lexical choices as well.}

\paragraph*{LF Structure and Lexical Choice Diversity (LFS-LC-D).}
We eventually aggregate the two terms to get their joint benefits, $\phi(\vx_s, \vy) = \alpha \phi_s(\vx_s, \vy) + \phi_c(\vx_s, \vy)$, where $\alpha$ is the weight that balances the importance of two terms. We normalize the two terms using quantile normalization~\citep{bolstad2003comparison} in order to conveniently tune $\alpha$.

\paragraph*{Hyperparameter Tuning.} {Because our setup is unique, we can not copy hyperparameters from existing works. The other efforts~\citep{duong2018active} get hyperparameters by evaluating algorithms on seed annotated data. To tune our AL hyperparameters, $\alpha$ and $\beta$, a straightforward practice using seed data is to sample multiple sets of examples from the source-language data, the target-language counterparts of which are in seed data, by varying different hyperparameter configurations and reveal their translations in the target language, respectively. The parser is trained on different bilingual datasets and evaluated on the \textit{target-side} dev set. We use the one, which results in the best parsing performance, as the experimental configuration.}

{Such a method still requires translation costs on the seed data. We assume if the selected examples help the parser generalize well in parsing source-language utterances, their translations should benefit the parser in parsing target languages. Given this assumption, we propose a \textit{novel} cost-free hyperparameter tuning approach. First, we acquire different sets of source-language samples by varying hyperparameters. Then, we train the parser on each subset and evaluate the parser on the \textit{source-side} dev set. Finally, we use the hyperparameters with the best dev set performance.} 

\section{Experiments}
\label{sec:exp}
\paragraph*{Datasets.}
Unlike~\cref{chap:meta}, where we utilized monolingual datasets, in this chapter, we experiment with two multilingual datasets: \geo~\citep{susanto2017neural} and \nlmap~\citep{haas2016}, as they offer an extensive collection of manually translated utterances across multiple languages. \geo utterances are in English (EN), German (DE), Thai (TH), and Greek (EL); \nlmap utterances are in English and German. Neither corpora include a development set, so we use 20\% of the training sets of \geo and \nlmap in each language as the development sets for tuning the hyperparameters. To simulate AL process, we consider English as the resource-rich language and others as the target languages. After the examples are selected from the English datasets, we reveal their translations in the target languages and add them to the training sets.

\paragraph*{AL Setting.} We perform six iterations, accumulatively selecting 1\%,  2\%, 4\%, 8\%, 16\% and 32\% of examples from English \geo and \nlmap.

\paragraph*{Baselines.} We compare four selection baselines and the Oracle setting:

\begin{itemize}
    \item \textit{Random} picks English utterances randomly to be translated.
    \item  \textit{S2S (FW)}~\citep{duong2018active} selects examples with the lowest parser confidence on their LFs. Please refer to~\eqref{eq:confidence} in~\sref{sec:pool_al} for the details of S2S (FW).
    \item  \textit{CSSE}~\citep{hu2021phrase} selects the most representative and diversified utterances for MT. Please refer to Eq.~\eqref{eq:csse} in~\sref{sec:pool_al} for the details of CSSE.
    \item  \textit{Max Compound}~\citep{oren2021finding} selects examples that diversify the atoms and compounds in the LFs. Please refer to~\sref{sec:parser_framework} for the descriptions of atoms and compounds.
    \item  \textit{ORACLE} trains the parser on the full bilingual training set.
\end{itemize}

\paragraph*{Evaluation.} We adopt the exact match accuracy of LFs for all the experiments. We only report the parser accuracy on the target languages as we found the influence of new data is negligible to the parser accuracy on English data.

\paragraph*{Base Parser.} We employ \bertlstm~\citep{moradshahi2020localizing} as our multilingual parser. The description of \bertlstm can be found at~\sref{chap3:sec:parser}. In our implementation, we specifically employ the Multilingual BERT variant as the encoder.

\paragraph*{Implementation Details.}

We tune the hyperparameters of \bertlstm on English data. For a fair comparison, we fix the hyperparameters of the parser when evaluating the AL methods. Specifically, we set the learning rate to 0.001, batch size to 128, LSTM decoder layers to 2, embedding size for each LF token to 256, and epochs to 240 and 120 for the training on \geo and \nlmap, respectively.

\subsection{Analysis}
\paragraph*{Hyperparameter Tuning.} 
\tref{tab:hyper} displays the experiment results with the hyperparameters tuned using only English data (EN) and the hyperparameters tuned using seed data on:

\begin{itemize}
    \item English data plus a small subset (10\% of train data plus development data) in the target language (EN + 10\%).
    \item The full bilingual data (EN + full).
    \item The same dataset in a different pair of languages from our experiment languages (Diff Lang).
    \item A different dataset in the same languages as our experiment (Diff Data).
\end{itemize}


\begin{table}[ht]
\centering
  \resizebox{0.6\textwidth}{!}{%
\begin{tabular}{ccccc}
\toprule
     & \multicolumn{3}{c}{\geo} & \nlmap \\
&DE & TH & EL& DE\\
\midrule
 EN (Ours)  &  73.86 & 74.57 &  77.57 &  69.43\\
 \hline
 EN + 10\%     &  73.86 & 74.57 &  77.57 &  69.02\\
EN + full &  73.86 & 74.57 &  77.14 &  69.43\\
Diff Lang     & 73.86 & 74.04 & 77.57&  -\\
Diff Data    &  71.36 & - &  -&67.72 \\
\bottomrule
\end{tabular}
  }
    \caption{The parsing accuracies on \geo and \nlmap test sets in various target languages after translating 16\% of the English examples selected by \lfslcd with the optimal hyperparameters obtained by different tuning approaches. We grid search the decay weight $\beta$ in 0, 0.25, 0.5, 0.75 and the weight balance rate $\alpha$ in 0.25, 0.5, 0.75, 1. The optimal hyperparameters for obtaining our main results are 0.75 and 0.75 for all language pairs of \geo and 0.75 and 0.25 for multilingual \nlmap.}
  \label{tab:hyper}
\end{table}}

From \tref{tab:hyper}, our approach takes significantly fewer annotation resources than others to find the optimal hyperparameters. Adding more target-language data does not help obtain better hyperparameters, validating our assumption that English data is enough for \lfslcd to obtain good hyperparameters. Surprisingly, the hyperparameters tuned on a different language pair do not significantly worsen the selection choices. However, tuning hyperparameters from other datasets results in inferior parsing performance, which is anticipated as different datasets include different LFs, but the performance of \lfslcd is closely related to the LF structures.

\begin{figure}[ht!]
    \centering
    \includegraphics[width=\textwidth]{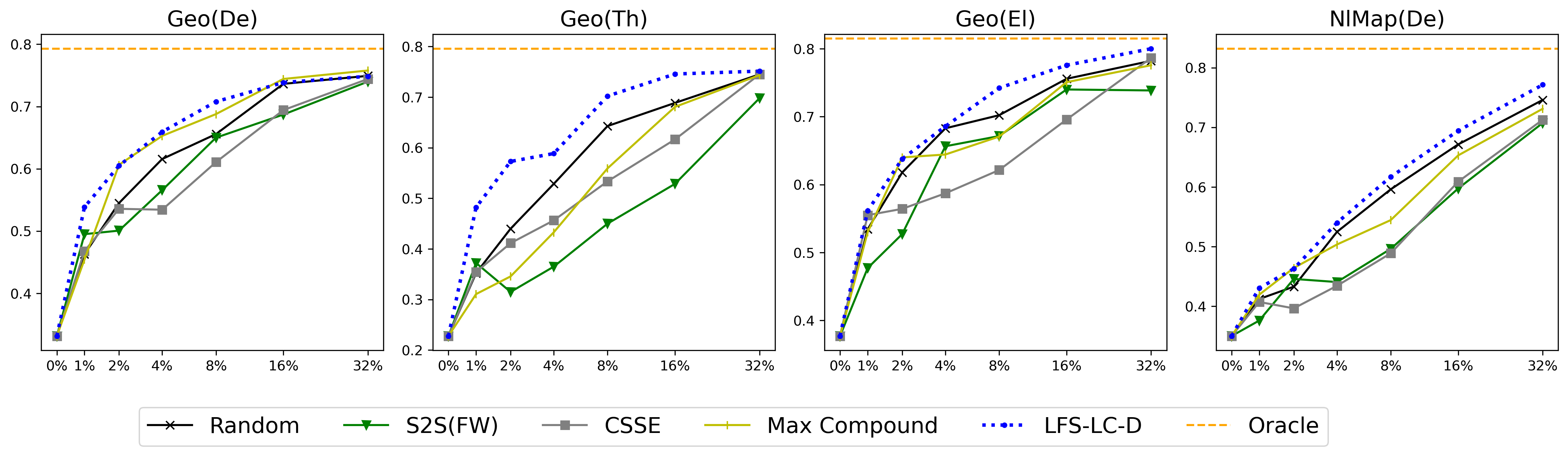}
    \caption{The parsing accuracies at different iterations on the test sets of \geo and \nlmap in German (De), Thai (Th) and Greek (El) using different selection approaches. All experiments are
run 5 times with different seeds. }
    \label{fig:main}
\end{figure}

\paragraph*{Active Learning Results.} \fref{fig:main} shows that only a small amount of target-language data significantly improves the parsing performance over the zero-shot performance. For example, merely 1\% of training data improves the parsing accuracies by up to 13\%, 12\%, 15\% and 6\% on \geode, \geoth, \geoel and \nlmapde, respectively. With the best selection approach \lfslcd, translating 32\% of instances yields parsing accuracies on multilingual \geo and \nlmap that are comparable to translating the whole dataset, with an accuracy gap of less than 5\%, showing that our \almsp might greatly minimize the translation effort.

\lfslcd consistently outperforms alternative baselines on both multilingual datasets when the sampling rate is lower than 32\%. In contrast, \ssfw consistently yields worse parser performance than the other baselines. Our inspection reveals that the parser is confident in instances with similar LFs. \maxcompound diversifies LF structures as \lfslcd, however, it does not perform well on \geoth. \csse diversifies utterances yet performs poorly. We hypothesize that diversifying LF structures is more advantageous to the semantic parser than diversifying utterances. \rand also performs consistently across all settings but at a lesser level than \lfslcd.

\begin{figure*}[ht!]
    \centering
    \includegraphics[width=1\textwidth]{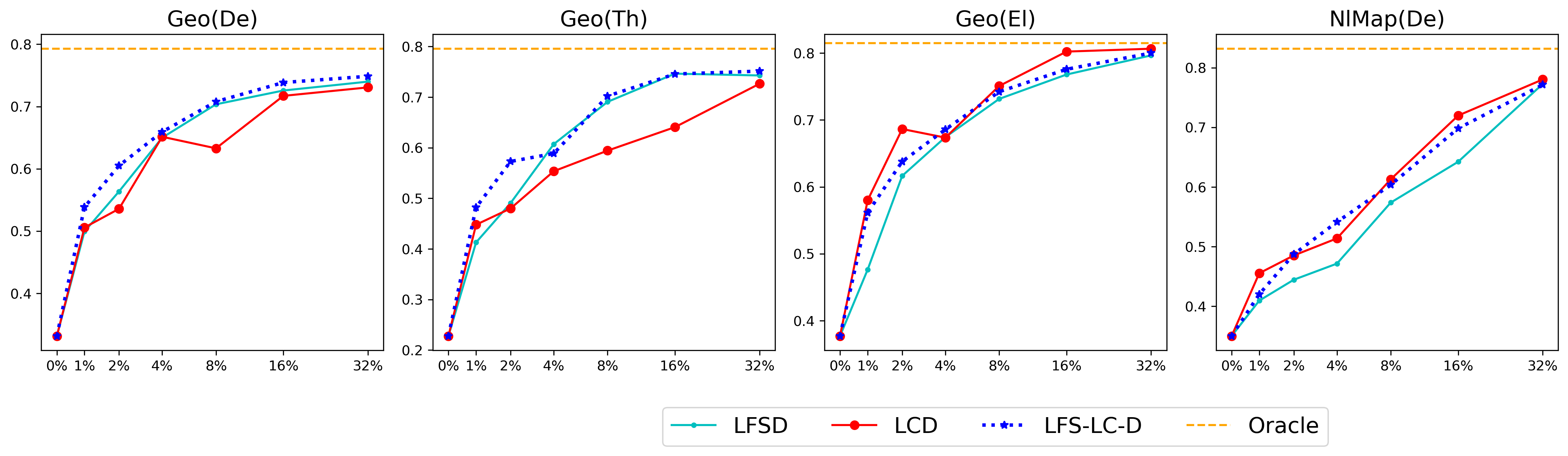}
    \caption{The parsing accuracies at different iterations on the test sets of \geo and \nlmap in German (De), Thai (Th) and Greek (El) using \lfsd, \lcd and \lfslcd, respectively.}
    \label{fig:al_msp_abl}
\end{figure*}

\paragraph*{Ablation of the Individual Terms.} As in~\fref{fig:al_msp_abl}, we also inspect each individual term, \lfsd and \lcd, in \lfslcd. Both terms have overall lower performance than \lfslcd, indicating the combination of two terms is necessary. Specifically, \lfsd performs poorly on \nlmap at the low sampling region. We inspect that \nlmap includes 5x more compounds than \geo. Therefore, it is difficult for the small number of chosen examples to encompass all types of compounds. \lcd performs poorly on \geoth. We notice that Thai is an analytic language linguistically distinct from English, German or Greek, so the entropy values of the probability $p(x_s|a)$ over lexicons in Thai (p$=$0.03) is statistically more different to the ones over English than German (p$=$5.80e-30), and Greek (p$=$1.41e-30)\footnote{We use the Student's t-test~\citep{demvsar2006statistical}.}. The two terms could benefit each other, so \lfslcd performs steadily across different settings. 

\paragraph*{Parser Accuracies on English Test Sets.}
\begin{figure}[ht!]
    \centering
    \includegraphics[width=0.75\textwidth]{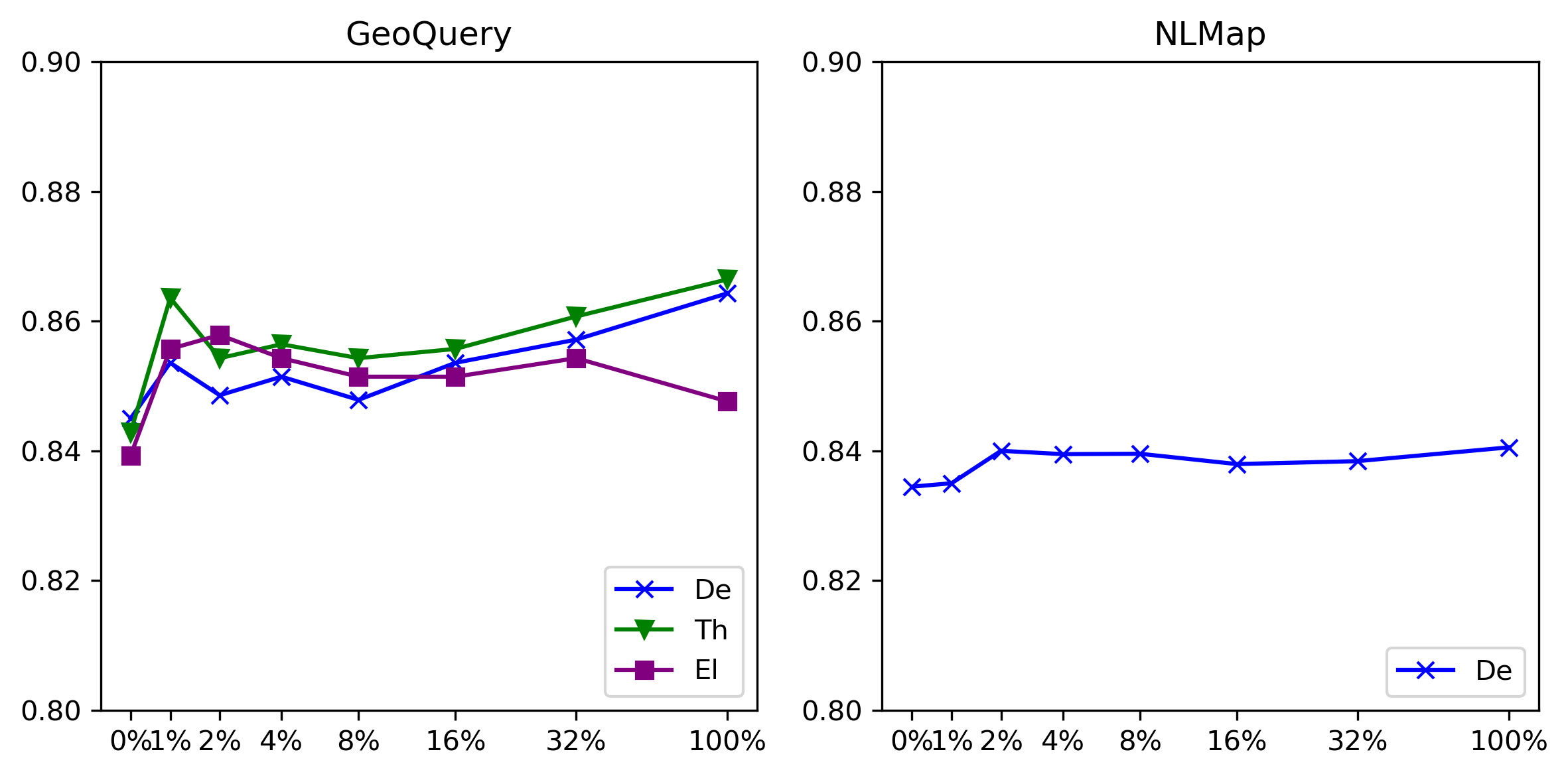}
    \caption{The parsing accuracies at different iterations on the English test sets of \geo and \nlmap after selecting data in German (De), Thai (Th) and Greek (El) using \lfslcd, respectively.}
    \label{fig:abl_en_res}
\end{figure}
As in~\fref{fig:abl_en_res}, training the parser on the data in the target language does not significantly influence the parser's performance on the English test sets. Therefore, in~\sref{sec:exp}, we only report the experimental results on the test sets in the target languages.

\section{Summary}
In this chapter, we conduct the first in-depth empirical study to investigate AL for MSP. In addition, we propose a method to select examples that maximize MSP performance in the target language and a cost-free hyperparameter tuning method. The experiments show that this method with the proper hyperparameters selects better examples than the other baselines. The AL procedure with the ideal example selection could significantly reduce the manual translation effort for adapting the multilingual semantic parser to a new language.
\chapter{Machine Translation and Active Learning for Multilingual Semantic Parsing}
\label{chap:comb_human_auto}

\epigraph{Quality is not an act, it is a habit.}{\textit{Aristotle}}

\fbox
{
\begin{minipage}{0.95\textwidth}
     Subsequent to PhD enrollment, research findings from~\cref{chap:comb_human_auto}, along with additional new results, have been submitted for publication and appear in the paper:\\
     Z. Li, L. Qu, P. R. Cohen, R. V. Tumuluri, G. Haffari ``The Best of Both Worlds: Combining Human and Machine Translations for Multilingual Semantic Parsing with Active Learning'' The 61st Annual Meeting of the Association for Computational Linguistics (ACL). 2023.
\end{minipage}
}

This chapter, which follows the previous one, addresses the data imbalance issue for MSP while adapting the parser to a new language. However, a limited annotation budget for manual translation and an MT system is now available.


Prior works proposed to utilize the translations by either humans~\citep{li2021mtop} or MT service~\citep{moradshahi2020localizing,sherborne2020bootstrapping} to alleviate the data imbalance issue for MSP. However, human translations (HTs) are expensive, while the MTs are cheap but prone to bias and error. This chapter introduces an AL approach that exploits the strengths of both humans and MTs by iteratively adding small batches of HTs into the machine-translated training set. Besides, we propose novel aggregated acquisition criteria that help the AL method select utterances to be manually translated. Our experiments demonstrate that an ideal utterance selection can significantly reduce the bias and error in the translated data, resulting in higher parser accuracies than the parsers merely trained on the machine-translated data with only a minimal translation cost.

\section{Introduction}


Adapting a multilingual semantic parser to new languages requires training data from the target languages. However, there is a severe imbalance of data availability among languages for current MSP research. The utterances in most current semantic parsing datasets are in English, while non-English data is scarce.

To overcome the data imbalance issue, prior studies, as well as our study in~\cref{chap:al_msp}, translate utterances in the MSP datasets from high-resource languages (e.g. English) to the target low-resource languages of interest by either human translators~\citep{susanto2017neural,duong2017multilingual,li2021mtop} or automatic MT ~\citep{moradshahi2020localizing,sherborne2020bootstrapping}. Unfortunately, HT, though effective, is cost-intensive and time-consuming. While the cost of MTs is much lower than that of HTs, the low quality of the machine-translated utterances severely weakens the performance of the MSP in the target languages.

We observe that the quality of MTs is lower than that of HTs, mainly due to translation bias and errors.  First, MT systems are likely to be influenced by algorithmic bias. Hence, the output of MT systems is generally less lexically and morphologically diverse than HTs~\citep{vanmassenhove2021machine}. So, there is a lexical distribution discrepancy between the machine-translated and the human-generated utterances. Second, MT systems are prone to generate translations with errors~\citep{daems2017identifying}. 

Prior study~\citep{moradshahi2020localizing} demonstrates that adding only a small portion of human-translated data into the complete set of machine-translated training data significantly improves the MSP performance on the test set of the target language. Given this observation, we propose a novel annotation strategy based on AL that benefits from both \textbf{H}uman translations and \textbf{A}utomatic machine \textbf{T}ranslations (\hatt). In contrast to \almsp, which we proposed in the previous chapter, we assume an available MT system in this chapter, which is reasonable given that MT services such as Google Translation~\citep{wu2016google} are currently easy to access and inexpensive. \hatt initially machine-translates all utterances in training sets from the high-resource languages to target languages. Then, for each iteration, \hatt selects a subset of utterances from the original training set to be translated by human translators, followed by adding the HT data to the MT training data. The multilingual parser is trained on the combination of both types of translated data.

We further investigate how \hatt can select utterances whose HTs maximally benefit the parser performance in the target languages. We assume the performance improvement is ascribed to the less biased and erroneous training set in a mixture of the MT and HT data. We have found that resolving the bias and error issues for the translations of the most semantically diversified and representative utterances improves the parser performance to the greatest extent. Given this assumption, we provide an \textbf{A}ggregated acquisition function that scores the utterances on how much their HTs can mitigate the \textbf{B}ias and \textbf{E}rror issues for learning the multilingual parsers (\amsp). \amsp aggregates four individual acquisition functions, two of which measure the bias and error degree for the translations of the source utterances. The other two encourage the selection of the most representative and semantically diversified utterances.

Our key contributions are as follows:
\begin{itemize}
 \item We propose a novel AL procedure, \hatt, that benefits from two popular annotation strategies for training the multilingual semantic parser. \hatt greatly boosts the performance of the parser trained on MT data while it requires only a small extra human annotation cost. With only 16\% of total utterances translated by humans, the parser accuracies on the \geo~\citep{susanto2017neural} and \nlmap~\citep{haas2016} test sets of the target languages can be improved by up to 28\% and 5\%, respectively, compared to the accuracies of those trained on machine-translated data, and are only up to 5\% away from the accuracies of the \oracle parsers trained on all human data.
 \item We propose an aggregated acquisition function, coined \amsp, specifically designed to select utterances where their HTs mitigate translation bias and error for learning a good multilingual semantic parser. Compared to other SOTA acquisition baselines, given the same selection budget, our experiments show \amsp consistently results in the less biased and erroneous training sets and higher parser accuracies on the test sets of the \geo and \nlmap in the target languages.
\end{itemize}

\section{Difficulties for MSP with Automatic MT}

\subsection{Evaluation} To delve deeper into the challenges of employing automatic MT for MSP, we start by outlining the metrics used for various aspects of evaluation. Specifically, we measure the performance of the multilingual parser in terms of accuracy. To measure the degree of bias in the training set, we employ metrics such as Jensen–Shannon divergence, MAUVE~\citep{pillutla2021mauve}, and MTLD~\citep{mccarthy2005assessment}. Additionally, the translation error within the training set is quantified using the Discrepancy Rate of Back-translations. 

\begin{description}
    \item[Accuracy:] To evaluate the performance of the multilingual semantic parser, we report the accuracy of exactly matched LFs at each query round. Considering that the parser accuracies on the English test sets are not relevant to evaluating the AL method, we only report the accuracies on the test sets in the \textit{target} languages.
    \item[Bias of the Training Set:] We use three metrics to measure the bias of the training data in the target language at each query round. 
    
    \begin{itemize}
      \item  \textit{Jensen–Shannon (JS) divergence}~\citep{pillutla2021mauve} measures the JS divergence between the n-gram frequency distributions of the utterances in training set $\hat{\mathcal{D}}_t \cup \bar{\mathcal{D}}^{q}_t$ generated by each AL method and test set $\mathcal{T}_{t}$.
      \item \textit{MAUVE} compares the learnt distribution from the training set to the distribution of human-written text in the test set $\mathcal{T}_{t}$ using Kullback–Leibler divergence~\citep{kullback1951information} frontiers. Here we use n-gram lexical features from the text when calculating MAUVE. \textit{JS} and \textit{MAUVE} together measure how lexically "human-like" the generated training set is. 
      \item As in Section~\ref{chap3:para_diversity_metrics}, we adopt \textit{MTLD} to evaluate the lexical diversity of the training set $\hat{\mathcal{D}}_t \cup \bar{\mathcal{D}}^{q}_t$. 
    \end{itemize}

\item[{Discrepancy Rate of the Back-translations (BTs):}] Since we do not possess bilingual knowledge, we use back-translation~\citep{tyupa2011theoretical} to access the translation quality. At each query round, we randomly sample 100 utterances from the utterances selected by each acquisition during five experiments using different random seeds. The back-translation is obtained by using Google Translation to translate the MTs of the 100 sampled utterances back into English. Two annotators manually check the percentage of the BTs that are not semantically equivalent to their original utterances. We only consider a BT discrepancy when both annotators agree. Ideally, the utterances with fewer mistranslations would see fewer semantic discrepancies between the BTs and the original.
\end{description}

\subsection{Discussion}
Although using MT for MSP is cost-effective, the parser performance is usually much lower than the one trained with human-translated data. For example, as shown in \tref{tab:bias_analysis}, the parsers trained on HTs all have significantly higher accuracies than those trained on MTs in different dataset settings. Such performance gaps lie with two major issues in the MT data, discussed in the following. 
\begin{table}[ht]
    \vspace{-2mm}
\centering
  \resizebox{0.9\textwidth}{!}{%
  \begin{tabular}{|c||cc| cc| cc|cc|}
    \toprule
    \multirow{2}{*}{Metrics} &
      \multicolumn{2}{c|}{\geode} &
      \multicolumn{2}{c|}{\geoth} &
      \multicolumn{2}{c|}{\geoel} &
      \multicolumn{2}{c|}{\nlmapde}\\
     & HT& MT & HT& MT & HT& MT & HT & MT  \\
      \hline \hline
       Accuracy$\uparrow$ &  78.14 & 47.21 & 79.29 & 56.93 & 80.57 & 68.5 & 81.57 &  67.86 \\
             \hline
           \hline
 BT Discrepancy Rate$\downarrow$ &  2\% & 11\% & 3\% & 12\% & 3\% & 10\% & 2\% &  10\% \\
      \hline
    \hline
          JS$\downarrow$ &  36.67 & 59.95 & 32.02 & 73.83 & 33.67 & 56.36 & 33.78 &  46.84 \\
    MAUVE$\uparrow$ & 96.01 & 22.37  & 97.52  &8.48 & 97.12 & 45.01 & 97.34 & 70.24 \\
\hline
MTLD$\uparrow$ & 26.02&  22.50  & 20.74  & 19.07 & 28.16& 27.08 & 44.80&42.38       \\
    \bottomrule
  \end{tabular}%
  }
    \caption{The scores of five metrics to measure the quality of the HTs and MTs in German (De), Thai (Th), and Greek (El) of the utterances in \geo and \nlmap. $\uparrow$/$\downarrow$ means the higher/lower score the better.}
  \label{tab:bias_analysis}

\end{table}
\paragraph*{Translation Bias.}
Many existing MT systems amplify biases observed in the training data~\citep{vanmassenhove2021machine}, leading to two problems that degrade the parsers' performance trained on MT data. 
\begin{itemize}
    \item The MTs lack lexical diversity~\citep{vanmassenhove2021machine}. 
As shown in~\tref{tab:bias_analysis}, MTLD~\citep{vanmassenhove2021machine} values show that the HTs of utterances in multilingual \geo and \nlmap are all more lexically diversified than MTs. The lexical diversity of training data is essential to improving the generalization ability of the parsers~\citep{shiri2022paraphrasing,xu2020autoqa,wang2015overnight}. 
\item The lexical distribution of the biased MTs is different to the human-written text. The two metrics, JS divergence~\citep{manning1999foundations} and MAUVE~\citep{pillutla2021mauve}, in~\tref{tab:bias_analysis} show the HTs of utterances in \geo and \nlmap are more lexically close to the human-generated test sets than MTs.
\end{itemize}

\begin{figure}[ht]
    \centering
    \includegraphics[width=0.75\textwidth]{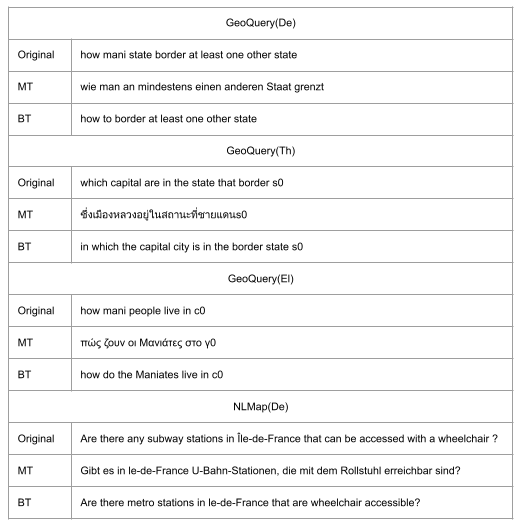}
    \caption{The source-language utterances from \geode, \geoel, \geoth, and \nlmapde, respectively, and their corresponding MTs and BTs generated by Google Translation System.}
  \label{fig:back_examples}
\end{figure}

\paragraph*{Translation Error.} 
MT systems often generate translation errors, which cause incongruence between the source and target texts. One common error type is mistranslation~\citep{vardaro2019translation}, which alters the semantics of the source sentences after translation. Training a multilingual semantic parser on the mistranslated data would cause incorrect parsing output, as LFs are the semantic abstraction of the utterances. 
BT Discrepancy Rate in~\tref{tab:bias_analysis} demonstrates that the mistranslation problem is more significant in the machine-translated datasets.
We observe that the BT discrepancy patterns vary as in~\fref{fig:back_examples}. For instance, the semantics of the BTs for Thai in \geo are altered dramatically due to the incorrect reorder of the words. Within \nlmap, the meanings of some German locations are inconsistent after the BT process. 

\section{Combining Human and Automatic Translations with Active Learning}


To mitigate the negative effect of translation bias and error in the MT data, we propose \hatt, which introduces extra human supervision to machine supervision when training the multilingual semantic parser. Two major intuitions motivate our training approach: 
\begin{itemize}
\item Adding the HTs to the training data could enrich its lexical and morphological diversity and ensure that the lexical distribution of the training data is closer to the human test set, thus improving the parsers' generalization ability~\citep{shiri2022paraphrasing,xu2020autoqa,wang2015overnight}. 
\item HTs are less erroneous than the MTs~\citep{freitag2021experts}. The parser could learn to predict correct abstractions with less erroneous training data.
\end{itemize}

Our \hatt AL setting considers only the \textit{bilingual} scenario, one of which is in the high-resource language, and the other one is in the low-resource. However, it is easy to extend our method to more than two languages. We assume access to a well-trained black-box multilingual MT system, $g^{mt}(\cdot)$, and a semantic parsing training set that includes utterances in a high-resource language $l_s$ (e.g., English) paired with LFs, $\mathcal{D}_s = \{(\vx_{s}^i, \vy^i)\}_{i=1}^{N}$, two human-generated test sets $\mathcal{D}^{test}_{s} = \{(\vx_{s}^i, \vy^i)\}_{i=1}^{M}$ and $\mathcal{D}^{test}_{t} = \{(\vx_{t}^i, \vy^i)\}_{i=1}^{M}$ with utterances in high and low-resource languages, respectively. Each utterance $\vx_{s}$ in $\mathcal{D}_s$ is translated into the utterance $\hat{\vx}_{t} = g^{mt}_{s\rightarrow t}(\vx_{s})$ in the target language $l_t$ by the MT system, $\hat{\mathcal{D}}_{t} = \{(\hat{\vx}^i_{t}, \vy^i)\}_{i=1}^{N}$. The goal of our AL method is to select an optimal set of utterances from the training data in the source language, $\tilde{\mathcal{D}}_s \in \mathcal{D}_s$, and ask human translators to translate them into the target language, denoted by $\bar{\mathcal{D}}_t = g^{ht}_{s\rightarrow t}(\tilde{\mathcal{D}}_s)$. The semantic parser is subsequently trained on the union of $\bar{\mathcal{D}}_t$ and $\hat{\mathcal{D}}_{t}$. It should be noted that the examples in $\hat{\mathcal{D}}_{t}$ are not replaced by those in $\bar{\mathcal{D}}_t$. 


The selection criterion is based on the \textit{acquisition functions} that score the source utterances. Following the conventional batch AL setting~\citep{duong2018active}, there are $Q$ rounds of selection. At $q$th round, AL selects utterances with a budget size of $K_q$.
The detailed \hatt AL procedure iteratively performs the following steps as in~\algoref{algo:al_auto}.




\begin{algorithm}[ht]
{\small
\SetKwData{Left}{left}\SetKwData{This}{this}\SetKwData{Up}{up}
\SetKwFunction{Union}{Union}\SetKwFunction{FindCompress}{FindCompress}
\SetKwInOut{Input}{Input}\SetKwInOut{Output}{Output}
\SetAlgoLined
\Input{Initial training set $\mathcal{D}^{0}=\mathcal{D}_s \cup \hat{\mathcal{D}}_t$, source utterance pool $\mathcal{D}_s$, budget size $K_q$, number of selection rounds $Q$, human annotators $g^{ht}(\cdot)$}
\Output{A well-trained multilingual parser $P_{\bm{\theta}}(\vy|\vx)$}

\textcolor{blue}{\# \textit{Train the initial parser on the initial data}} \\
     Update $\bm{\theta}$ of $P_{\bm{\theta}}(\vy|\vx)$ with $\nabla_{\bm{\theta}}\mathcal{L}(\bm{\theta})$ on $\mathcal{D}^{0}$ \\
Evaluate $P_{\bm{\theta}}(\vy|\vx)$ on $\mathcal{D}^{test}_s$ and $\mathcal{D}^{test}_t$ \\
Estimate the acquisition function $\phi(\cdot)$\\
$\bar{\mathcal{D}}^{0}_t = \emptyset$ \textcolor{blue}{\# \textit{Empty set of human-translated data}} \\
$\bar{\mathcal{D}}^{0}_s = \mathcal{D}_s$ \textcolor{blue}{\# \textit{Initial source utterance pool}}\\
\For{$q \gets 1$ to $Q$} 
{
\textcolor{blue}{\# \textit{Select a subset $\tilde{\mathcal{D}}^{q}_s \in \mathcal{D}^{q-1}_s$ of the size $K_q$ with the highest scores ranked by the acquisition function $\phi(\cdot)$}} \\

$\tilde{\mathcal{D}}^{q}_s = \text{TopK} (\phi(\bar{\mathcal{D}}^{q-1}_s), K_q)$ \\
$\bar{\mathcal{D}}^{q}_s = \bar{\mathcal{D}}^{q-1}_s \setminus \tilde{\mathcal{D}}^{q}_s$ \\
\textcolor{blue}{\# \textit{Translate the utterances in $\tilde{\mathcal{D}}^{q}_s$ into the target language $l_t$ by human annotators}} \\
$\mathcal{D}^{q}_t = g^{ht}_{s\rightarrow t}(\tilde{\mathcal{D}}^{q}_s)$\\
\textcolor{blue}{\# \textit{Merge all human-translated data}} \\
$\bar{\mathcal{D}}^{q}_t = \bar{D}^{q-1}_t \cup D^{q}_{t}$  \\
\textcolor{blue}{\# \textit{Add the human-translated data into the training data}} \\
$\mathcal{D}^{q} = \mathcal{D}_s \cup \hat{\mathcal{D}}_t \cup \bar{\mathcal{D}}^{q}_t$ \\
\textcolor{blue}{\# \textit{Train the parser on the updated data}} \\
Update $\bm{\theta}$ of $P_{\bm{\theta}}(\vy|\vx)$ with $\nabla_{\bm{\theta}}\mathcal{L}(\bm{\theta})$ on $\mathcal{D}^{q}$ \\
Evaluate $P_{\bm{\theta}}(\vy|\vx)$ on $\mathcal{D}^{test}_s$ and $\mathcal{D}^{test}_t$\\
Re-estimate  $\phi(\cdot)$
}
}
\caption{\hatt procedure
}
\label{algo:al_auto}
\end{algorithm}

\subsection{Acquisition Functions} 
\label{sec:acq}

The acquisition functions assign higher scores to those utterances whose HTs  can boost the parser's performance more than the HTs of the other utterances.
The prior AL works~\citep{sener2018active,zhdanov2019diverse,nguyen2004active} suggested that the most representative and diversified examples in the training set improve the generalization ability of the machine learning models the most. Therefore, we provide a hypothesis that \textit{we should select the representative and diversified utterances in the training set, whose current translations have significant bias and errors}. We postulate fixing problems of such utterances improves the parsers' performance the most. We derive four acquisition functions from this hypothesis to score the utterances. Then, \amsp aggregates these acquisition functions to gain their joint benefits. In each AL round, the utterances with the highest \amsp scores are selected.

\paragraph*{Translation Bias.} 

We assume an empirical conditional distribution, $P_{e}^{q}(\vx_t|\vx_s)$, for each utterance $\vx_s$ in $\mathcal{D}_{s}$ at $q$th AL selection round. 

Intuitively, if an utterance $\vx_s$ has highly biased translations, it is likely to have fewer corresponding translations in the target languages than other utterances. This scarcity of training data makes it challenging for the parser to generalize to the LF associated with $\vx_s$. In statistical terms, an utterance $\vx_s$ with the most biased translations would be characterized by a highly skewed empirical conditional distribution.
Therefore, we measure the translation bias by calculating the entropy of the empirical conditional distribution, $H(P_{e}^{q}(\vx_{t}|\vx_s))$, and select the $\vx_s$ with the lowest entropy. Since the translation space  $\mathcal{X}_{t}$ is exponentially large, it is intractable to calculate the entropy directly. Following~\citet{settles-craven-2008-analysis}, we adopt two approximation strategies, \textit{N-best Sequence Entropy} and \textit{Maximum Confidence Score}, to approximate the entropy.

\noindent\textit{$\bullet$ N-best Sequence Entropy:}

\begin{equation}
    \phi_b(\vx_s) = - \sum_{\hat{\vx}_t \in \mathcal{N}} \hat{P_{e}^{q}}(\hat{\vx}_t|\vx_s) \log \hat{P_{e}^{q}}(\hat{\vx}_t|\vx_s)
\end{equation}

\noindent\ignorespaces where $\mathcal{N} = \{\hat{\vx}^{1}_t,...,\hat{\vx}^{N}_t\}$ are the $N$-best hypothesis sampled from the empirical distribution $P_{e}^{q}(\vx_{t}|\vx_s)$. $\hat{P_{e}^{q}}(\hat{\vx}_t|\vx_s)$ is re-normalized from $P_{e}^{q}(\hat{\vx}_t|\vx_s)$ over $\mathcal{N}$, which is only a subset of $\mathcal{X}_{t}$. For more details, please refer to~\eqref{eq:n_best_entropy} in~\sref{sec:pool_al}.



\noindent\textit{$\bullet$ Maximum Confidence Score (MCS)}:

\begin{align}
    \phi_b(\vx_s) & =  \log P_{e}^{q}(\vx'_t|\vx_s) \\
    s.t. \vx'_t & = \argmax_{\vx_t} P_{e}^{q}(\vx_{t}|\vx_s)
\end{align}

\noindent\ignorespaces The empirical distribution is difficult to obtain because it is a mixture of the HT and MT distributions, neither of which are available to us. Therefore, we use distillation training~\citep{hinton2015distilling} to train a translation model that estimates $P_{e}^{q}(\vx_t|\vx_s)$ on all the bilingual pairs $(\vx_s, \vx_t)$ in the MSP training data $\mathcal{D}^{q}$. Another challenge is that $\mathcal{D}^{q}$ is too small to distil a good translation model that imitates the mixture distribution. Here, we apply a Bayesian factorization trick that factorizes $P_{e}^{q}(\vx_{t}|\vx_s) = \sum_{\vy \in \mathcal{Y}} P_{e}^{q}(\vx_t|\vy) P_{e}^{q}(\vy|\vx_{s})$, where $\vy$ ranges over LFs representing the semanics. As there is a deterministic mapping between $\vx_s$ and the LF, $P_{e}^{q}(\vy|\vx_{s})$ is an one-hot distribution. Thus, we only need to estimate the entropy, $H(P^{q}_{e}(\vx_t|\vy))$. This has  a nice intuition:  the less diversified data has less lexically diversified utterances per each LF. Note that if we use this factorization, all $\vx_s$ that share the same LF have the same scores.

We use the lightweight Seq2Seq model to estimate $P_{e}^{q}(X_t|\vx_s)$ or $P^{q}_{e}(\vx_t|\vy)$. 
Ideally, every time a new source utterance $\vx_s$ is selected, $P_{e}^{q}(\vx_t|\vx_s)$ should be re-estimated. However, we only re-estimate $P_{e}^{q}(\vx_t|\vx_s)$ once at the beginning of each selection round to reduce the training cost.

\paragraph*{Translation Error.} 
Similar to~\citet{haffari-etal-2009-active}, we leverage back-translations to measure the translation error. We conjecture that if the translation quality for one source utterance $\vx_s$ is good enough, the semantic parser should be confident in the LF of the source utterance conditioned on its back-translations. 
Therefore, we measure the translation error for each $\vx_s$ as the expected parser's negative log-likelihood in its corresponding LF $\vy_{\vx_s}$ over all the back-translations of $\vx_s$, $\mathbb{E}_{P^{q}_{e}(\vx_{t}|\vx_s)}[-\log(P^{q}_{\bm{\theta}}(\vy_{\vx_s}|g_{t \rightarrow s}^{mt}(\vx_t)))]$, where $P^{q}_{\bm{\theta}}$ is the parser trained at $q$th round. To approximate the expectation, we apply two similar strategies as mentioned in \textit{Translation Bias}.

\noindent\textit{$\bullet$ N-best Sequence Expected Error:}

\begin{align}
    \phi_e(\vx_s) & = -\sum_{\hat{\vx}_t \in \mathcal{N}_{\vy_{\vx_s}}} \hat{P}_{e}^{q}(\hat{\vx}_t|\vx_s) \log P_{\bm{\theta}}(\vy_{\vx_s}|g_{t \rightarrow s}^{mt}(\vx_t)) 
\end{align}

\noindent\ignorespaces where $\mathcal{N}_{\vy_{\vx_s}}$ is the set of translations in $\mathcal{D}^q$ that share the same LF $\vy_{\vx_s}$ with $\vx_s$. We only back-translate utterances in $\mathcal{D}^q$ to reduce the cost of back-translations.


\noindent\textit{$\bullet$  Maximum Error:}

\begin{align}
    \phi_e(\vx_s) & = - \log P^{q}_{\bm{\theta}}(\vy_{\vx_s}|g^{mt}_{t \rightarrow s}(\vx'_t)) \\
        s.t. \vx'_t & = \argmax_{\vx_t} P^{q}_{e}(\vx_{t}|\vx_s)
\end{align}

\noindent\ignorespaces We use the same distilled translation model $P^{q}_{e}(\vx_{t}|\vx_s)$ used in \textit{Translation Bias}.


\paragraph*{Semantic Density.}
The previous AL works~\citep{nguyen2004active,donmez2007dual} have found that the most \textit{representative} examples improve the model performance the most. Therefore we expect to reduce the translation bias and error for the translations of the most representative source utterances. As such,  the utterances should be selected from the dense regions in the semantic space,
\begin{equation}
    \phi_s(\vx_s) = \log P(\vx_s).
\end{equation}
We use kernel density estimation~\citep{botev2010kernel} with the exponential kernel to estimate $P(\vx_s)$, while other density estimation methods could be also used. The feature representation of $\vx_s$ for density estimation is the average pooling of the contextual sequence representations from the MSP encoder. The density model is re-estimated at the beginning of each query selection round.

\paragraph*{Semantic Diversity.} The role of the semantic diversity function is twofold. First, it prevents the AL method from selecting similar utterances. Resolving the bias and errors of similar utterances in a small semantic region does not resolve the training issues for the overall dataset. Second, semantic diversity correlates with lexical diversity; hence improving it also enriches lexical diversity.
\begin{align}
    \phi_d(\vx_s) =     
    \begin{cases}
      0 & \text{if $c(\vx_s) \notin \bigcup_{\vx^{i}_s \in \mathcal{S}} c(\vx^{i}_s)$}\\
      -\infty & \text{Otherwise}
    \end{cases}
\end{align}
\noindent\ignorespaces where $c(\vx_s)$ maps each utterance $\vx_s$ into a cluster id and $\mathcal{S}$ is the set of cluster ids of the selected utterances. We use a clustering algorithm to diversify the selected utterances as in~\citet{ni2020merging,nguyen2004active}. The source utterances are partitioned into $|\mathcal{C}|$ clusters. We select one utterance at most from each cluster. Notice the number of clusters should be greater than or equal to the total budget size until the current selection round, $|\mathcal{C}| \geq \sum^{q}_{i=1}K_{i}$. The clusters are re-estimated every round. To ensure the optimal exploration of semantic spaces across different query rounds, we adopt Incremental K-means~\citep{liu2020active} as the clustering algorithm. At each new round, Incremental K-means considers the selected utterances as the fixed cluster centres, and learns the new clusters conditioned on the fixed centres. The feature representation of $\vx_s$ for Incremental K-means is from the MSP encoder as well.

\paragraph*{Aggregated Acquisition.}
We aggregate the four acquisition functions into one,
\begin{align}\nonumber
\label{eq:sample}
    & \phi_{A}(\vx_s) = \sum_{k}\alpha_{k} \phi_{k}(\vx_s) 
\end{align}
\noindent where $\alpha_{k}$'s are the coefficients. Each $\phi_{k}(\vx_s)$ is normalized using quantile normalization~\citep{bolstad2003comparison}. Given the approximation strategies we used for \textit{Translation Bias} and \textit{Translation Error}, we provide two types of aggregations, \amspnbest and \amspmax. \amspnbest applies \textit{N-best Sequence Entropy} and \textit{N-best Sequence Expected Error}. \amspmax applies \textit{Maximum Confidence Score} and \textit{Maximum Error}.

\section{Experiments}
\label{sec:experiments}
\paragraph*{Datasets.}
Following~\cref{chap:al_msp}, we evaluate our AL method for MSP on datasets, \geo and \nlmap with multilingual human-translated versions. \geo includes 600 utterances-LF pairs as the training set and 280 pairs as the test set. \nlmap includes 1500 training examples and 880 test examples. 

In our work, we consider English as the \textit{resource-rich} source language and use Google Translate System\footnote{https://translate.google.com/} to translate all English utterances in \geo into German (De), Thai (Th), Greek (El) and the ones in \nlmap into German, respectively. The AL methods actively sample English utterances, the HTs of which are obtained from the multilingual \geo and \nlmap. 


\paragraph*{AL Setting.} The \hatt procedure performs five rounds of query, which accumulatively samples 1\%,  2\%, 4\%, 8\%, and 16\% of total English utterances in \geo and \nlmap. We only perform five rounds as we found the performance of the multilingual parser is saturated after sampling 16\% of examples with most acquisition functions.

\paragraph*{Base Parser.} Similar to our works in Chapters~\ref{chap:auto} and~\ref{chap:al_msp}, we use \bertlstm~\citep{moradshahi2020localizing} as our multilingual parser. 

\paragraph*{Baselines.} We compare \amsp with eight acquisition baselines and an oracle baseline.

\begin{itemize}
    \item \textbf{Random} randomly selects English utterances in each round.
    \item \textbf{Cluster}~\citep{ni2020merging} partitions the utterances into different groups using K-means and randomly selects one example from each group.
    \item \textbf{LCS (FW)}~\citep{duong2018active} selects English utterances for which the parser is least confident in their corresponding LFs, $\vx = \argmin_{\vx}p_{\theta}(\vy|\vx)$.
    \item \textbf{LCS (BW)}~\citep{duong2018active}, on the opposite of LCS (BW), trains a text generation model to generate text given the LF. The English utterances are selected for which the text generation model is least confident conditioned on their corresponding LFs, $\vx = \argmin_{\vx}p_{\theta}(\vx|\vy)$.
    \item \textbf{Traffic}~\citep{sen2020uncertainty} selects utterances with the lowest perplexity and highest frequency in terms of their corresponding LFs.
    \item \textbf{CSSE}~\citep{hu2021phrase} combines the density estimation and the diversity estimation metrics to select the most representative and semantically diversified utterances.
    \item \textbf{RTTL}~\citep{haffari-etal-2009-active} uses BLEU~\citep{papineni2002bleu} to estimate the translation information losses between the back-translations and the original utterances, and select utterances with the highest losses.
    \item \textbf{LFS-LC-D} is the selection method for AL-MSP proposed in~\cref{chap:al_msp}. Please see~\cref{chap:al_msp} for more details. 
    \item \textbf{ORACLE} trains the parser on the combination of English data, machine-translated data and the complete set of human-translated data.
\end{itemize}

\subsection{Main Results and Discussion}
\begin{figure}[ht!]
     
    \centering
    \includegraphics[width=\textwidth]{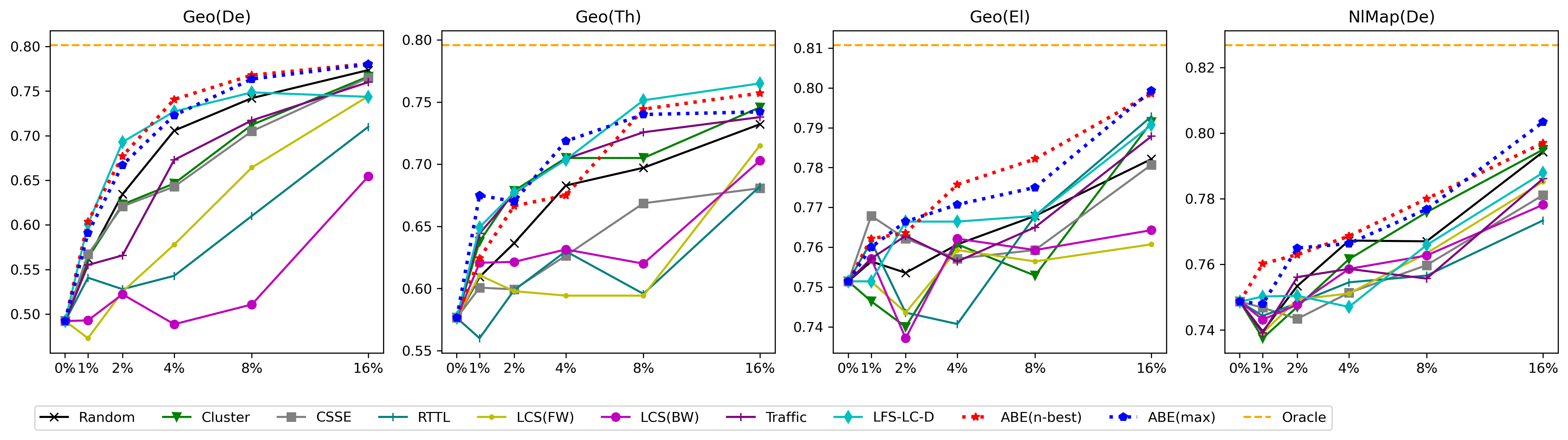}
    \caption{The parser accuracies at different query rounds using various acquisitions on the test sets of \geode, \geoth, \geoel, and \nlmapde. Orange dash lines indicate the accuracies of \oracle multilingual parsers. All experiments are run 5 times with a different seed for each run.}
    \label{fig:main_al_result}
    
\end{figure}

\paragraph*{Effectiveness of \hatt.} Fig.~\ref{fig:main_al_result} shows that \hatt significantly improves the parser accuracies on all test sets by adding only a small amount of HTs into the machine-translated training set. For example, with 16\% of English utterances translated by humans, \hatt improves the parser accuracies by up to 28\% and 25\%, respectively, on \geode and \geoth test sets. On the other hand, on \geoel and \nlmapde test sets, the accuracy improvement by \hatt is only up to 5\% because the parser has already achieved a decent performance after it is trained on the MT data. According to Table~\ref{tab:bias_analysis}, we speculate that the training sets of \geoel and \nlmapde are less biased than those of \geoth and \geode. Overall for all dataset settings, if we apply \hatt with \amsp as its selection method, the multilingual parsers can perform comparably to the \oracle parsers with no more than 5\% differences in terms of accuracies, at an extra expense of manual translating 16\% of English utterances. In addition, \hatt requires less cost of human annotation than~\almsp proposed in~\cref{chap:al_msp}, which requires the manual translation of 32\% of English utterances to achieve the same level of parser performance.



\paragraph*{Effectiveness of \amsp.} In contrast to the acquisition baselines, which focus on only one or two measurements as mentioned in Sec.~\ref{sec:acq} and do not solve the bias and error problems, \amsp jointly combines four important measurements, leading to consistently superior performance over the baselines on both datasets in various languages. For instance, \cluster, \csse, and \lfslcd select utterances with the highest semantic diversity and representativeness. \traffic selects utterances given its model perplexity and LF frequency. As a result, these baselines can perform well in some datasets and languages, but they fail to perform well across all settings consistently. Three baselines, \lcsfw, \lcsbw and \rttl, consistently perform lower than the others. \lcsfw tends to select similar examples, which lack semantic diversity. \rttl is designed to choose the utterances with the most erroneous translations. However, we inspect that such utterances with translation errors are mostly the tail examples in the training set. \amsp overcomes this issue by balancing the \textit{Translation Error} term with the \textit{Semantic Density}. \lcsbw has an opposite objective with our \textit{Translation Bias}, which means it selects the utterances with the least translation bias. Therefore, though \lcsbw performs well in the AL scenario in~\citet{duong2018active} for semantic parsing, it has the worst performance in our AL scenario.

\paragraph*{Bias, Error and Parser Performance.} As in~\tref{tab:main_diversity_error}, we also measure the bias of the training set and the BT discrepancy rates of the selected utterances at the final selection round for each acquisition function. We can see that the parser accuracy directly correlates with the training set's bias degree. The bias of the training set acquired by \rand, \traffic, \cluster, \lfslcd, and \amsp score better in general than the other baselines in terms of the bias metrics, resulting in a better parser accuracy. On the contrary, the training sets built with \rttl, \lcsfw and \lcsfw depict the larger bias. Thus the parsers perform worse. \rttl and \lcsfw that select utterances with more erroneous translations do not necessarily guarantee better accuracies for parsers. Our following ablation study shows that the parser performance can be improved by correcting the translation errors for the most representative utterances.


\begin{table*}[ht]
\centering
  \resizebox{\textwidth}{!}{%
  \begin{tabular}{|c|c|cccccccc|cc|c|}
    \toprule
     Metric & No\ HT & \rand & \cluster & \csse & \rttl & \lcsfw & \lcsbw & \traffic & \lfslcd & \amspnbest & \amspmax & \oracle  \\
      \midrule
          BT Discrepancy Rate & - & 11\%  & 14\% & 11\% & 21\% & \textbf{22\%} & 8\% & 14\% & 10\% & 17\% & 18\% & - \\
     \hline
          \hline
    JS$\downarrow$ & 59.95 &  54.15 & 54.71 & 55.53 & 54.56 & 54.38 & 56.13 & 54.58 & 54.26 &54.16 & \textbf{53.97} & 45.12 \\
    MAUVE$\uparrow$ & 22.37 &  36.99 & 36.12 & 34.52 & 35.53 & 31.61 & 29.75 & 35.67 & 36.87 &\textbf{38.96} & 35.13 & 73.04 \\
\hline
\hline
  MTLD$\uparrow$ & 22.50 & 23.79 &  23.32  & 22.65  & 22.89 & 23.00 & 22.27& 23.42 & \textbf{23.97}&23.80 & 23.78  & 24.23 \\
          \hline
    \bottomrule
  \end{tabular}
  }
    \caption{Using different acquisitions at the final query round, we depict the scores of the metrics to measure the bias of the training sets in \geode and the discrepancy rates of BTs of the total selected utterances. 
      }
  \label{tab:main_diversity_error}
\end{table*}


\begin{figure}
    \centering
     \begin{subfigure}[b]{0.45\linewidth}
              \centering
    \includegraphics[width=\textwidth]{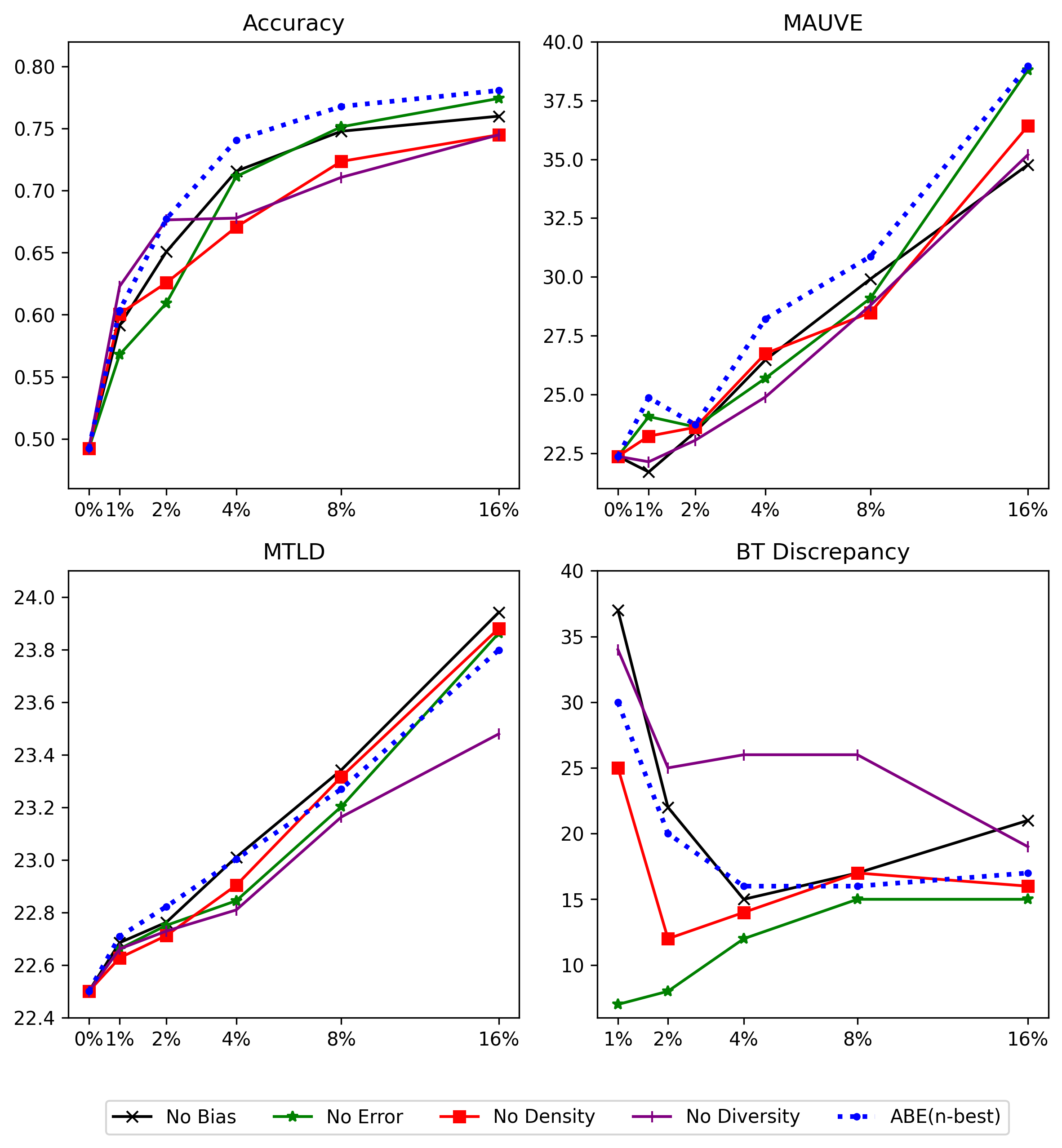}
          \caption{The parser accuracies at different query rounds after removing every single term from \amspnbest.}
           \label{fig:remove_al_result}
     \end{subfigure}
  \begin{subfigure}[b]{0.45\linewidth}
           \centering
    \includegraphics[width=\textwidth]{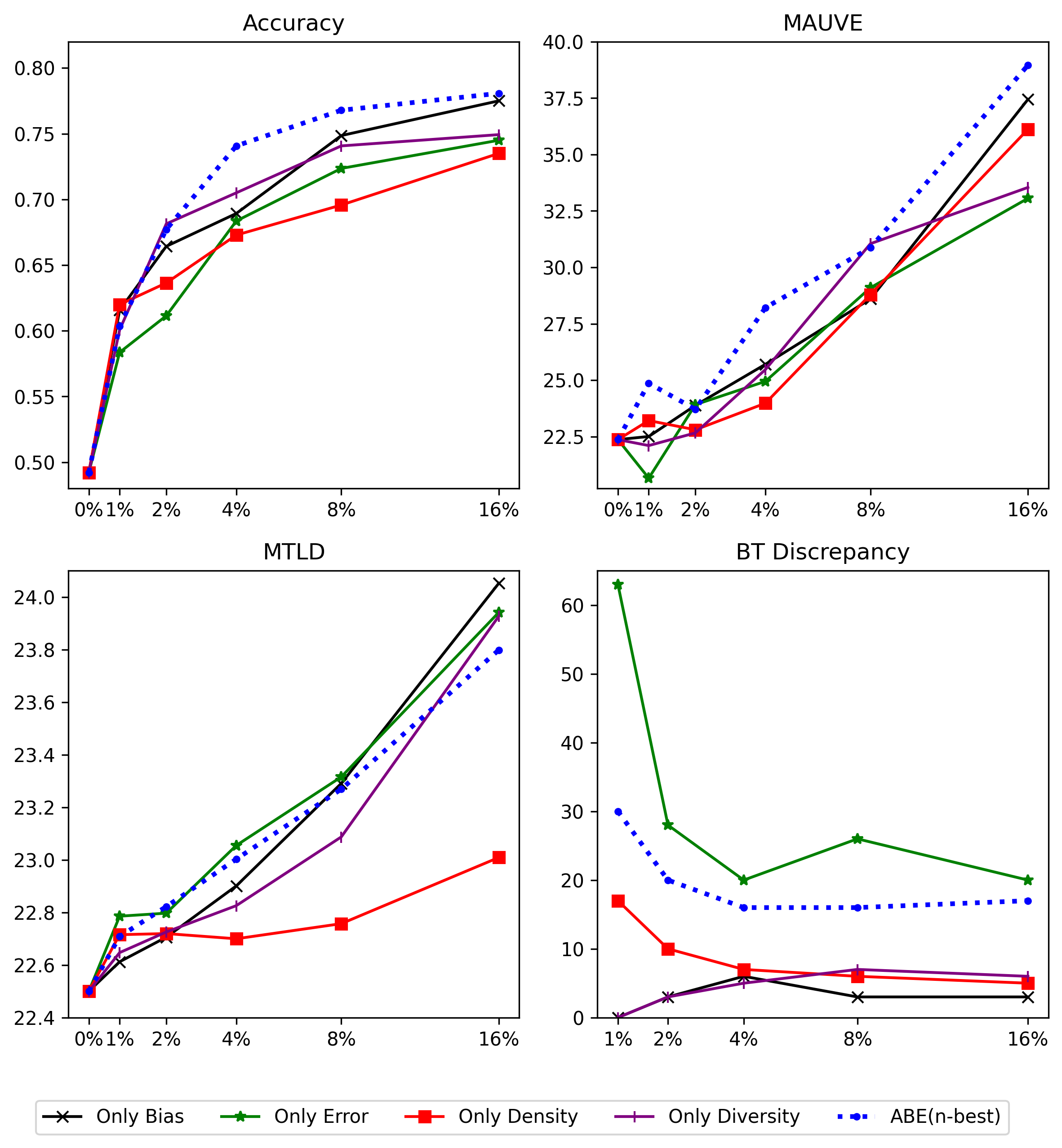}
      \caption{The parser accuracies at different query rounds using each single term from \amspnbest.}
       \label{fig:only_al_result}
  \end{subfigure} 
 \caption{The parser accuracies at different query rounds across various settings.}
\end{figure}

\subsection{Ablation Study}
\paragraph*{Influence of Individual Acquisition Terms.} As in~\fref{fig:remove_al_result}, we evaluate the effectiveness of each individual acquisition term by observing how removing each acquisition from \amspnbest influences the parser performance, the bias of the training set and the BT Discrepancy rate of the selected utterances. We can see that removing all terms degrades the parser's performance. However, each acquisition contributes to the parser's accuracy due to different reasons.

Translation Bias and Semantic Diversity contribute to the parser performance mainly by alleviating the bias of the training set. Excluding Translation Bias does not influence the lexical diversity, while the lexical similarity between the training and test sets becomes lower. Removing Semantic Diversity drops the lexical similarity as well. But it drops lexical diversity more seriously when the sampling rates are high. 



Removing Translation Error significantly decreases the parser accuracy and BT Discrepancy rate in the low sampling regions. However, when the selection rate increases, gaps in parser accuracies and BT Discrepancy rates close immediately. Translation Error also reduces the bias by introducing correct lexicons into the translations.

Removing Semantic Density also drops the parser performance as well. We inspect that Semantic Density contributes to parser accuracy mainly by combing with the Translation Error term. As in~\fref{fig:only_al_result}, using Translation Error or Semantic Density independently results in inferior parser performance. We probe that Translation Error tends to select tail utterances from the sparse semantic region. 

\begin{figure}[ht]
    \centering
    \includegraphics[width=0.9\textwidth]{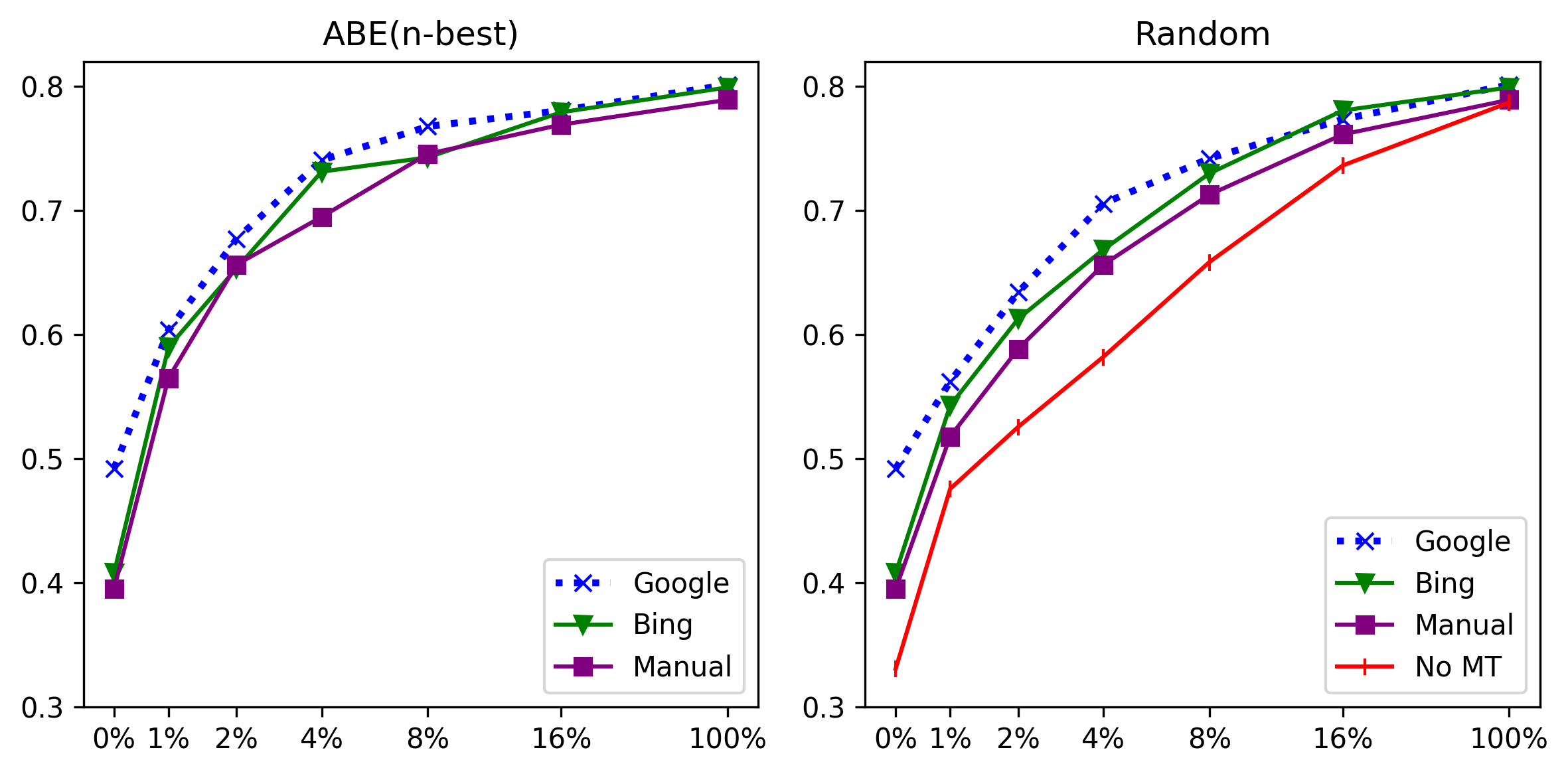}
    \caption{The parser accuracies at different query rounds on the test set of \geode using \amspnbest (Left) and \rand (Right) with different MT systems or using no MT data. }
    \label{fig:mt_al_result}
 
\end{figure}

\paragraph*{Influence of MT Systems.}
As in~\fref{fig:mt_al_result} (Right), at all query rounds, the multilingual parsers perform better with MT data in the training set, showing that MT data is essential for improving the parser's performance when a large number of HTs is not feasible. The quality of the MT data also significantly influences the parser performance when having no HT data in the training set. The accuracy difference between the parsers using Google and Bing translated data is greater than 10\% when AL has not been performed. However, after obtaining the HT data by \hatt, the performance gaps close immediately, although the MT data of better quality brings slightly higher performance. When all utterances are translated by humans, the performance differences between parsers with different MT systems can be negligible.

\fref{fig:mt_al_result} also demonstrates that \amspnbest outperforms \rand, a strong acquisition baseline, with all three different MT systems. \amspnbest is also more robust to the MT systems than \rand. The performance gaps for the parsers with \amspnbest are much smaller than those with \rand when applying different MT systems.

\section{Summary}
This chapter addresses the issues of data imbalance when adapting a multilingual semantic parser to a low-resource language. To address this issue, this chapter assumes a realistic scenario involving an MT system and budget constraints for human annotation. In particular, this chapter presents an AL method to collect a small amount of human-translated data within an abundant supply of machine-translated data to reduce the bias and error of the overall training data. Experiments demonstrate that by manually translating only 16\% of the total dataset, a parser trained on such mixture data can achieve superior performance compared to parsers trained solely on machine-translated data and comparable performance to the \oracle parser trained on a complete human-translated set.



\part{Limited Computational Resource for Semantic Parsing}
\label{part3}
\chapter{Continual Learning for Semantic Parsing}
\label{chap:continual}
\epigraph{Anyone who stops learning is old, at twenty or eighty. Anyone who keeps learning stays young.}{\textit{Henry Ford}}

\fbox
{
\begin{minipage}{0.95\textwidth}
     This chapter is based on:\\
     Z. Li, L. Qu, G. Haffari. ``Total Recall: a Customized Continual Learning Method for Neural Semantic Parsers'' The 2021 Conference on Empirical Methods in Natural Language Processing (EMNLP). 2021. 
\end{minipage}
}

To answer RQ5, this chapter investigates CL for semantic parsing when only limited computational resources are available. In this setting, a neural semantic parser learns tasks sequentially without accessing full training data from previous tasks. As described in~\cref{sec:back_cl}, direct application of the SOTA CL algorithms to this problem fails to achieve comparable performance with re-training models with all seen tasks, because they have not considered the special properties of structured outputs, yielded by semantic parsers. Therefore, we propose \tr, a CL method designed for neural semantic parsers from two aspects: i) a sampling method for memory replay that diversifies LF templates and balances distributions of parse actions in a memory;
ii) a two-stage training method that significantly improves the generalization capability of the parsers across tasks. We conduct extensive experiments to study the research problems involved in continual semantic parsing and demonstrate that a neural semantic parser trained with \tr achieves superior performance than the one trained directly with the SOTA continual learning algorithms, and achieves a 3-6 times speedup compared to retraining from scratch. 

\section{Introduction}
In the recent market research report published by MarketsandMarkets\footnote{https://www.marketsandmarkets.com/PressReleases/smart-speaker.asp}, it is estimated that the smart speaker market is expected to grow from USD 7.1 billion in 2020 to USD 15.6 billion by 2025. Commercial smart speakers, such as Alexa and Google Assistant, often need to translate users' commands and questions into actions. Therefore, semantic parsers are widely adopted in dialogue systems to map NL utterances to executable programs or LFs~\citep{damonte2019practicalSemanticParsing,rongali2020seq2seqSemanticParsing}. Due to the increasing popularity of such speakers, software developers have implemented a large volume of skills for them, and the number of new skills grows quickly every year. For instance, as of 2020, the ecosystem of Amazon Alexa voice assistant boasted over 100,000 unique `skills,' where each skill refers to a specific task that Alexa can perform. Moreover, an average of 24 new skills were being introduced daily during that year\footnote{https://voicebot.ai/2020/10/25/amazon-alexa-skill-growth-has-slowed-further-in-2020/}. Although machine learning-based semantic parsers achieve SOTA performance, they face the following challenges due to the fast-growing number of tasks. 

Given new tasks, one common practice is to retrain the parser from scratch on the training data of all seen tasks. However, it is both economically and computationally expensive to retrain semantic parsers because of a fast-growing number of new tasks~\citep{lialin2020clNeuralSemanticParsing}. To achieve its optimal performance, training a deep model on all 8 tasks of NLMap~\citep{lawrence2018improving} takes approximately 14 times longer than training the same model on one of those tasks. 
In practice, the cost of repeated retraining for a commercial smart speaker is much higher, e.g. Alexa needs to cope with the number of tasks which is over 10,000 times more than the one in NLMap\footnote{A rough estimation: retraining of our semantic parser for 100,000 tasks will take more than 138 days (2 mins of training time per NLMap task$\times$100,000) on a single-GPU machine.}. In contrast, \textit{CL} provides an alternative cost-effective training paradigm, which learns tasks sequentially without accessing full training data from the previous tasks, such that the computational resources are utilized only for new tasks. 

Privacy leakage has gradually become a major concern in many Artificial Intelligence (AI) applications. Recently, on-device machine learning has been a promising option for promoting privacy by training the model on data stored on local devices (e.g., mobile phones) with limited memory. Although training one parser per task could prevent catastrophic forgetting, in most on-device occasions, the limited memory can neither store the data of all the tasks for retraining nor multiple running parsers for different tasks. In addition, as most computing environments are not 100\% safe, it is not desirable to always keep a copy of the training data including identifiable personal information. Thus, it is almost not feasible to assume that complete training data of all known tasks is always available for retraining a semantic parser~\citep{irfan2021lifelong}. For the semantic parser of a privacy-sensitive AI system, e.g. personalized social robot, \textit{CL} provides a solution to maintain the knowledge of all learned tasks when the complete training data of those data is not available anymore due to security reasons. 
%


A major challenge of CL lies in \textit{catastrophic forgetting} that the (deep) models easily \textit{forget} the knowledge learned in the previous tasks when they learn new tasks~\citep{french1991using,mi2020continual}. Another challenge is to learn what kind of knowledge the tasks share in common and support fast adaptation of models for new tasks. Methods are developed to mitigate catastrophic forgetting~\citep{lopez2017gradient,han2020continual} and facilitate forward knowledge transfer~\citep{li2017learning}. Instead of directly measuring the speedup of training, those methods assume that there is a small fixed-size memory available for storing training examples or parameters from the previous tasks. The memory limits the size of training data, thus proportionally reducing training time. However, we empirically found that direct application of those methods to neural semantic parsers leads to a significant drop in test performance on benchmark datasets compared to retraining them with all available tasks each time.

In this chapter, we investigate the applicability of existing CL methods to semantic parsing in-depth, and we have found that most methods have not considered the special properties of structured outputs, distinguishing semantic parsing from the multi-class classification problem. Therefore, we propose \tr (\TR), a CL method specially designed to address the semantic parsing problems from two perspectives. First, we customize the sampling algorithm for memory replay, which stores a small sample of examples from each previous task when continually learning new tasks. The corresponding sampling algorithm called \textbf{D}iversified \textbf{L}ogical \textbf{F}orm \textbf{S}election (\dlfs), diversifies LF templates and maximizes the entropy of the parse action distribution in memory. 
Second, motivated by findings from cognitive neuroscience~\citep{goyal2020inductiveBias}, we facilitate knowledge transfer between tasks by proposing a two-stage training procedure, called \textbf{F}ast \textbf{S}low \textbf{C}ontinual \textbf{L}earning (\awu). It updates only unseen action embeddings in the fast-learning stage and all model parameters in the follow-up stage. As a result, it significantly improves the generalization capability of parsing models.

Our key contributions are as follows:
\begin{itemize}
    \item We conduct the \textit{first} in-depth empirical study of the problems encountered by neural semantic parsers to learn a \textit{sequence} of tasks continually in various settings. The most related work~\citep{lialin2020clNeuralSemanticParsing} only investigated incremental learning between two semantic parsing tasks.
    \item We propose \dlfs, a sampling algorithm for memory replay that is customized for semantic parsing. As a result, it improves the best sampling methods of memory replay by 2-11\% on Overnight~\citep{wang2015overnight}.
    \item We propose a two-stage training algorithm, coined \awu, that improves the test performance of parsers across tasks by 5-13\% in comparison with using only Adam~\citep{kingma2014adam}.
    \item In our extensive experiments, we investigate the applicability of the SOTA CL methods to semantic parsing with \textit{three} different task definitions, and show that \TR outperforms the competitive baselines by 4-9\% and achieves a speedup by 3-6 times compared to training from scratch. 
\end{itemize}

\section{Semantic Parser}

A semantic parser learns a mapping $\pi_{\bm{\theta}} : \mathcal{X} \rightarrow \mathcal{Y}$ to convert an NL utterance $\vx \in \mathcal{X}$ into its corresponding LF $\vy \in \mathcal{Y}$. Most SOTA neural semantic parsers formulate this task as translating a word sequence into an output sequence, whereby an output sequence is either a sequence of LF tokens or a sequence of parse actions that construct an LF. For a fair comparison between different CL algorithms, we adopt the same base model for them, as commonly used in prior works~\citep{lopez2017gradient,wang2019sentence,han2020continual}.

Similar to the one in~\cref{chap:meta}, the base parser converts the utterance $\vx$ into a sequence of actions $\va = \{a_1, ..., a_t\}$. 
As an LF can be equivalently parsed into an abstract syntax tree (AST),
the actions $\va$ sequentially construct an AST deterministically 
in the depth-first order, wherein each action $a_t$ at time step $t$ either i) expands an intermediate node according to the production rules from a grammar, 
or ii) generates a leaf node. Applying the template normalization described in~\cref{chap:meta}, the idioms (frequently occurring AST fragments) are collapsed into single units. 
The AST is further mapped back to the target LF. 

The parser employs the attention-based sequence-to-sequence (\seqseq) architecture~\citep{luong2015effective} for estimating action probabilities.

\vspace{-3mm}
\begin{small}
\begin{equation}
    P(\va | \vx ) = \prod_{t = 1}^{|\va|} P(a_t | \va_{<t},\vx)
\end{equation}
\end{small}

\paragraph*{Encoder.} The encoder in \seqseq is a standard bidirectional Long Short-term Memory (LSTM) network~\citep{hochreiter1997long}, which encodes an utterance $\vx$ into a sequence of contextual word representations.


\paragraph*{Decoder.} 
The decoder applies an LSTM to generate action sequences. At time $t$, the decoder produces an action representation $\mathbf{s}_t$, which is yielded by concatenating the hidden representation $\mathbf{h}_t$ produced by the LSTM and the context vector $\mathbf{o}_t$ produced by the soft attention~\citep{luong2015effective}.

We maintain an embedding for each action in the embedding table. The probability of an action $a_t$ is estimated by:

\vspace{-3mm}
\begin{small}
\begin{equation}
\label{eq:action_prob_cl}
 P(a_t | \va_{<t},\vx) =  \frac{\exp(\mathbf{c}_{a_t}^{\intercal} \mathbf{s}_t)}{\sum_{a' \in \mathcal{A}_t}\exp(\mathbf{c}_{a'}^{\intercal} \mathbf{s}_t)}
\end{equation}
\end{small}
where $\mathcal{A}_t$ is the set of applicable actions at time $t$, and $\mathbf{c}_{a}$ is the embedding of the action $a_{t}$, which is referred to as \textit{action embedding} in the following.
\section{Continual Semantic Parsing}
\paragraph*{Problem Formulation.}
We consider a widely adopted CL setting~\citep{lopez2017gradient,han2020continual} that a parser $\pi_{\bm{\theta}}$ is trained continually on a sequence of $K$ distinct tasks \{$\mathcal{T}^{(1)}$, $\mathcal{T}^{(2)}$,...,$\mathcal{T}^{(K)}$\}. 
In both training and testing, we know which task an example belongs to. As the definition of tasks is application-specific and parallel data of semantic parsing is often created by domain experts, it is easy to identify the task of an example in practice. We further assume that there is a fixed-size memory $\mathcal{M}_{k}$ associated with each task $\mathcal{T}^{(k)}$ for e.g. storing a small number of replay instances, as adopted in~\citet{rebuffi2017icarl,wang2019sentence}. This setting is practical for personalized conversational agents because it is difficult for them to re-collect past information except by reusing the ones in their memories.

\subsection{Challenges}
\label{sec:challenges}
We demonstrate catastrophic forgetting in continual semantic parsing by training the base parser sequentially with each task from the \overnight corpus~\citep{wang2015overnight} and report the test accuracy of exactly matched LFs of all seen tasks combined (More evaluation details are in \sref{sec:cl_exp}).

\begin{figure}[ht]
     
    \centering
    \includegraphics[width=0.9\textwidth]{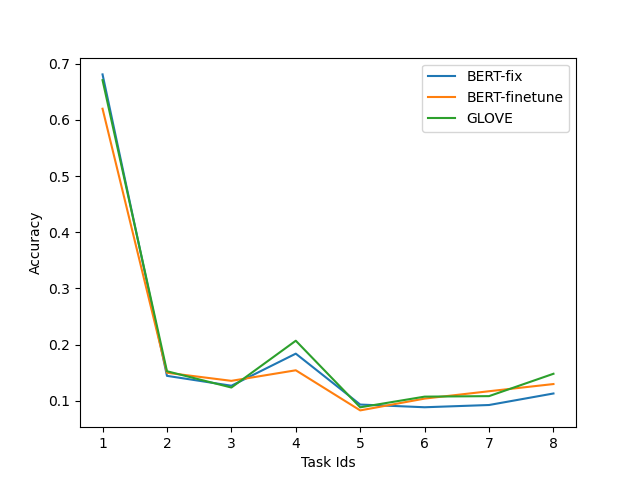}
    \caption{The accuracy on \overnight after the parser being trained on each task. The parser uses GLOVE~\citep{pennington2014glove}, BERT~\citep{devlin2019bert} with parameters updated (BERT-fine-tune) and fixed (BERT-fix) as the input embedding layer. }
    \label{fig:all_embeddings_acc_overnight}

\end{figure}
\fref{fig:all_embeddings_acc_overnight} shows the performance of continually training the base parser with BERT~\citep{devlin2019bert} (with and without fine-tuning BERT parameters) and GlOVE respectively by using the standard cross entropy loss. The accuracy on the combined test set drops dramatically after learning the \textit{second} task. The training on the initial task appears to be crucial in mitigating catastrophic forgetting. The BERT-based parser with/without fine-tuning obtains no improvement over the one using GLOVE. The forgetting with BERT is even more serious compared with using GLOVE. The same phenomenon is also observed in~\citet{arora2019forgettingBaldwin} that the models with pre-trained language models obtain inferior performance than LSTM or CNN, when fine-tuning incrementally on each task. They conjecture that it is more difficult for models with large capacity to mitigate catastrophic forgetting. 

\begin{figure}[ht]
    
    \centering
    \includegraphics[width=0.9\textwidth]{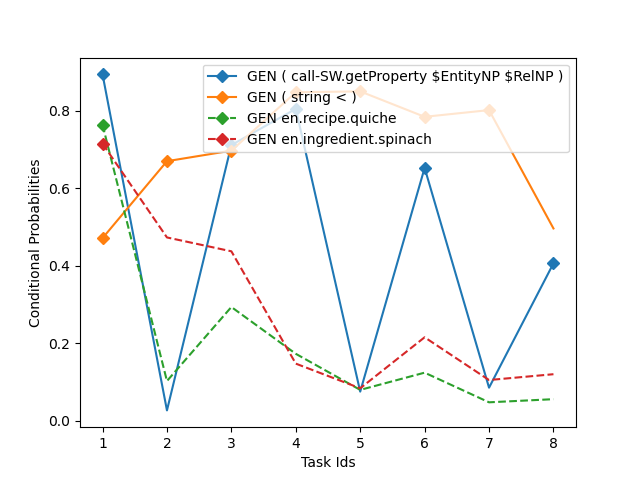}
    \caption{The average conditional probabilities $P(a_t | \va_{<t},\vx)$ of the representative cross-task (solid) and task-specific (dash) actions till the seen tasks on \overnight after learning on each task sequentially. The boxes at $i$th task indicate the actions from the initial task also exist in the $i$th task. }
   \label{fig:agg_acc_overnight}
         
\end{figure}
We further investigate which parse actions are easy to forget. To measure the degree of forgetness of the actions, after training the parser in the first task, we average the probabilities $P(a_t | \va_{<t},\vx)$ produced by the parser on the training set of the first task. We recompute the same quantity after learning each task sequentially and plot the measures. \fref{fig:agg_acc_overnight} depicts the top two and the bottom two actions are easiest to forget on average. Both top two actions appear only in the first task. Thus, it is difficult for the parser to remember them after learning new tasks. In contrast, cross-task actions, such as \textit{GEN ( string < )}, may even obtain improved performance after learning on the last task. Thus, it indicates the importance of differentiating between task-specific actions and cross-task actions when designing novel CL algorithms.

\subsection{Total Recall}
To save training time for each new task, we cannot use all training data from previous tasks. Thus, we introduce a designated sampling method in the sequel to fill memories with the examples most likely mitigating catastrophic forgetting. We also present the two-stage training algorithm \awu to facilitate knowledge transfer between tasks.

\paragraph*{Sampling Method.}

\dlfs improves Episodic Memory Replay (\emr)~\citep{wang2019sentence,chaudhry2019tiny} by proposing a designated sampling method for continual semantic parsing. \emr utilizes a memory module $\mathcal{M}_{k} = \{(\vx_1^{(k)},\vy_1^{(k)}),...,(\vx_M^{(k)},\vy_M^{(k)})\}$ to store a few examples sampled from the training set of task $\mathcal{T}^{(k)}$, where $(\vx_m^{(k)},\vy_m^{(k)}) \in \mathcal{D}_{train}^{(k)}$ and $M$ is the size of the memory. The training loss of \emr takes the following form:
\begin{equation}
    \mathcal{L}_{\emr} = \mathcal{L}_{\mathcal{D}_{train}^{(k)}} + \sum_{i=1}^{k-1}\mathcal{L}_{\mathcal{M}_{i}}
\end{equation}
where $\mathcal{L}_{\mathcal{D}_{train}^{(k)}}$ and $\mathcal{L}_{\mathcal{M}_{i}}$ denotes the loss on the training data of current task $\mathcal{T}^{(k)}$ and the memory data from task $\{\mathcal{T}\}^{(k-1)}_{i=1}$, respectively. The training methods for memory replay often adopt a subroutine called \textit{replay training} to train models on instances in the memory. Furthermore, prior works~\citep{aljundi2019gradient,wang2019sentence,han2020continual,mi2020continual,chrysakis2020online,kim2020imbalanced} discovered that storing a small amount of \textit{diversified} and \textit{long-tailed} examples helps tackle catastrophic forgetting for memory-based methods. Semantic parsing is a structured prediction problem. We observe that semantic parsing datasets are highly imbalanced w.r.t. LF structures. Some instances with similar LF structures occupy a large fraction of the training set. Therefore, we presume storing the diversified instances in terms of the corresponding LF structures would alleviate the problem of catastrophic forgetting in continual semantic parsing. 

To sample instances with the diversified LF structures, our method \dlfs partitions the LFs in $\mathcal{D}_{train}$ into $M$ clusters, followed by selecting representative instances from each cluster to maximize the entropy of actions in memory. To characterize differences in structures, we first compute similarities between LFs by $sim(\vy_i,\vy_j) = (Smatch(\vy_i,\vy_j) + Smatch(\vy_j,\vy_i))/2$, where \textit{Smatch}~\citep{cai2013smatch} is an asymmetrical similarity score between two LFs yielded by calculating the overlapping percentage of their triples. Then, we run a flat clustering algorithm using the distance function $1- sim(\vy_i,\vy_j)$ and the number of clusters is the same as the memory size. We choose \textit{K-medoids}~\citep{park2009simple} in this work to easily interpret clustering results.

We formulate the problem of balancing action distribution and diversifying LF structures as the following constrained optimization problem. In particular, it i) aims to balance the actions of stored instances in the memory module $\mathcal{M}$ by increasing the entropy of the action distribution, and ii) requires that each instance $m$ in $\mathcal{M}$ belongs to a different cluster $c_j$. Let the function $c(m)$ return the cluster id of an instance in a memory $\mathcal{M}$ and $m_i$ denote its $i$th entry, we have
\begin{align}\nonumber
\label{eq:sample}
 \max_{\mathcal{M}}& -\sum_{a_{i} \in \mathcal{A}} p_{\mathcal{M}}(a_{i})\log p_{\mathcal{M}}(a_{i})\\
    s.t. & \forall m_i, m_j \in \mathcal{M}, c(m_i) \neq c(m_j)
\end{align}
\noindent where $p_{\mathcal{M}}(a_{i})=\frac{n_{i}}{\sum_{{a_j} \in \mathcal{A}} n_{j}}$, with $n_i$ being the frequency of action $a_{i}$ in $\mathcal{M}$ and $\mathcal{A}$ being the action set included in the training set $\mathcal{D}_{train}$. In some occasions, the action set $\mathcal{A}$ is extremely large (e.g. 1000+ actions per task), so it may be infeasible to include all actions in the limited memory $\mathcal{M}$. We thus sample a subset of $h$ actions, $\mathcal{A}' \subseteq \mathcal{A}$, given the distribution of $P_{D}(\mathcal{A})$ in $\mathcal{D}_{train}$ where $p_{D}(a_{i})=\frac{n_{i}}{\sum_{{a_j} \in \mathcal{A}} n_{j}}$, with $n_i$ being the frequency of $a_{i}$ in $\mathcal{D}_{train}$. In that case, our method addresses the optimization problem over the actions in $\mathcal{A}'$. We solve the above problem by using an iterative updating algorithm, whose details can be found at~\algoref{algo:sample}. The closest works~\citep{chrysakis2020online,kim2020imbalanced} maintain only the balanced label distribution in the memory, while our work maintains the balanced memory w.r.t. both the LF and action distributions. 

\begin{algorithm}[t]
{\small
\SetKwData{Left}{left}\SetKwData{This}{this}\SetKwData{Up}{up}
\SetKwFunction{Union}{Union}\SetKwFunction{FindCompress}{FindCompress}
\SetKwInOut{Input}{Input}\SetKwInOut{Output}{Output}
\SetAlgoLined
\Input{Training set $\mathcal{D}$, memory size $M$}
\Output{The memory $\mathcal{M}$}
     Partition $\mathcal{D}$ into $M$ clusters, denoted as $C$, with the similarity/distance metric\\
    Randomly sample the memory  $\mathcal{M}$ of size $M$ from $D$\\
    $H_{old} \gets -\inf$\\
     Compute the entropy $H_{new}$ of the action distribution $P_{\mathcal{M}}(\mathcal{A})$ in $\mathcal{M}$ as in Eq. \ref{eq:sample}\\
     \While {$H_{new} > H_{old}$}{
    \For{$i$-th cluster $c_i \in \mathcal{C}$}{
     Compute the entropy $H_{c}$ of $P_{\mathcal{M}}(\mathcal{A})$\\
    \For{$instance\ n \in c_i$}{
        Replicate $\mathcal{M}$ with $\mathcal{M}'$\\
        Replace the $i$-th instance $m_i$ in $\mathcal{M}'$ with $n$\\
        Compute the entropy $H'_{c}$ of $\mathcal{M}'$\\
        \If{$H'_{c} > H_{c}$}
        {
        $\mathcal{M} \gets \mathcal{M}'$\\
        $E_{c} \gets H'_{c}$
        }
    }
    $H_{old} \gets H_{new}$\\
    Compute the entropy $H_{new}$ of $P_{\mathcal{M}}(\mathcal{A})$\\
    }
    }
}
\caption{\blfst
}
\label{algo:sample}
\end{algorithm}

\paragraph*{Fast-Slow CL.}
CL methods are expected to learn what the tasks have in common and in what the tasks differ. If there are some shared structures between tasks, it is possible to transfer knowledge from one task to another. Inspired by findings from cognitive neuroscience, the learning should be divided into slow learning of stationary aspects between tasks and fast learning of task-specific aspects~\citep{goyal2020inductiveBias}. This is an inductive bias that can be leveraged to obtain cross-task generalization in the space of all functions.

We implement this inductive bias by introducing a two-stage training algorithm. In the base model, action embeddings $\mathbf{c}_a$ (Eq. \eqref{eq:action_prob_cl}) are task-specific, while the remaining parts of the model, which builds representations of utterances and action histories, are shared to capture common knowledge between tasks. Thus, in the \textit{fast-learning} stage, we update only the embeddings of unseen actions $\mathbf{c}^{(i)}_{a}$ with the cross-entropy loss, in the \textit{slow-learning} stage, we update all model parameters. Fast-learning helps parsers generalize to new tasks by giving the unseen actions good initialized embeddings and reduces the risk of forgetting prior knowledge by focusing on minimal changes of model parameters for task-specific patterns.

In the fast-learning stage, the unseen actions $A^{(k)}_{u}$ of the $k$-th task are obtained by excluding all historical actions from the action set of current task $\mathcal{T}^{(k)}$, namely $A^{(k)}_{u} = A^{(k)}\setminus A^{(1:k-1)}$, where $A^{(k)}$ denotes the action set of the $k$-th task. All actions are unseen in the first task, thus we update all action embeddings by having $A^{(0)}_{u} = A^{(0)}$. In the slow-learning stage, we differ updating parameters w.r.t. current task from updating parameters w.r.t. memories of previous tasks. For the former, the parameters $\bm{\theta}_g$ shared across tasks are trained w.r.t all the data while the task-specific parameters $\bm{\theta}^{(i)}_s$ are
trained only w.r.t. the data from task $\mathcal{T}^{(i)}$. For the latter, the task-specific parameters learned from the previous tasks are frozen to ensure they do not forget what is learned from previous tasks. 
\begin{algorithm}[t]
{\small
\SetKwData{Left}{left}\SetKwData{This}{this}\SetKwData{Up}{up}
\SetKwFunction{Union}{Union}\SetKwFunction{FindCompress}{FindCompress}
\SetKwInOut{Input}{Input}\SetKwInOut{Output}{Output}
\SetAlgoLined
\Input{Training set $\mathcal{D}^{(k)}_{train}$ of $k$-th task $\mathcal{T}^{(k)}$, memory data $\mathcal{M} = \{\mathcal{M}_1,...,\mathcal{M}_{k-1}\}$, the known action set $\mathcal{A}^{(1:k-1)}$ before learning $\mathcal{T}^{(k)}$}
Extract action set $\mathcal{A}^{(k)}$ from $\mathcal{D}^{(k)}_{train}$\\
Obtain unseen actions $\mathcal{A}^{(k)}_u$ by excluding $\mathcal{A}^{(1:k-1)}$ from $\mathcal{A}^{(k)}$\\
\textcolor{blue}{\# \textit{fast-learning on unseen action embeddings}}\\
\For{$i \gets 1$ to $epoch_1$} 
{
Update $\mathbf{c}^{(k)}_{a}$ with $\nabla_{\mathbf{c}^{(k)}_{a}}\mathcal{L}(\bm{\theta}_{g},\bm{\theta}^{(k)}_{s})$ on $\mathcal{D}^{(k)}_{train}$
}
\textcolor{blue}{\# \textit{slow-learning stage}}\\
    \For{$i \gets 1$ to $epoch_2$}{
    \textcolor{blue}{\# \textit{fine-tune all cross-task parameters and task-specific parameters of the current task}}\\
     Update $(\bm{\theta}_{g},\bm{\theta}^{(k)}_{s})$ with $\nabla_{(\bm{\theta}_{g},\bm{\theta}^{(k)}_{s})}\mathcal{L}(\bm{\theta}_{g},\bm{\theta}^{(k)}_{s})$ on $\mathcal{D}^{(k)}_{train}$\\
     \textcolor{blue}{\# \textit{replay training with task-specific parameters of the previous tasks frozen}}\\
    \For{$\mathcal{M}_i \in \mathcal{M}$}{
        Update $\bm{\theta}_{g}$ with $\nabla_{\bm{\theta}_{g}}\mathcal{L}(\bm{\theta}_{g},\bm{\theta}^{(i)}_{s})$ on $\mathcal{M}^{(i)}$\\
    }
    }
}
\caption{Fast-Slow Training for the $k$-th task }
\label{algo:adapter}
\end{algorithm}
More details can be found in \algoref{algo:adapter}.
This training algorithm is closely related to Invariant Risk Minimization~\citep{arjovsky2019invariant}, which learns invariant structures across different training environments. However,~\citet{arjovsky2019invariant} assume the same label space across environments and have access to all training environments at the same time. 

\paragraph*{Loss} During training, we augment the \emr loss with the Elastic Weight Consolidation (\ewc) regularizer~\citep{kirkpatrick2017overcoming} to obtain the training loss $L_{CL} = L_{EMR} + \Omega_{EWC}$, where
$
    \Omega_{\ewc} = \lambda \sum^{N}_{j=1} F_{j}(\bm{\theta}_{k,j}-\bm{\theta}_{k-1,j})^{2}
$
, $N$ is the number of model parameters, $\bm{\theta}_{k-1,j}$ is the model parameters learned until $\mathcal{T}^{(k-1)}$ and $F_{j} = \nabla^{2}\mathcal{L}(\bm{\theta}_{k-1,j})$ w.r.t. the instances stored in $\mathcal{M}$. \ewc slows down the updates of parameters that are crucial to previous tasks according to the importance measure $F_{j}$.
\section{Experiments}

\label{sec:cl_exp}
\paragraph*{Datasets and Task Definitions.}
In this work, we consider three different scenarios: 

\begin{itemize}
    \item Different tasks are in different domains, and there are task-specific predicates and entities in LFs.
    \item There are task-specific predicates in LF templates.
    \item There are a significant number of task-specific entities in LFs.
\end{itemize}
All tasks in the latter two are in the same domain. Unlike previous chapters, where smaller datasets were sufficient, our current study demands larger datasets that can be partitioned into multiple tasks. To meet this requirement, we select Overnight~\citep{wang2015overnight} and NLMapV2~\citep{lawrence2018improving} to simulate the proposed three CL scenarios, coined \overnight, \nlmapq and \nlmapc, respectively.


Overnight includes around 18,000 queries involving eight domains. The data in each domain includes 80\% training instances and 20\% test instances. Each domain is defined as a task. 

NLMapV2 includes 28,609 queries involving 79 cities and categorizes each query into one of 4 different question types and their sub-types. In the \nlmapq setting, we split NLMapV2 into 4 tasks with queries in different types. In the setting of \nlmapc, NLMapV2 is split into 8 tasks with queries of 10 or 9 distinct cities in each task. Each city includes a unique set of point-of-interest regions. In both \nlmapc and \nlmapq, each task is divided into 70\%/10\%/20\% of training/validation/test sets, respectively.

We attribute different distribution discrepancies between tasks to different definitions of tasks. Overall, the distribution discrepancy between tasks on \overnight is the largest while the tasks in the other two settings share relatively smaller distribution discrepancies because tasks of \nlmapq and \nlmapc are all in the same domain.

\paragraph*{Baselines.}
Our proposed method is compared with 8 CL baselines and 1 Oracle setting. 

\begin{itemize}
    \item \textbf{FINE-TUNE} fine-tunes the model on the new tasks based on previous models.
    \item \textbf{EWC}~\citep{kirkpatrick2017overcoming} adds an L2 regularization to slow down the update of model parameters important to the historical tasks. 
    \item \textbf{HAT}~\citep{serra2018overcoming} activates a different portion of parameters for each task with task-specific mask functions.
    \item \textbf{GEM}~\citep{lopez2017gradient} stores a small number of instances from previous tasks and uses the gradients of previous instances as the constraints on directions of gradients w.r.t. current instances.
    \item  \textbf{EMR}~\citep{chaudhry2019tiny,wang2019sentence} trains the model on the data from the current task along with mini-batches of memory instances. 
    \item \textbf{EMAR}~\citep{han2020continual} is an extension of \emr using memory instances to construct prototypes of relation labels to prohibit the model from overfitting on the memory instances.
    \item  \textbf{ARPER}~\citep{mi2020continual} adds an adaptive \ewc regularization on the \emr loss, where the memory instances are sampled with a unique sampling method called \prior. 
    \item \textbf{ProtoParser} is the parser we proposed in~\cref{chap:meta}. It utilizes prototypical networks~\citep{snell2017prototypical} to improve the generalization ability of semantic parsers on the unseen actions in the new task. We customize it by training the \proto on the instances on current task as well as the memory instances. 
     \item \textbf{ORACLE} (All Tasks) setting trains the model on the data of all tasks combined, considered as an upper bound of CL.
\end{itemize}


\paragraph*{Evaluation.} To evaluate the performance of continual semantic parsing, we report the accuracy of exactly matched LFs. We further adopt two common evaluation settings in CL. 
One setting measures the performance by averaging the accuracies of the parser on test sets of all seen tasks $\{\mathcal{D}_{test}^{(1)}$,...,$\mathcal{D}_{test}^{(k)}\}$ after training the model on
task $\mathcal{T}^{(k)}$, i.e. $\text{ACC}_{\text{avg}} = \frac{1}{k} \sum^{k}_{i=1} acc_{i,k}$~\citep{lopez2017gradient,wang2019sentence}.
The other one evaluates the test sets of all seen tasks combined after finishing model training on task $\mathcal{T}^{(k)}$, $
    \text{ACC}_{\text{whole}} = acc_{\mathcal{D}^{(1:k)}_{test}}
$~\citep{wang2019sentence,han2020continual}.

\paragraph*{Implementation Details}
\label{sec:repro}
The hyper-parameters are cross-validated on the training set of \overnight and validated on the validation set of \nlmapq and \nlmapc. We train the semantic parser on each task with a learning rate of 0.0025, batch size of 64 and for 10 epochs. The fast-learning training epochs are 5. We use the 200-dimensional GLOVE embeddings~\citep{pennington2014glove} to initialize the word embeddings for utterances. As different task orders influence the performance of the continual semantic parsing, all experiments are run on 10 different task orders with a different seed for each run. We report the average $\text{ACC}_{\text{avg}}$ and $\text{ACC}_{\text{whole}}$ of 10 runs. 
In addition, we use one GPU of Nvidia V100 to run all our experiments. The sizes of hidden states for LSTM and the action embeddings are 256 and 128, respectively. The default optimizer is Adam~\citep{kingma2014adam}. For our \dlfs method, we sample the subsets of the action sets with sizes 300 and 500 on \nlmapc and \nlmapq, respectively. The number of our model parameters is around 1.8 million. We grid-search the training epochs from \{10,50\}, the learning rate from \{0.001,0.0025\}. The coefficient for \ewc is selected from \{50000, 200000\}. Here we also provide the experiment results on the validation sets as in~\tref{tab:valid}.
\begin{table}[ht]
\centering
  \resizebox{0.65\textwidth}{!}{%
  \begin{tabular}{|c|c c|c c|cc|}
    \toprule
    \multirow{2}{*}{Methods} &
      \multicolumn{2}{c|}{\overnight} &
      \multicolumn{2}{c|}{\nlmapq} &
      \multicolumn{2}{c|}{\nlmapc} \\
      & W  & A & W  & A & W  & A \\
      \midrule
    \hline
    \aemrt & 52.38 & 53.30 & 68.78 &66.83  & 68.51  & 66.23 \\
    \aemrt (+\ewc) & 56.84 & 54.85  & 70.74 &69.21 & 74.84  & 71.19 \\
    \hline
    \bottomrule
  \end{tabular}%
  }
    \caption{Results on the validation sets.}
  \label{tab:valid}
\end{table}

\begin{table}[ht]
    \vspace{-2mm}
\centering
  \resizebox{0.95\textwidth}{!}{%
  \begin{tabular}{|c|c c|c c|cc|}
    \toprule
    \multirow{2}{*}{Methods} &
      \multicolumn{2}{c|}{\overnight} &
      \multicolumn{2}{c|}{\nlmapq} &
      \multicolumn{2}{c|}{\nlmapc} \\
      & W  & A & W  & A & W  & A \\
      \midrule
    Fine-tune & 14.40 & 14.37 &  60.22  & 55.53  & 49.29  & 48.11  \\
        \hline
    \ewc & 38.57 & 40.45 & 65.44 & 62.25  & 59.28& 57.56 \\
    \hatt & 15.45 & 15.71 & 64.69 & 60.88 & 53.30 & 52.41 \\
    \gem & 41.33  & 42.13 & 63.28 & 59.38  & 55.14 &54.37    \\
    \emr & 45.29 & 46.01 & 59.75 &55.59 & 58.36 & 56.95 \\
    \emar & 46.68 & 48.57 & 14.25 &12.89  & 51.79 & 50.93 \\
    \arper & 48.28 & 49.90  & 67.79 &64.73  & 62.62 &60.61\\
    \proto & 48.15 & 49.71  & 67.10 &63.73 & 62.58 &60.87 \\
    \hline
    \aemrt & 54.40 & 54.73 & 70.06 &67.77  & 62.96  & 61.59 \\
    \aemrt (+\ewc) & \textbf{59.02} & \textbf{58.04}  & \textbf{72.36} & \textbf{70.66} & \textbf{67.15}  & \textbf{64.89} \\
    \hline
    \oracle (All Tasks) & 63.14 & 62.76 & 73.20 & 71.65 & 69.48 & 67.18 \\
    \bottomrule
  \end{tabular}%
  }
    \caption{LF Exact Match Accuracy (\%) on two datasets with three settings after model learning on all tasks. “W” stands for the Whole performance $\text{ACC}_{\text{whole}}$, and “A” stands for the Average performance $\text{ACC}_{\text{avg}}$. All the results are statistically significant (p$<$0.005) compared with \aemrt(+\ewc) according to the Wilcoxon signed-rank test~\citep{woolson2007wilcoxon}. All experiments are run 10 times with different sequence orders and seeds. The memory size is 50 for all memory-based methods.}
  \label{tab:main}
  \vspace{-2mm}
\end{table}

\subsection{Results and Discussion}

\begin{figure}[t]
    \centering
    \includegraphics[width=0.5\textwidth]{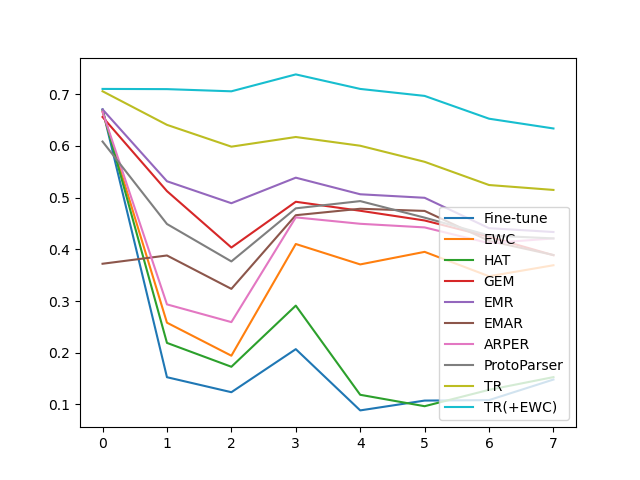}
    \caption{$\text{ACC}_{\text{whole}}$ till the seen tasks on \overnight after learning on each task sequentially. \vspace{-4mm}}
    \label{fig:acc_overnight}

\end{figure}

As shown in~\tref{tab:main}, the base parser trained with our best setting, \aemrt(+\ewc), significantly outperforms all the other baselines (p$<$0.005) in terms of both $\text{ACC}_{\text{avg}}$ and $\text{ACC}_{\text{whole}}$. The performance of \aemrt(+\ewc) is, on average, only 3\% lower than the \oracle setting. Without \ewc, \aemrt still performs significantly better than all baselines except it is marginally better than \arper and \proto in the setting of \nlmapq. From~\fref{fig:acc_overnight}, we can see that our approaches are more stable than other methods, and demonstrate less and slower forgetting than the baselines.

The dynamic architecture method, \hatt, performs worst on \overnight while achieving much better performance on \nlmapq and \nlmapc. Though the performance of the regularization method, \ewc, is steady across different settings, it ranks higher among other baselines on \nlmapc and \nlmapq than on \overnight. In contrast, the memory-based methods, \gem, and \emr, rank better on \overnight than on \nlmapq and \nlmapc. 

We conjecture that the overall performance of CL approaches varies significantly in different settings due to different distribution discrepancies as introduced in \textit{Datasets}. The general memory-based methods are better at handling catastrophic forgetting than the regularization-based and dynamic architecture methods, when the distribution discrepancies are large. However, those memory-based methods are less effective when the distribution discrepancies across tasks are small. Ideally, if we can anticipate the data distribution of upcoming tasks—especially if these future distributions are influenced by previously encountered ones—we could design an algorithm to dynamically choose the most suitable method, be it memory-based or otherwise. This represents an exciting avenue for future research. 
Another weakness of memory-based methods is demonstrated by \emar, which achieves only 14.25\% of $\text{ACC}_{\text{whole}}$ on \nlmapq, despite it being the SOTA method on continual relation extraction. A close inspection shows that the instances in the memory are usually insufficient to include all actions when the number of actions is extremely large (i.e., more than 1000 actions per task in \nlmapq) while \emar relies on instances in memory to construct prototypes for each label. 
Furthermore, large training epochs for memory-based methods usually lead to severe catastrophic forgetting of the previous tasks, while the regularization method could largely alleviate this effect.


\arper and \parser are the two best baselines. Similar to \aemrt, \arper is a hybrid method combining \emr and \ewc. Thus, the joint benefits lead to consistently superior performance over the other baselines except \parser in all three settings. 
The generalization capability to unseen actions in new tasks also seems critical in continual semantic parsing. Merely combining \proto and \emr yields a new baseline, which performs surprisingly better than most existing CL baselines. From that perspective, the parser with \awu performs well in CL also because of its strength in generalizing to unseen actions. 

We also observe the action probabilities $P(a_t | \va_{<t},\vx)$ for each method. In methods not incorporating memory, such as \finetune, we find that the conditional probabilities associated with task-specific actions typically decline consistently, with only minor fluctuations. In contrast, these probabilities tend to maintain high levels in methods like \emr and \tr across tasks. However, probabilities in \emr fluctuate more substantially, whereas the \tr method stabilizes these probabilities at a high level with fewer fluctuations. These high probabilities are expected as task-specific actions are stored in memory and retrained for future tasks.

\paragraph*{Influence of Sampling Strategies.}
\begin{table}[ht]
\centering
  \resizebox{0.95\textwidth}{!}{%
  \begin{tabular}{|c|ccc|ccc|ccc|}
    \toprule
    \multirow{2}{*}{Methods} &
      \multicolumn{3}{c|}{\overnight} &
            \multicolumn{3}{c|}{\nlmapq} &
      \multicolumn{3}{c|}{\nlmapc}
        \\
    &10 &
      25 &
      50 & 
     10 &
      25 &
      50&
     10 &
      25 &
      50\\

      \midrule
    \random & 37.63 & 45.18 & 50.63 & 65.63 & 66.20 & 67.21  &  58.82 & 59.96 & 60.64 \\
    \fss & 38.64 & 47.08 & 52.63 & 66.51 & 67.15 & 67.67 & 58.94 & 59.67 & 60.89   \\
    \gss & 34.01 & 39.45 & 43.87 & 65.89 & 66.60&67.07 & 57.97 & 59.40&60.36 \\
    \prior & 37.60  & 44.84 & 50.14  &66.08  & 67.47 &67.42 &54.39  & 54.09 &53.43 \\
    \balance & 38.21 & 45.33 & 48.64  & 65.37 & 66.21 &66.99 & 58.39 & 59.89 &61.19 \\
    \hline
    \lfs & 38.24 & 44.66 & 53.06 & 66.08 &67.20 & 67.53 & 58.88 & 59.81 & 60.76  \\
    \blfst & 39.24 & 48.21 & 54.73  & 65.21 & 66.60 & 67.77 & 59.31 & 60.58 & 61.59 \\
    \bottomrule
  \end{tabular}%
  }
    \caption{$\text{ACC}_{\text{avg}}$ (\%) of \aemr with different sampling strategies and memory sizes 10, 25 and 50.}
  \label{tab:sample}
 
\end{table}

\tref{tab:sample} reports the evaluation results of \aemrt with different sampling strategies and sizes on \overnight, \nlmapc, and \nlmapq. \textbf{RANDOM} randomly samples instances from the train set of each task. \textbf{FSS}~\citep{aljundi2019gradient,wang2019sentence,mi2020continual}, \textbf{GSS}~\citep{aljundi2019gradient} and \textbf{LFS} partition the instances into clusters w.r.t. the spaces of utterance encoding features, instance gradients, and LFs, respectively, and then select the instances which are closest to the centroids. \textbf{PRIOR}~\citep{mi2020continual} selects instances that are most confident to the models and diversifies the entities in the utterances of memory. \textbf{BALANCE}~\citep{chrysakis2020online,kim2020imbalanced} balances the action distribution in a memory.

Overall, our sampling method consistently outperforms all other baselines on both \overnight and \nlmapc. On \overnight with memory size 50, the gap between \blfst and \gss is even up to 11\% and 2\% between \blfst and \fss, the best baseline. However, on \nlmapc, the performance differences across various sampling methods are smaller than those on \overnight. 
A similar observation applies to the influence of different sample sizes. 
We conclude that the smaller distribution discrepancy reduces the differences in sampling methods as well as the sample sizes in the memory-based methods.

\random performs steadily across different settings, though it is usually in mediocre performance. \fss, \gss, and \prior are model-dependent sampling methods. The gradients and model confidence scores are not stable features for the sample selection algorithms. We inspect that the instances selected with \gss are significantly different even when model parameters are slightly disturbed. For the \prior, the semantic parsing model is usually confident to instances with similar LF templates. Diversifying entities does not necessarily lead to diversities of LF templates since the LFs with different entities may share similar templates. Therefore, \gss and \prior can only perform well in one setting. In contrast, the utterance encoding features are much more reliable. \fss can achieve the second-best performance among all methods. Either balancing action distribution (\balance) or selecting centroid LFs from LF clusters (\lfs) alone performs no better than \blfst, proving it is advantageous to select an instance in a cluster that balances the memory action distribution over directly using the centroid. 

\paragraph*{Ablation Study of FSCL Training.}

\begin{table}[ht]
    \vspace{-2mm}
\centering
  \resizebox{0.9\textwidth}{!}{%
  \begin{tabular}{|c|c c|c c|cc|}
    \toprule
    \multirow{2}{*}{Methods} &
      \multicolumn{2}{c|}{\overnight} &
      \multicolumn{2}{c|}{\nlmapq} &
      \multicolumn{2}{c|}{\nlmapc} \\
      & W  & A & W  & A & W  & A \\
      \midrule
    
    \aemrt(+\ewc) & \textbf{59.02} & \textbf{58.04}  & \textbf{72.36}  & \textbf{70.66} & \textbf{67.15}  & \textbf{64.89} \\
    \quad - fast & 56.22 & 54.77  & 69.12 & 65.97 & 64.88 & 62.42\\
    -/+ fast/lwf & 55.80 & 54.54  & 69.45 & 66.37 & 65.14 & 62.70\\
    -/+ fast/emar & 56.93 & 56.05  & 69.43 & 66.42 &  64.89 & 62.51\\
    \hline
    \aemrt & 54.40 & 54.73 & 70.06  & 67.77  & 62.96  & 61.59 \\
    \quad - fast & 49.28 & 49.48  & 60.22 & 54.84 & 57.53 & 55.75\\
    -/+ fast/lwf & 47.47 & 47.77 & 58.96 & 53.54  & 55.24 & 56.89\\
    -/+ fast/emar & 49.63 & 48.74  & 64.98 & 61.09 & 56.89 & 55.24\\
    \bottomrule
  \end{tabular}%
  }
    \caption{The ablation study results of \awu.}
  \label{tab:abl_adap}
      \vspace{-3mm}
\end{table}
\tref{tab:abl_adap} shows the ablation study of 
\awu training by removing (-) or replacing (-/+) the corresponding component/step.

\begin{figure}
    \centering
    \begin{subfigure}[b]{0.32\linewidth}
        \centering
    \includegraphics[width=\textwidth]{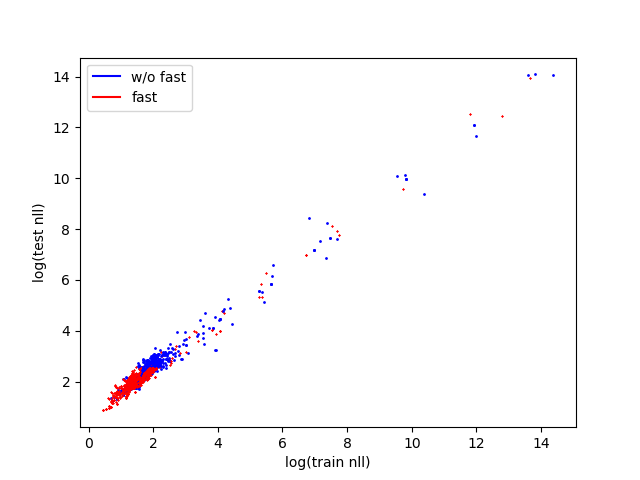}
        \label{fig:error_overnight}
    \end{subfigure}
    \begin{subfigure}[b]{0.32\linewidth}
        \centering
    \includegraphics[width=\textwidth]{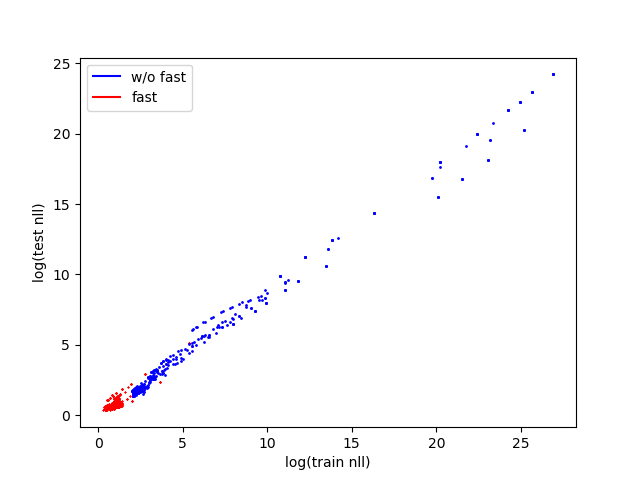}
        \label{fig:error_nlmapq}
    \end{subfigure}
        \begin{subfigure}[b]{0.32\linewidth}
        \centering
    \includegraphics[width=\textwidth]{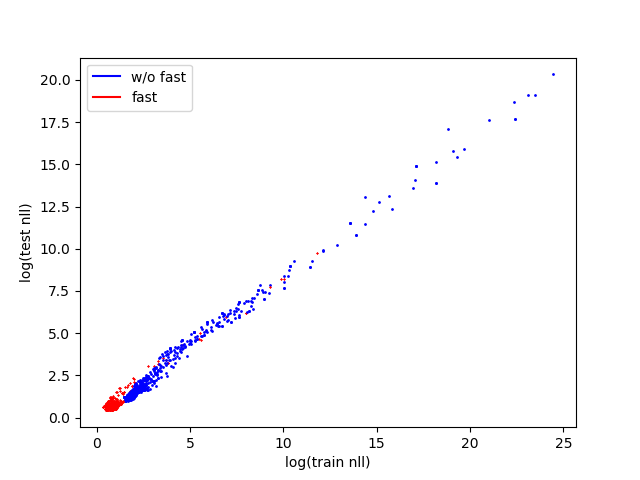}
        \label{fig:error_nlmapc}
    \end{subfigure}
    \caption{The training and test errors of the base parser with/without fast-learning on \overnight, \nlmapc, and \nlmapq. }
\label{fig:error_plot}
\end{figure}

The fast-learning with action embeddings is the most critical step in \awu training. Removing it causes up to 13\% performance drops. To study this step in depth, we also replace our fast-learning with fine-tuning all task-specific parameters except in the first task, as done in LwF~\citep{li2017learning}, or fine-tuning all parameters, as done in \emar~\citep{han2020continual}, in the fast-learning stage. The corresponding performance is no better than removing it in most cases. We also plot the training errors and test errors with or without this step in \fref{fig:error_plot}. This step clearly leads to a dramatic improvement in both generalization and optimization.

Another benefit of this fast-learning step is in the first task. We observe that a good optimization on the first task is crucial to the model learning on the following tasks. Our preliminary study shows that by applying the fast-learning only to the first task, the model can still keep a close-to-optimal performance. As shown in~\fref{fig:acc_overnight}, our method with this fast-learning step is better optimized and generalized on the initial tasks than all the other baselines and largely alleviates the forgetting problem caused by learning on the second task.

\paragraph*{Influence of Pre-trained Language Models.}
We study the impact of pre-trained language models for semantic parsing in supervised learning and CL, respectively. In both settings, we evaluate the base parsers using BERT~\citep{devlin2019bert} as its embedding layer in two configurations: fine-tuning the parameters of BERT (BERT-fine-tune) and freezing BERT's parameters (BERT-fix). As in~\tref{tab:bert}, BERT slightly improves the overall performance of the base parsers in supervised training (the \oracle setting) on \overnight. In contrast, in the CL setting, base parsers with the BERT embedding perform worse than the ones with the GLOVE embedding.
On \nlmapq, the accuracy of \finetune with GLOVE embedding is 30\% and 20\% higher than that with BERT's embedding updated and fixed, respectively. We conjecture that the deeper neural models suffer more from catastrophic forgetting. However, the average training speeds of parsers with BERT-fix and BERT-fine-tune are 5-10 times and 20-40 times, respectively, slower than those with GLOVE on each task. Overall, our method outperforms other SOTA CL methods except that \ewc with BERT-fix performs comparably with ours on \nlmapc. In contrast, the performance of \proto, the best baseline with GLOVE, is highly unstable on NLMap with BERT. 
\begin{table}[ht]
 
\centering
  \resizebox{\textwidth}{!}{%
  \begin{tabular}{|c|c c|c c|cc|}
    \toprule
    \multirow{2}{*}{Methods} &
      \multicolumn{2}{c|}{\overnight} &
      \multicolumn{2}{c|}{\nlmapq} &
      \multicolumn{2}{c|}{\nlmapc} \\
      & W  & A & W  & A & W  & A \\
            \midrule
      \multicolumn{7}{|c|}{BERT-finetune} \\
        \midrule
    Fine-tune & 16.19 & 14.89 &  30.33  & 29.22  & 39.47  & 36.07 \\
        \hline
            \ewc  & 13.97 & 13.39 &  45.97 &  44.19   & 35.74& 34.17 \\
    \emr  & 51.74 &51.40 & 37.39 & 34.61 & 54.89 & 52.61 \\
            \proto & 47.74  & 46.99 &  15.10 & 13.03 & 34.56 & 32.27 \\
    \aemrt & 52.58 & 52.86 & 64.07 &61.35  & 58.38  & 56.35 \\

    \hline
    \oracle (All Tasks) & 65.40 & 64.39 & 72.70 & 71.60 & 67.51 & 68.80 \\
                \midrule
        \multicolumn{7}{|c|}{BERT-fix} \\
                  \midrule
        Fine-tune  & 14.10 & 12.09 &  40.45  & 36.06 & 49.15  & 46.48 \\
        \hline
            \ewc  & 17.69 & 18.56 & 54.20 & 51.32   &  64.10 & 61.67  \\
    \emr& 39.10 &39.03 & 42.46& 39.50 & 57.55 &55.42 \\
    \proto & 41.69 & 41.79 & 51.39 & 48.90 & 45.66 & 43.92 \\
    \aemrt & 47.14 & 48.12 & 59.41 & 55.84& 64.48  & 62.66 \\
    \hline
    \oracle (All Tasks) & 64.70 & 63.90 & 73.20 & 71.91 &69.80 & 67.57 \\
    \bottomrule
  \end{tabular}%
  }
    \caption{$\text{ACC}_{\text{whole}}$ and $\text{ACC}_{\text{avg}}$ (\%) of parsers using BERT by fine-tuning (Up) and fixing (Bottom) BERT's parameters.  }
  \label{tab:bert}
 
\end{table}

\paragraph*{Training Time Analysis}
\begin{table}[ht!]
\centering
  \resizebox{\textwidth}{!}{%
  \begin{tabular}{|c|ccc|}
    \toprule
    Methods & \overnight & \nlmapq
       &
      \nlmapc \\

      \midrule
    Fine-tune & 54.19 &  360.09 & 110.12  \\
        \hline
    \ewc &  71.66 &  541.20 & 223.02  \\
    \hatt &  53.85 &  212.63 & 128.67  \\
    \gem &  94.67 &  389.04 & 259.80     \\
    \emr &  102.97 &  399.05 & 214.95  \\
    \emar &  139.44 &  402.64 & 240.64  \\
    \arper & 160.05 &  901.33 & 654.51 \\
    \proto & 148.04 &  490.95 & 304.45  \\
    \hline
    \aemrt &  117.63 &  549.22 & 275.09  \\
    \aemrt (+\ewc) & 124.49 &  540.68 & 282.87  \\
    \hline
    \oracle (All Tasks) &  712.43 &  1876.60 & 1531.05  \\
    \bottomrule
  \end{tabular}%
  }
    \caption{The average training time (seconds) of each continual learning method on one task. The training time of the \oracle setting is reported with training on all tasks. All the methods are running on a server with one Nvidia V100 and four cores of Intel i5 5400.}
  \label{tab:walltime}
\end{table}
The average training times of different CL approaches for \overnight, \nlmapc, and \nlmapq are depicted in~\tref{tab:walltime}. On average, the training time of \finetune is 13, 5, and 14 times faster than training the parser from scratch for the \overnight, \nlmapc, and \nlmapq tasks, respectively. In general, the training times of memory-based methods are longer than regularization and dynamic architecture methods due to the replay training. Since our method, \tr, is a memory-based method, its training time is comparable to the other memory-based methods such as \gem, \emr, and \emar. In addition, \ewc slows the convergence speed of the parsers on \nlmapc, and \nlmapq, thereby increasing the training time required for the parsers to achieve optimal performance for each task. Therefore, the hybrid method, \arper, that utilizes both \emr and \ewc takes the longest training time among all CL methods. However, our \awu could speed up the convergence of the base parser even with \ewc; thus, the training time of \tr (+EWC) is much less than that of \arper.

\section{Summary}
In this chapter, we conduct a comprehensive empirical investigation into CL in order to address the problems for semantic parsing under the limited computational resources.
To cope with catastrophic forgetting and facilitate knowledge transfer between tasks, this chapter presents \tr, consisting of a sampling method specifically designed for semantic parsing and a two-stage training method implementing an inductive bias for CL. On three benchmarks, the resulting parser achieves promising performance with only up to 3\% less accuracy than the \oracle parser, while the \oracle parser requires 3-6 times as much training time as \tr to adapt to a new task. In three benchmark settings, \tr also outperforms the existing CL baselines. Additionally, the ablation studies illustrate why \tr is effective.

\chapter{Conclusion}
\label{chap:conclusion}
\epigraph{In literature and in life we ultimately pursue, not conclusions, but beginnings.}{\textit{Sam Tanenhaus}}

This chapter concludes the contributions made to address the research questions under various limited resource conditions, followed by a discussion on the future directions for enhancing semantic parsing in these conditions.

\section{Summary}
This thesis proposes various techniques, including automatic data curation, knowledge transfer, AL, and CL, to tackle semantic parsing challenges in three limited resource scenarios:
\begin{enumerate}
\item No or limited parallel training data is available to adapt the semantic parser to a task in a new domain (\pref{part1}).
\item A constrained annotation budget exists for adapting a multilingual parser to a new language (\pref{part2}).
\item Limited computational resources are available for adapting the semantic parser to a new task (\pref{part3}).
\end{enumerate}

\paragraph*{Limited Parallel Data for Semantic Parsing.} To address semantic parsing in low-resource conditions with minimal or no parallel training data, Chapters \ref{chap:auto} and \ref{chap:meta} present novel automatic data curation and knowledge transfer approaches. These leverage different prior resources to generalize the semantic parser to a new domain or task.

\cref{chap:auto} introduces an automatic data curation method to produce an extensive amount of synthetic data. This aids in adapting a semantic parser to a new domain with no parallel training data, using only the structured data stored in a database. Inspired by earlier works~\citep{wang2015overnight,xu2020autoqa}, this method employs SCFG, written based on the database schema and data, to generate a vast collection of clunky utterance-LF pairs. These clunky utterances are then paraphrased into fluent NL utterances. The primary contribution of \cref{chap:auto} lies in evaluating the efficacy of various paraphrasing strategies, demonstrating that the performance of the semantic parser can be enhanced by selecting optimal paraphrasing techniques.

\cref{chap:meta} tackles issues of semantic parsing in an FSL scenario, where only one or two parallel utterance-LF pairs per new predicate for the target task are available while abundant data exists for the source task. The key contribution is a meta-learning technique combined with attention regularization, facilitating fast adaptation of the \parser, a transition-based neural semantic parser, to a new task by transferring knowledge from the source task. Experimental results confirm the efficiency of our method, showing its superior performance over all baselines in both one-shot and two-shot settings.

\paragraph*{Limited Annotation Budget for Multilingual Semantic Parsing.} Addressing challenges for multilingual semantic parsing in situations with a restricted budget for manual translation yet ample data in the source language, Chapters \ref{chap:al_msp} and \ref{chap:comb_human_auto} suggest approaches that use AL and MT to adapt the semantic parsers to new languages.

\cref{chap:al_msp} presents \almsp, a unique AL framework designed to gather training data for adapting the semantic parser to the target language. Given a restricted translation budget for HT and abundant data in the source language, \almsp iteratively samples examples. The HT of these examples is subsequently used to train the semantic parser for the target language. Additionally, this chapter introduces \lfslcd, a sampling method employed in \almsp that diversifies LF structures and lexical choices in the chosen examples. Experimental results show that \lfslcd outperforms other selection baselines given an equal translation budget. Using \lfslcd, \almsp notably reduces the HT cost, achieving comparable performance to a parser trained on a full human-translated training set in the target language with only 32\% of the source-language dataset manually translated.

In conditions similar to \cref{chap:al_msp} but with an additional machine translator, \cref{chap:comb_human_auto} introduces \hatt, an AL method for acquiring training data to maximize semantic parser performance in the target language. \hatt employs a machine translator to translate all source-language data and iteratively samples examples. The HT of these examples is then integrated into the machine-translated data, improving overall data quality. Ablation studies indicate that \amsp, the sampling method introduced in this chapter, substantially reduces biases and errors in the overall training data, leading to better parser performance compared to other sampling baselines. Moreover, given a manual translation budget, a parser trained on the mixed data generated by \hatt significantly outperforms those trained exclusively on machine-translated data or human-translated data obtained through \almsp.

\paragraph*{Limited Computational Resource for Semantic Parsing.}
\cref{chap:continual} introduces \tr, a CL method tailored for neural semantic parsers. \tr aims to decrease training time and memory usage while adapting the semantic parser to a new task. Specifically, the method comprises \blfs, a technique that samples diversified examples to boost the ability of EMR~\citep{wang2019sentence} in preventing catastrophic forgetting, and \awu, a strategy that fine-tunes action embeddings in a specific manner to encourage forward knowledge transfer across tasks. Experiments reveal that semantic parser training can be accelerated 3-6 times across three benchmarks without compromising performance on prior tasks.

\section{Future Directions}

This section discusses potential future research directions for tackling the challenges associated with semantic parsing under limited resource conditions.

\paragraph*{Distribution Mismatch.} The automatic data curation approach automatically generates a large amount of synthetic data, whose data distribution differs from the true data distributions of the current task. Consequently, training the parser on such data compromises the performance of the parser. Multiple approaches~\citep{herzig2019don,qiu2021improving} suggested resolving this issue by learning the true distribution from unlabeled or labeled data. These approaches, however, require extensive data resources for the current task, which are sometimes unavailable due to limited data resource constraints. One promising research direction is to leverage prior knowledge from various sources, such as the source task with a close data distribution to the current task or the pre-trained language model, to obtain the true distribution. Then, the true distribution can improve the quality of the synthetic data generated.

\paragraph*{Multilingual Transfer Learning.} In a multilingual transfer scenario, one potential topic of future research would be to improve the semantic parser in the low-resource language by transferring knowledge from the high-resource languages. Transfer learning can be employed with AL or MT to enhance the low-resource multilingual semantic parser.

\paragraph*{Continual Learning for Semantic Parsing with Pre-trained Language Models.} \cref{chap:continual} proposed a hybrid CL method for a vanilla \Seq semantic parser. However, the semantic parser using a pre-trained language model has demonstrated strong generalization ability in many semantic parsing studies. \cref{chap:continual} conducted a preliminary study in CL for the semantic parsers with the pre-trained language models, yet many areas still need to be explored. For example, how to prevent catastrophic forgetting for the pre-trained language model in the semantic parsing task? How can the forward transfer for semantic parsing with the pre-trained language model be facilitated? These questions shed light on promising future directions.







\addtocontents{toc}{\vspace{2em}} 

\appendix 

\chapter{An Appendix}

\section{Appendix for~\cref{chap:meta}}
\subsection{Template Normalization}
\label{appendix:meta}
\begin{algorithm}[t]
{\small
\SetKwData{Left}{left}\SetKwData{This}{this}\SetKwData{Up}{up}
\SetKwFunction{Union}{Union}\SetKwFunction{FindCompress}{FindCompress}
\SetKwInOut{Input}{Input}\SetKwInOut{Output}{Output}
\SetAlgoLined
\Input{A set of abstract trees $\mathcal{T}$, a minimal support $\tau$}
\Output{A set of normalized trees}
$O$ := mapping of subtrees to their occurrences in $\mathcal{T}$. \\
\For{tree $t$ in $\mathcal{T}$}{
    update occurrence of all leaf nodes $v$ of $t$ to $O[v]$
}
\While{$O$ updated with new trees}{
\For{tree $t$, occur\_list $l$ in $\mathcal{O}$}{
   build occurrence list $l'$ for supertree $t'$ of $t$\\
   \If{size($l'$) $\geq$ size($l$)}
   {
     $O[t'] = l'$ 
   }
}
}
\For{tree $t$, occur\_list $l$ in $\mathcal{O}$}{
    \If{size($l$) $\geq \tau$}
    {
        collapse $t$ into a node for all $t'$ in $l$.
    }
}
}
\caption{Template Normalization}
\label{algo:normalization}
\end{algorithm}

Many LF templates in the existing corpora have shared subtrees in the corresponding abstract semantic trees. The tree normalization algorithm aims to treat those subtrees as single units. The identification of such a shared structure is conducted by finding frequent subtrees. Given an LF dataset, the \textit{support} of a tree $t$ is the number of LFs that occur as a subtree. We call a tree \textit{frequent} if its support is greater and equal to pre-specified minimal support.

We also observe that in an LF dataset, some frequent subtrees always have the same supertree. For example, \pre{ground\_fare \$1} is always the child of \textit{=( $\dots$, \$0 )} in the whole dataset. We call a subtree \textit{complete} w.r.t. a dataset if any of its supertrees in the dataset occur significantly more often than that subtree. Another observation is that some tree nodes have fixed siblings. In order to check if two tree nodes sharing the same root are fixed siblings, we merge the two tree paths together. If the merged tree has the same support as that of the two trees, we call the two trees pass the fixed sibling test. In the same manner, we collapse tree nodes with fixed siblings, as well as their parent node into a single tree node to save unnecessary parse actions.

Thus, the normalization is conducted by collapsing a frequent complete abstract subtree into a tree node. We call a tree \textit{normalized} if all its frequent complete abstract subtrees are collapsed into the corresponding tree nodes. The pseudocode of the tree normalization algorithm is provided in Algorithm~\ref{algo:normalization}.	

\addtocontents{toc}{\vspace{2em}}  
\backmatter

\label{Bibliography}
\lhead{\emph{Bibliography}}  
\bibliographystyle{acl_natbib}  
\bibliography{Bibliography}  

\end{document}